\def\paperID{*****}
\def\confName{CVPR}
\def\confYear{2026}

\def\paperTitle{DyWeight: Dynamic Gradient Weighting for Few-Step Diffusion Sampling \vspace{-4mm}}

\def\authorBlock{
    Tong Zhao$^{1,2,*}$, 
    Mingkun Lei$^{2,*}$, 
    Liangyu Yuan$^{3,*}$, \\
    Yanming Yang$^{2}$, 
    Chenxi Song$^{2}$, 
    Yang Wang$^{2}$, 
    Beier Zhu$^{4}$, 
    Chi Zhang$^{2,\dagger}$ \\ [6pt]
    $^{1}$Zhejiang University, \quad $^{2}$AGI Lab, Westlake University, \\ 
    $^{3}$TongJi University, \quad $^{4}$University of Science and Technology of China \\[4pt]
    
    {\url{https://github.com/Westlake-AGI-Lab/DyWeight}}
}

\newif\ifreview 
\newif\ifarxiv \newcommand{\arxiv}{\arxivtrue}
\newif\ifcamera 
\newif\ifrebuttal 
\arxiv 
\pdfoutput=1
\documentclass[10pt,twocolumn,letterpaper]{article}
\ifreview \usepackage[review]{cvpr} \fi
\ifarxiv \usepackage[pagenumbers]{cvpr} \fi
\ifrebuttal \usepackage[rebuttal]{cvpr} \fi
\ifcamera \usepackage{cvpr} \fi

\usepackage{amsfonts,bm}
\usepackage{nccmath}
\usepackage{graphicx}	
\usepackage{amsmath}	
\usepackage{amssymb}	
\usepackage{booktabs}
\usepackage{times}
\usepackage{microtype}
\usepackage{epsfig}
\usepackage{caption}
\usepackage{float}
\usepackage{placeins}
\usepackage{color, colortbl}
\usepackage{stfloats}
\usepackage{enumitem}
\usepackage{tabularx}
\usepackage{xstring}
\usepackage{multirow}
\usepackage{xspace}
\usepackage{url}
\usepackage{subcaption}
\usepackage{xcolor}
\usepackage[hang,flushmargin]{footmisc}
\usepackage{algorithm}
\usepackage{wrapfig}
\usepackage{algorithmic}
\usepackage{titletoc}

\usepackage[most]{tcolorbox} 

\newtcolorbox{observationbox}{
  colback=gray!8,      
  colframe=gray!40,    
  boxrule=0.5pt,       
  arc=3pt,             
  left=6pt, right=6pt, top=4pt, bottom=4pt,
  enhanced, breakable, 
}

\if camera \usepackage[accsupp]{axessibility} \fi

% ...

\if arxiv  \fi

\newcommand{\R}[1]{{%
    \textbf{%
        \ifstrequal{#1}{1}{\textcolor{red}{R#1}}{%
        \ifstrequal{#1}{2}{\textcolor{blue}{R#1}}{%
        \ifstrequal{#1}{3}{\textcolor{magenta}{R#1}}{%
        \ifstrequal{#1}{4}{\textcolor{teal}{R#1}}{%
                           \textcolor{cyan}{R#1}%
        }}}}%
    }%
}}

\def\eqref#1{equation~\ref{#1}}

\def\1{\bm{1}}

\DeclareMathAlphabet{\mathsfit}{\encodingdefault}{\sfdefault}{m}{sl}
\SetMathAlphabet{\mathsfit}{bold}{\encodingdefault}{\sfdefault}{bx}{n}

\usepackage{xr-hyper}

\makeatletter
\newcommand*{\addFileDependency}[1]{
  \typeout{(#1)}
  \@addtofilelist{#1}
  \IfFileExists{#1}{}{\typeout{No file #1.}}
}

\makeatother

\definecolor{cvprblue}{rgb}{0.21,0.49,0.74}
\usepackage[pagebackref,breaklinks,colorlinks,allcolors=cvprblue]{hyperref}
\usepackage[capitalize]{cleveref}
\crefname{section}{Sec.}{Secs.}
\crefname{table}{Table}{Tables}
\crefname{figure}{Fig.}{Figs.}

\ifarxiv \crefname{appendix}{App.}{Apps.}
\else \crefname{appendix}{Suppl.}{Suppls.} \fi

\begin{document}
\newcommand{\ours}{\texttt{DyWeight}}

\newcommand{\tableCellHeight}{1}
\newcommand{\tabstyle}[1]{
  \setlength{\tabcolsep}{#1}
  \renewcommand{\arraystretch}{\tableCellHeight}
  \centering
  \small
}

\definecolor{tabhighlight}{HTML}{e5e5e5}
\definecolor{lightCyan}{rgb}{1,0.95,0.98} 
\definecolor{paramcolor}{rgb}{0.9, 0.2, 0.5}  
\definecolor{commentcolor}{rgb}{0.0, 0.0, 1.0} 
\definecolor{color_blue}{HTML}{2E86AB}  
\definecolor{color_red}{HTML}{A23B72}  

\newtheorem{definition}{Definition}
\newtheorem{theorem}{Theorem}
\newtheorem{assumption}{Assumption}
\newtheorem{lemma}{Lemma}
\newtheorem{proposition}{Proposition}
\newtheorem{corollary}{Corollary}
\title{\paperTitle}
\vspace{-4mm}
\author{\authorBlock}
\twocolumn[{%
\renewcommand\twocolumn[1][]{#1}%
\maketitle
\vspace{-10mm}
\begin{center}
    \centering
    \includegraphics[width=\linewidth]{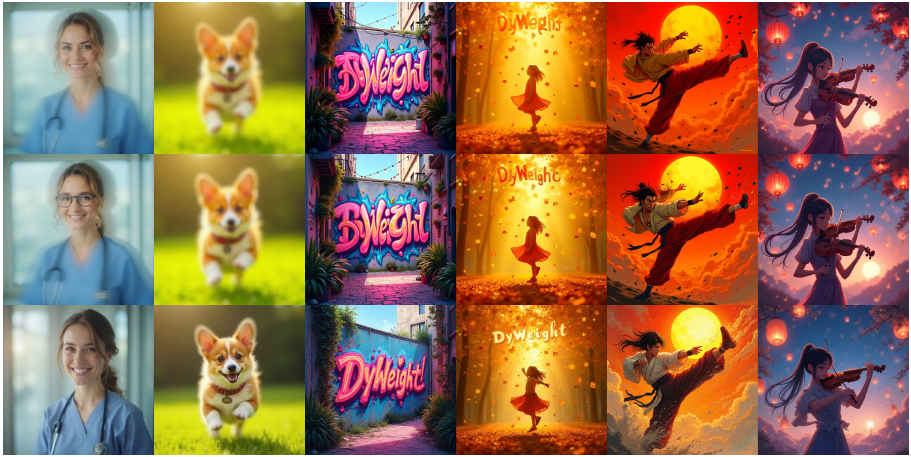}
    \vspace{-5mm}
    \captionof{figure}{Qualitative comparison on FLUX.1-dev \cite{flux,flux_paper} at 7 NFEs. From top to bottom, the rows show results from DPM-Solver++(2M) \cite{dpmpp}, iPNDM(2M) \cite{PNDM,iPNDM}, and our \ours. Our method delivers superior visual fidelity, prompt alignment, and structural coherence.}
    \label{fig:teaser}
\end{center}
}]

\begin{NoHyper}
\let\thefootnote\relax\footnotetext{
    $^{*}$Equal contribution. $^{\dagger}$Corresponding author. \\
    Emails: \texttt{\{zhaotong68, chizhang\}@westlake.edu.cn}
}
\end{NoHyper}
\begin{abstract}
Diffusion Models (DMs) have achieved state-of-the-art generative performance across multiple modalities, yet their sampling process remains prohibitively slow due to the need for hundreds of function evaluations. Recent progress in multi-step ODE solvers has greatly improved efficiency by reusing historical gradients, but existing methods rely on handcrafted coefficients that fail to adapt to the non-stationary dynamics of diffusion sampling. To address this limitation, we propose \texttt{Dy}namic Gradient \texttt{Weight}ing (\ours), a lightweight, learning-based multi-step solver that introduces a streamlined implicit coupling paradigm. By relaxing classical numerical constraints, \ours~learns unconstrained time-varying parameters that adaptively aggregate historical gradients while intrinsically scaling the effective step size. This implicit time calibration accurately aligns the solver's numerical trajectory with the model's internal denoising dynamics under large integration steps, avoiding complex decoupled parameterizations and optimizations. Extensive experiments on CIFAR-10, FFHQ, AFHQv2, ImageNet64, LSUN-Bedroom, Stable Diffusion and FLUX.1-dev demonstrate that \ours~achieves superior visual fidelity and stability with significantly fewer function evaluations, establishing a new state-of-the-art among efficient diffusion solvers. Code is available at \url{https://github.com/Westlake-AGI-Lab/DyWeight}.
\end{abstract}    
\section{Introduction}
\label{sec:intro}

Diffusion Models (DMs) \cite{diffusion,DDPM,score_based,Flow} have established a new state-of-the-art in generative modeling, capable of synthesizing data with unprecedented fidelity and diversity across modalities such as images \cite{flux,sd3,qwen}, audio \cite{audio,audio_step}, and video \cite{wan,sora,hunyuanvideo}. At their core, DMs transform a simple noise distribution into complex data through the reversal of a predefined noising process, which can be mathematically formulated as solving a corresponding Probability Flow Ordinary Differential Equation (PF-ODE) \cite{DDIM,SDE}. 

Despite their remarkable performance, diffusion models suffer from inherently slow sampling speed \cite{DDPM,DDIM}. Generating a single high-quality sample typically requires evaluating a large neural network (\eg, U-Net \cite{unet}) hundreds or even thousands of times, leading to substantial computational latency. This inefficiency poses a significant barrier for real-time and interactive applications, motivating extensive research on efficient numerical solvers to accelerate diffusion sampling \cite{DDIM,EDM,PNDM,iPNDM,dpmsolver,dpmpp,dpmv3,UniPC,geometric,gits}. Among various acceleration approaches, explicit multi-step ODE solvers (\eg, DPM-Solver++\cite{dpmpp} and iPNDM\cite{iPNDM}) have emerged as particularly promising due to their ability to exploit historical gradient information. By reusing the model’s past denoising predictions as a ``free lunch'', these solvers can improve accuracy at negligible extra cost, offering a compelling trade-off between quality and speed.

Although multi-step solvers improve efficiency, their handcrafted coefficients are derived under the assumption that denoising gradients at neighboring timesteps share similar statistics \cite{PNDM,iPNDM}. This stationarity assumption, however, does not hold in diffusion sampling, where the dynamics are highly non-stationary across time. As a result, a fixed weighting scheme fails to adapt to these changing dynamics, early noisy gradients may overpower or conflict with later, more reliable ones, causing accumulated bias and degraded image quality, particularly under large step sizes.
Furthermore, the coefficients in these solvers are designed based on classical ODE assumptions of smooth, low-order trajectories within small integration intervals \cite{numerical_analysis,numerical_ode,adam}. When extended to few-step sampling with large step gaps, these assumptions break down, introducing severe truncation errors. 

Recently, learning-based solvers have attempted to address these limitations by optimizing both solver coefficients and time schedules \cite{LD3,DSS,DLMS,s4s,AMED,EPD}. However, these approaches typically adopt an \emph{explicitly decoupled} optimization paradigm, maintaining an independent time sequence alongside the solver coefficients. To implement this, both components must be parameterized independently: the gradient weights are typically constrained to satisfy classical numerical rules, while the time sequence is parameterized with principled mathematical functions, such as cumulative softmax, to ensure valid and strictly decreasing steps. While functionally valid, this decoupled formulation yields a notoriously complex optimization landscape when solver coefficients and time discretization variables are jointly optimized, often necessitating cumbersome alternating strategies (\eg, multi-loop bi-level optimization) \cite{s4s}.

To overcome these bottlenecks in few-step sampling, we propose \texttt{Dy}namic Gradient \texttt{Weight}ing (\ours), a learning-based multi-step solver that introduces a streamlined \emph{implicit coupling} paradigm. Rather than maintaining a separate, heavily constrained time sequence, \ours~elegantly relaxes the classical numerical constraint that gradient weights must sum to one (\ie, not requiring $\sum w = 1$). This simple yet profound relaxation yields a dual benefit: the normalized part of the weights adaptively learns to aggregate historical gradients, while the unnormalized sum inherently acts as an \emph{implicit time shifting} mechanism that dynamically adjusts the macro-level integration boundary. This eliminates the need for complex timeline parameterizations, transforming a difficult joint-optimization problem into a highly efficient, single-round learning process. Furthermore, to correct the misalignment between the numerical timeline and the network's internal noise expectations \cite{exposure_alleviating,exposure_elucidating} caused by large step gaps, we supplement our method with a micro-level \emph{time scaling} to adjust the exact query time fed into the neural network. Together, implicit time shifting and scaling constitute our time calibration mechanism. 

To optimize these parameters efficiently, we employ a training-data-free distillation framework using on-the-fly teacher generation \cite{AMED,EPD}. By relying solely on endpoint supervision \cite{LD3} rather than rigid intermediate path supervision, the solver is granted the flexibility to deviate from the teacher mid-way while achieving more accurate alignment with the data manifold at the final step. Our contributions are summarized as follows:

\begin{itemize}
\item We propose \ours, a learning-based solver that replaces complex, decoupled parameterization with a relaxed weight formulation, implicitly coupling gradient aggregation with timestep adjustments.

\item We introduce a \emph{time calibration} strategy combining implicit time shifting and time scaling, enabling the solver to maintain temporal alignment with the diffusion model's internal dynamics in the few-step sampling regime.

\item Extensive experiments demonstrate that ours method achieves state-of-the-art visual fidelity with significantly fewer NFEs across diverse benchmarks, including complex flow-matching models like FLUX.1-dev \cite{flux,flux_paper}, while requiring minimal optimization overhead.
\end{itemize}
\section{Related work}
\begin{figure*}[t]
\centering
    \includegraphics[width=.95\linewidth]{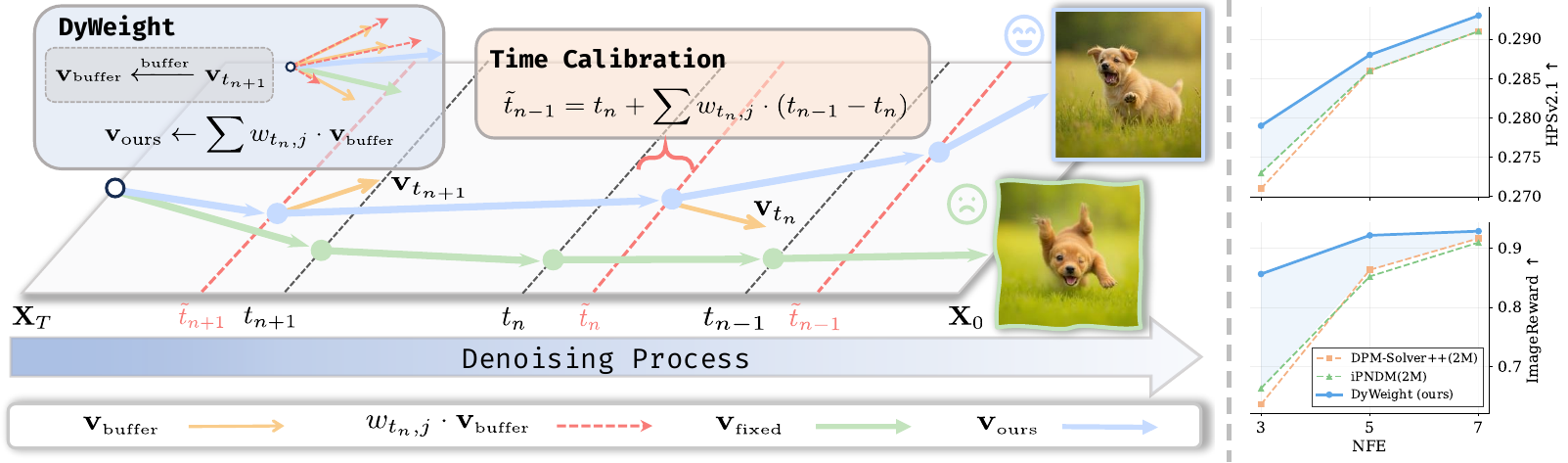}
    \caption{Overview of \ours. Our method dynamically aggregates buffered historical gradients ($w_{t_n,j} \cdot \mathbf{v}_{\text{buffer}}$) using unconstrained learned weights (\ie, not requiring $\sum w = 1$). The unnormalized sum part adjusts the effective step size via time calibration ($t_n \rightarrow \textcolor[RGB]{248, 123, 119}{\tilde{t}_{n}}$). As illustrated on the right, \ours~exhibits a distinct advantage in few-step generation scenarios.}
    \label{fig_pipeline}
\end{figure*}

\noindent
\textbf{Multi-step diffusion solvers.} A primary approach to accelerating DPM sampling is the design of efficient, training-free ODE solvers that reduce truncation error under few steps. Solvers are broadly categorized as single-step \cite{DDIM,EDM,dpmsolver} or multi-step \cite{dpmpp,dpmv3,PNDM,iPNDM,UniPC}. Explicit multi-step solvers \cite{dpmpp,iPNDM,PNDM,UniPC} are popular as they exploit historical gradients for higher accuracy without extra NFEs. Their main limitation is the use of handcrafted coefficients, typically derived from generic polynomial assumptions \cite{adam,numerical_analysis,numerical_ode}, which may be suboptimal for diffusion dynamics, especially under large step sizes and few steps.

\vspace{0.3em}
\noindent
\textbf{Learning-based solvers.} This paradigm optimizes solver parameters by distilling knowledge from a high-NFE teacher, keeping the diffusion model fixed \cite{bespoke,learningfast,s4s,D-ODE}. These learnable parameters vary, such as time schedule \cite{LD3,timesteptuner,DMN,ays} or single-step gradient scaling \cite{AMED,EPD}. Recent methods, including S4S \cite{s4s}, DLMS \cite{DLMS}, and DSS \cite{DSS}, share our motivation to move beyond handcrafted weights but diverge fundamentally in the execution paradigms. DLMS \cite{DLMS} predicts solver coefficients at runtime with an auxiliary network, introducing additional inference cost and optimization complexity. Both DLMS \cite{DLMS} and DSS \cite{DSS} rely on path supervision, encouraging the solver to closely follow the teacher trajectory at each step, whereas our method uses end-point supervision, allowing flexibility to discover more efficient trajectories while still aligning with the data manifold. Moreover, all these methods decouple time steps and solver coefficients, learning them as independent variables or explicitly predicting the next time step. This decoupled modeling often necessitates complex, multi-round optimization strategies (\eg, S4S-Alt \cite{s4s} requires a $K=8$ loop bi-level optimization). Conversely, our method implicitly couples temporal adjustment with gradient weights via a relaxed weight sum formulation, enabling adaptive aggregation of historical gradients while modulating the effective step size. This avoids separate time parameterization and enables efficient single-round optimization.
\section{Methods}
\label{sec:method}

In this section, we first introduce the preliminaries in \cref{preliminary}. \cref{sec:numerical} presents a simple numerical experiment that analyzes the behavior of multi-step solvers, and \cref{dy-solver-method} provides a detailed description of our proposed methods.

\subsection{Preliminary}
\label{preliminary}
\noindent
\textbf{Diffusion model.} Diffusion Models (DMs) \cite{diffusion,DDPM,score_based} progressively map a data distribution $\mathbf{x}_0 \!\sim\! {p}_{\text{data}}$ to a simple prior, typically Gaussian noise $\mathbf{x}_T \!\sim\!\mathcal{N}(0, \mathbf{I})$, via a forward noising process. At inference, the model reverses this process, transforming noise back into data through a learned denoising function $\bm{\epsilon}_\theta(\mathbf{x}_t,t)$, which can be directly modeled as an ordinary differential equation (ODE) \cite{SDE,vdm}:
\begin{equation}
    \frac{\mathrm{d}\mathbf{x}_t}{\mathrm{d}t}=f(\mathbf{x}_t,t)+\frac{g^2(t)}{2\sigma_t}\bm{\epsilon}_\theta(\mathbf{x}_t,t),
    \label{ode}
\end{equation}
where $f(\mathbf{x}_t,t)$ is the drift coefficient, $g(t)$ is the the diffusion coefficient, $\sigma_t$ represents the noise level at time $t$. In this paper, we follow the parameterization of EDM \cite{EDM}, which sets $f(\mathbf{x}_t,t) = 0$, $g(t) = \sqrt{2t}$, and $\sigma_t = t$. This parameterization significantly simplifies \eqref{ode} to:  
\begin{equation}
    \frac{\mathrm{d}\mathbf{x}_t}{\mathrm{d}t}=\bm{\epsilon}_\theta(\mathbf{x}_t,t)=\frac{\mathbf{x}_t-\mathcal{D}_\theta(\mathbf{x}_t,t)}{t}.
    \label{ode_simplified}
\end{equation}

Here, $\bm{\epsilon}_\theta(\mathbf{x}_t, t)$ is the noise prediction model \cite{DDPM,DDIM}. The second equality defines the data prediction model $\mathcal{D}_\theta(\mathbf{x}_t, t) = \mathbf{x}_t - t \cdot \bm{\epsilon}_\theta(\mathbf{x}_t, t)$, which is trained to directly estimate the clean data $\mathbf{x}_0$ \cite{EDM,dpmpp,UniPC}. Therefore, the generation process is equivalent to solving the ODE in \eqref{ode_simplified} from an initial condition $\mathbf{x}_T$. We can express the exact solution for a single step:
\begin{equation}
    \mathbf{x}_{t_{n-1}} 
    =\mathbf{x}_{t_n} + \int_{t_n}^{t_{n-1}} \bm{\epsilon}_\theta(\mathbf{x}_t,t) \mathrm{d}t.
    \label{integral}
\end{equation}

Since the integral of the neural network $\bm{\epsilon}_\theta$ over the unknown path $\mathbf{x}_t$ is intractable, numerical methods are required to approximate this integral. A baseline approach, used by DDIM \cite{DDIM}, applies the first-order Euler method:
\begin{equation}
    \mathbf{x}_{t_{n-1}}=\mathbf{x}_{t_n} + 
    \bm{\epsilon}_\theta(\mathbf{x}_{t_n},t_n)\cdot(t_{n-1} - t_{n}).
    \label{sum}
\end{equation}

The core challenge of diffusion sampling thus becomes how to accurately and efficiently approximate the intractable integral in \eqref{integral}. This necessitates the use of high-order numerical solvers \cite{PNDM,EDM} beyond the simple first-order method.

\vspace{0.5em}
\noindent
\textbf{Multi-step diffusion solvers.} To improve upon the first-order approximation and reduce truncation error, a variety of higher-order solvers have been developed. Among them, implicit single-step methods (\eg, Heun’s method \cite{EDM}, DPM-Solver \cite{dpmsolver}) achieve higher accuracy but require evaluating the model $\bm{\epsilon}_\theta$ at future timesteps, typically doubling the number of function evaluations (NFE) per step, and often degrade when the total number of steps is very small.

Consequently, explicit multi-step solvers \cite{PNDM,iPNDM,dpmpp,UniPC} are often preferred. These methods leverage historical information from previous steps to achieve a high-order approximation without additional NFE. A general form for a $k$-order explicit multi-step solver can be written as:
\begin{equation}
    \mathbf{x}_{t_{n-1}}=\mathbf{x}_{t_n} + (t_{n-1} - t_{n})
     \sum_{i=0}^{k-1}
    w_{i}\cdot\underbrace{\bm{\epsilon}_\theta(\mathbf{x}_{t_{n+i}},t_{n+i})}_{\text{gradient term}}.
    \label{mult_gereral}
\end{equation}

Different solvers utilize different strategies to derive these weights $w_{i}$. DPM-Solvers \cite{dpmsolver,dpmpp,dpmv3} leverage the semi-linear property of the underlying ODE and apply a Taylor expansion to incorporate high-order information. Meanwhile, iPNDM \cite{PNDM,iPNDM} approximates the integrand $\bm{\epsilon}_\theta(\mathbf{x}_t, t)$ by constructing an extrapolating polynomial $\mathbf{P}_k(t)$ that passes through the $k$ most recent gradients. When the time steps are uniform, \eg, $t_{n-1} - t_{n} = t_{n-2} - t_{n-1}=h$, the integral of the Lagrange polynomial simplifies to the classic $k$-th order Adams-Bashforth (AB) method \cite{adam,numerical_analysis,numerical_ode}:
{\small \begin{equation}
h \cdot
\left\{
\begin{aligned}
&\left( 3\bm{\epsilon}_n - \bm{\epsilon}_{n+1} \right)/2, \!\!& \text{if } k=2, \\[4pt]
&\left( 23\bm{\epsilon}_n - 16\bm{\epsilon}_{n+1} + 5\bm{\epsilon}_{n+2} \right)/12, \!\! & \text{if } k=3, \\[4pt]
&\left( 55\bm{\epsilon}_n - 59\bm{\epsilon}_{n+1} + 37\bm{\epsilon}_{n+2} - 9\bm{\epsilon}_{n+3} \right)/24, \!\!& \text{if } k=4.
\end{aligned}
\right.
\label{eq:ab}
\end{equation}}

While the classic AB methods in \cref{eq:ab} are derived for uniform steps $h$, practical time schedules \cite{DDPM,EDM,dpmsolver,UniPC} are often highly non-linear in $t$. To better handle this, solvers are often formulated and implemented in a re-parameterized space, such as the log-signal-to-noise ratio (log-SNR \cite{vdm,dpmsolver,UniPC}), to ensure stability.

\subsection{Numerical study of solver coefficients}
\label{sec:numerical}
The core hypothesis of our work is that the time-invariant, handcrafted coefficients of classical high-order ODE solvers are suboptimal for accelerating diffusion sampling, particularly under two challenges: $(\mathbf{1})$ the extreme complexity of the denoising network's learned function and $(\mathbf{2})$ the aggressive reduction in solver steps required for fast sampling. To investigate this hypothesis, we first step away from the ``black-box'' diffusion model to build a ``white-box'' numerical experiment where all variables are known and controllable.

\vspace{0.5em}
\noindent
\textbf{Setup.} To isolate the impact of solver coefficients under controlled yet realistic dynamics, we construct a synthetic ODE system $\mathbf{y}' = \mathbf{f}(t, \mathbf{y})$ on $t \in [0,1]$ with state dimension $D = 50$. The ground-truth trajectory $\mathbf{y}_\text{gt}(t)$ is defined as a $D$-dimensional polynomial of degree $K$, whose coefficients $\mathbf{C} \in \mathbb{R}^{D \times (K+1)}$ are sampled from $\mathcal{N}(0, 4\mathbf{I})$. This design provides closed-form access to both $\mathbf{y}_\text{gt}(t)$ and its derivative $\mathbf{y}'_\text{gt}(t)$, allowing us to compute numerical errors exactly. We then define the ODE drift as:
\begin{equation}
    \mathbf{f}(t, \mathbf{y}) = \mathbf{y}'_\text{gt}(t) + \mathbf{A}\big[\mathbf{y}(t) - \mathbf{y}_\text{gt}(t)\big],
\end{equation}
where $\mathbf{A}$ is a random coupling matrix that introduces stiffness and cross-dimensional interactions. This term mimics the complex, highly coupled dynamics induced by high-capacity diffusion backbones such as U-Net~\cite{sd1.5,sdxl,unet} or DiT~\cite{dit,sd3,flux,flux_paper}.

On this testbed, we evaluate Adams–Bashforth (AB) solvers \cite{adam} of orders 1–4 in two variants:
\begin{itemize}
    \item \textbf{Standard}: Classical AB solvers with fixed analytical coefficients from numerical analysis~\cite{numerical_analysis,numerical_ode}.
    \item \textbf{Optimized}: AB solvers with learnable, time-dependent coefficients. For an $O$-th order solver with $N$ steps, we optimize a weight matrix of size $N \times O$, where each row parameterizes the coefficients used at a specific timestep.
\end{itemize}

All methods integrate over $t \in [0,1]$, initialized from the same four ground-truth states to ensure an identical history for multi-step updates. The optimized solvers are trained to minimize the terminal error $||\mathbf{y}_\text{pred}(1) - \mathbf{y}_\text{gt}(1)||_2$ using Adam~\cite{opt_adam} for 2000 iterations per configuration. We initialize the learnable coefficients from the standard AB values to ensure a fair and stable comparison.

\begin{figure}[t]
\centering
\hspace{-2ex}
\subfloat[Fixed Steps $S=14$.]{
    \includegraphics[width=0.49\linewidth]{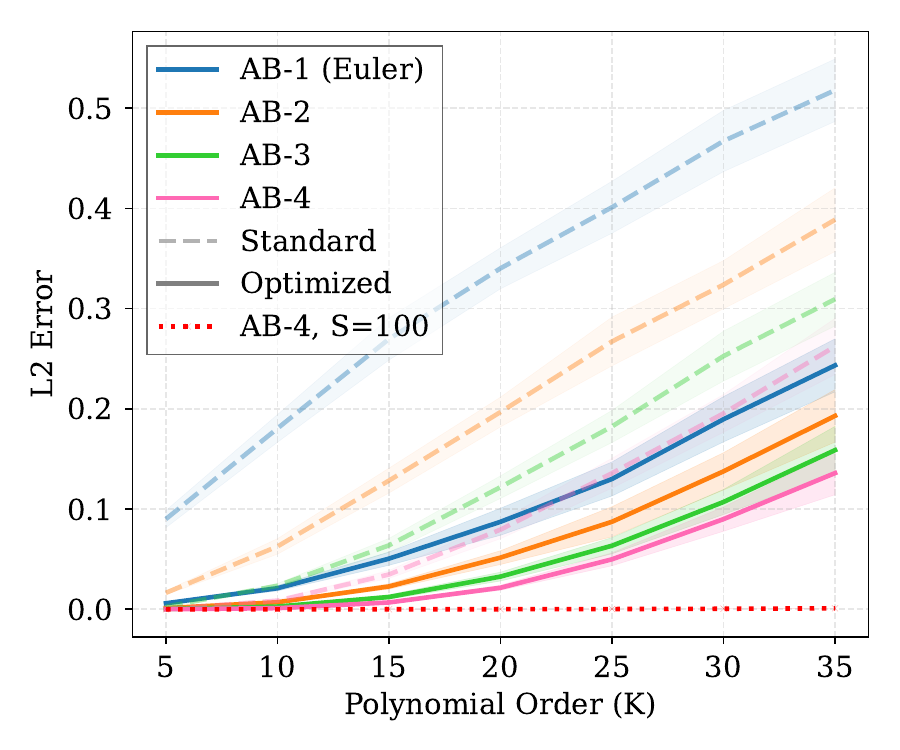}
    \label{fig:3_a}} 
\hspace{-2ex}
\subfloat[Fixed Complexity $K=20$.]{
    \includegraphics[width=0.49\linewidth]{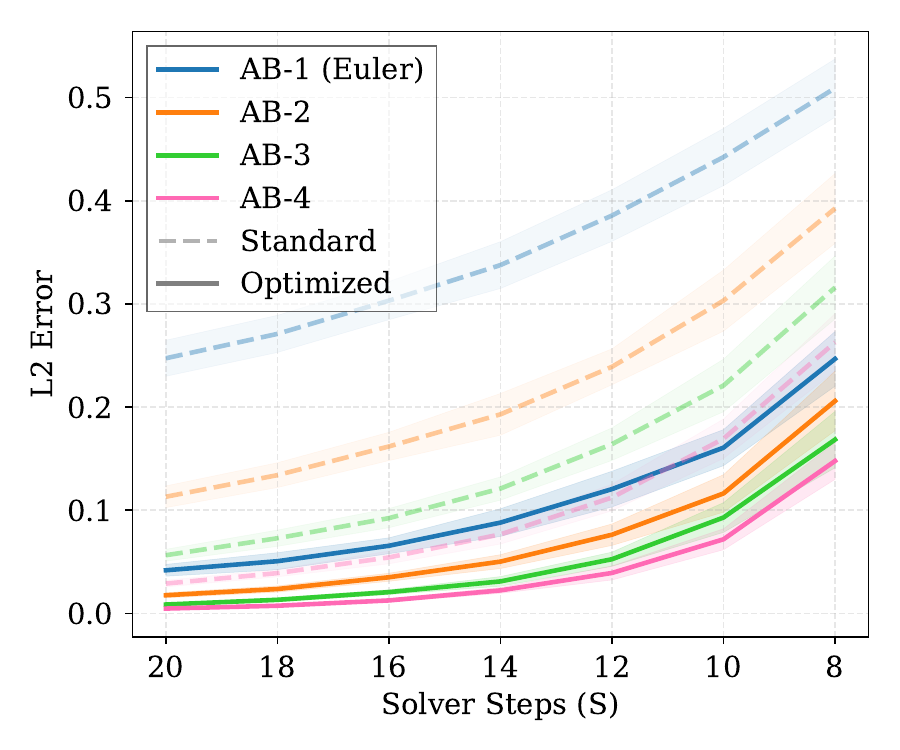}
    \label{fig:3_b}
    }
    \caption{L2 error under different complexity and steps settings. (a) Error vs. Complexity: We fix the solver steps ($S$) and increase the problem complexity (polynomial order $K$). (b) Error vs. Steps: We fix the complexity and decrease the number of solver steps.}
    \label{fig_toy}
    \vspace{-2mm}
\end{figure}

\noindent
\textbf{Observations.} We conduct two experiments, shown in \cref{fig_toy}, to analyze solver error under varying function complexity ($K$) and step counts ($S$), yielding four key observations:

\begin{observationbox}
\textit{Observation 1. }\textit{Multi-step methods effectively reduce global error.}
\end{observationbox}

Across both \cref{fig:3_a,fig:3_b}, a consistent hierarchy $\text{Error(AB-1)}\!>\!\text{Error(AB-2)}\!>\!\text{Error(AB-3)}\!>\!\text{Error(AB-4)}$ confirms the superiority of multi-step methods, which exploit historical gradients for improved accuracy over single-step Euler (AB-1).

\begin{observationbox}
\textit{Observation 2. }\textit{Error escalates with both increasing complexity and fewer steps.}
\end{observationbox}

As shown in \cref{fig:3_a}, higher polynomial orders $K$ lead to larger errors, reflecting the inherent difficulty of integrating complex dynamics akin to high-capacity diffusion networks \cite{sd3}. Similarly, \cref{fig:3_b} shows that reducing the number of solver steps $S$—as in fast diffusion sampling—leads to a sharp degradation in performance, consistent with the few-step quality drop reported in prior work \cite{dpmsolver,UniPC}.

\begin{observationbox}
\textit{Observation 3. }\textit{Handcrafted gradient weights are comparably suboptimal.}
\end{observationbox}

A pronounced gap between Standard (dashed) and Optimized (solid) solvers appears in all settings. For each order, the learned coefficients achieve consistently lower errors, suggesting that the classical coefficients, designed for general-purpose and small-step-size integration, are suboptimal for the high-error regime of few-step sampling.

\begin{observationbox}
\textit{Observation 4. }\textit{High-step trajectories provide reliable supervision.}
\end{observationbox}

In \cref{fig:3_a}, the ``Teacher'' baseline (red dotted line), which uses a standard AB-4 \cite{numerical_ode} solver with high precision ($S\!=\!100$), maintains a negligible, near-zero error across all complexities. This validates that a high-step-count trajectory serves as a reliable proxy for the true, (and in a real problem, unknowable) ground truth solution, consistent with prior findings in diffusion research \cite{AMED,LD3,DLMS}.

\vspace{0.5em}
\noindent
\textbf{Summary. }These results highlight the challenge of diffusion sampling: solving complex dynamics (high $K$, Obs. 2) with few steps (low $S$, Obs. 2), where classical multi-step solvers suffer large truncation errors from handcrafted, suboptimal coefficients (Obs. 3). This limitation can potentially be mitigated by using a high-step teacher as supervision (Obs. 4).

\subsection{Dynamic gradient weighting}

\begin{figure}[t]
\centering
\hspace{-2ex}
\subfloat[Standard deviation.]{
    \includegraphics[width=0.49\linewidth]{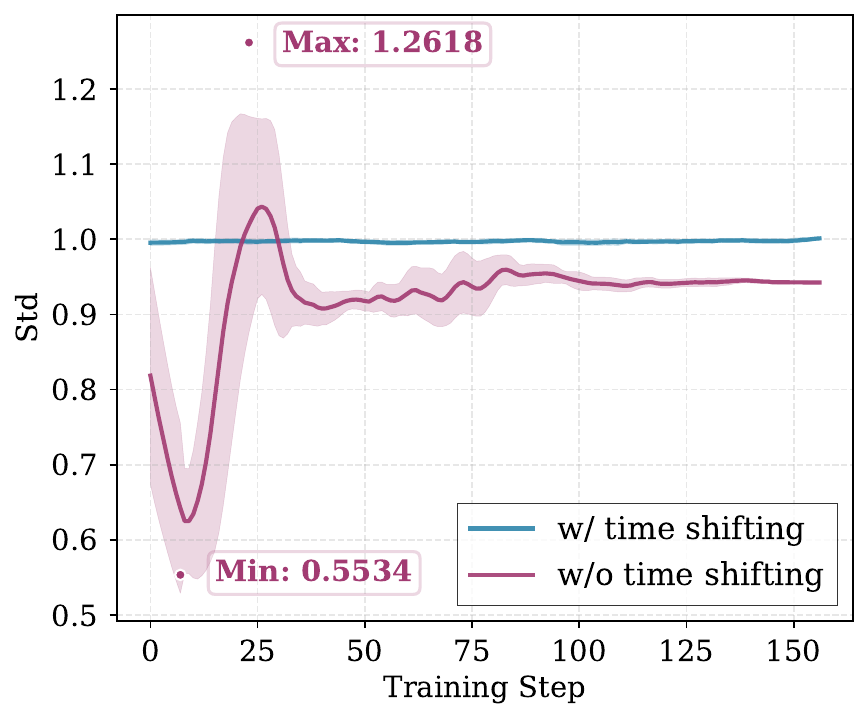}
    \label{fig4_a} 
    }
\hspace{-2ex}
\subfloat[Training loss.]{
    \includegraphics[width=0.49\linewidth]{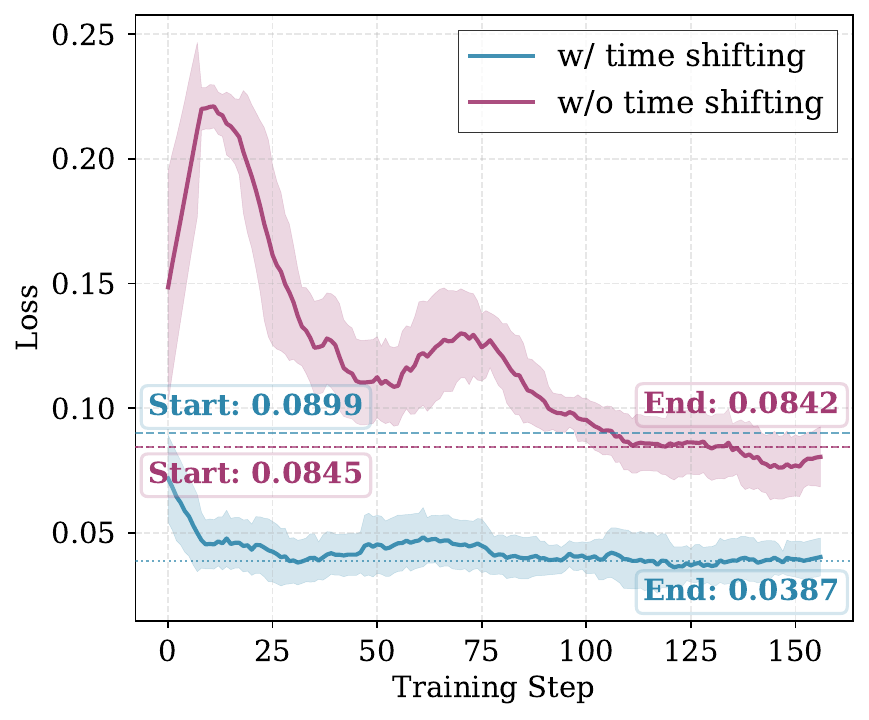}
    \label{fig4_b}
    }
    \vspace{-2mm}
    \caption{\textbf{Effect of Time Shifting on Training Stability.} We compare the (a) standard deviation of a gradient term specific to an intermediate point (expected$=\!1$) and (b) training loss for a 5-NFE student solver on CIFAR-10 \cite{cifar}. The solver w/o time shifting (\textcolor{color_red}{red}) exhibits extreme variance and an unstable loss. In contrast, our proposed time shifting (\textcolor{color_blue}{blue}) stabilizes the gradient variance around the expected value of 1.0 and achieves a smoother, lower-variance, and faster-converging training loss.}
    \label{fig_timeshifting}
    \vspace{-2mm}
\end{figure}

\label{dy-solver-method}
The results from our numerical experiments lead directly to the core motivation of \ours: rather than relying on time-invariant, handcrafted coefficients, we can train a specialized, time-varying student solver. Specifically, as illustrated in \cref{fig_pipeline}, our framework buffers historical gradients and adaptively aggregates them via unconstrained, learnable weights. This relaxed formulation enables the weights to not only determine the optimal gradient combination but also implicitly modulate the effective step size for dynamic time calibration. By using a pre-computed, multi-step teacher trajectory as a supervision signal, we can optimize the solver's coefficients, empowering it to \textit{dynamically weight} its history and \textit{align its temporal trajectory}, ultimately minimize the integration error tailored to a specific diffusion model and few-step sampling budget.

\vspace{0.5em}
\noindent
\textbf{Formulation.} We begin by re-examining the general k-order explicit multi-step solver in \eqref{mult_gereral}. The update rule can be equivalently expressed as:
{\small \begin{equation}
     \!\mathbf{x}_{t_{n-1}}\!\!=\mathbf{x}_{t_n}\!+ \!\underbrace{\sum_{i=0}^{k-1}w_{n,i}h_n}_\text{step scaling} 
     \underbrace{\sum_{i=0}^{k-1}\!\frac{w_{n,i}}{\sum_{i=0}^{k-1}w_{n,i}} \bm{\epsilon}_\theta(\mathbf{x}_{t_{n+i}},t_{n+i})}_\text{gradients weighting}.
    \label{dy-sum_split}
\end{equation}}

Here, $h_n\!=\!t_{n-1}\!-t_n$ denotes the time step size. This decomposition highlights two distinct mechanisms that emerge when the learnable weights $w_{n,i}$ are unconstrained (\ie, not requiring $\sum_{i=0}^{k-1} w_{n,i} = 1$). The normalized term acts as the \emph{gradient weighting}, which learns the optimal combination of historical gradients. The unnormalized pre-factor $\sum_{p=0}^{k-1} w_{n,i} h_n$, functions as an implicit \emph{step scaling}, which modulates the overall magnitude of the update. Classical solvers (\eg, Adams-Bashforth \cite{adam}) inherently constrain $\sum w_{n,i}\!=\!1$. We relax this constraint, enabling the solver to learn both the aggregation strategy and the update magnitude simultaneously.

\begin{algorithm}[t]
\caption{\texttt{DyWeight} Training}
\label{alg:training}
\begin{algorithmic}[1]
\STATE \textbf{Given:}\\
\hspace{1em}Teacher solver $\mathcal{S}$, teacher step $M$, student step $N$\\

\hspace{1em}Time schedules $\mathcal{T}^{\text{tea}}_M$, $\mathcal{T}_N^{\text{stu}}$ ($M \gg N$), order $K$\\

\hspace{1em}Network $\mathcal{D}_\theta$, condition $\mathbf{c}$, loss $\mathcal{L}$, learning rate $\eta$

\STATE \textbf{Initialize:} \\
\hspace{1em}Learnable params $\textcolor{paramcolor}{\Phi} = \{\textcolor{paramcolor}{\mathbf{W}}^{N \times K}, \textcolor{paramcolor}{\mathbf{s}}^{N \times 1}\}$ 

    \REPEAT
        \STATE Sample $\mathbf{x}_{T} \sim \mathcal{N}(0,\sigma_T^2\mathbf{I})$
        \vspace{1ex}
        
        \STATE $\mathbf{x}_{0}^{\text{tea}} \leftarrow \mathcal{S}(\mathcal{D}_\theta,\mathbf{x}_{T},\mathbf{c},\mathcal{T}^{\text{tea}}_M)$
        \vspace{1ex}
        
        \STATE $\mathbf{x}_{0}^{\text{stu}} \leftarrow \texttt{DyWeight}(\mathcal{D}_\theta,\mathbf{x}_{T},\mathbf{c},\textcolor{paramcolor}{\Phi},\mathcal{T}_N^{\text{stu}})$
        \vspace{1ex}
        
        \STATE Update $\textcolor{paramcolor}{\Phi} \leftarrow \textcolor{paramcolor}{\Phi}-\eta\nabla_{\textcolor{paramcolor}{\Phi}}\mathcal{L}(\mathbf{x}_{0}^{\text{stu}},\mathbf{x}_{0}^{\text{tea}})$
        \vspace{0.3ex}
    \UNTIL{converged}
    
\STATE \textbf{Return:} optimized $\textcolor{paramcolor}{\Phi}^*$
\end{algorithmic}
\end{algorithm}

\vspace{0.5em}
\noindent
\textbf{Gradient weighting parameterization.} Instead of using fixed, time-invariant coefficients, we parameterize the weights $w_{n,i}$ as time-varying and learnable. Specifically, for a $K$-order solver executing $N$ student steps, we introduce a learnable weight matrix $\mathbf{W}\!\in\!\mathbb{R}^{N \times K}$. At each integration step $n$ (from $N$ to 1), the solver utilizes the $n$-th row of this matrix, $\mathbf{w}_n\!=\![w_{n,0}, \dots, w_{n,i}]$, as the coefficients to aggregate the $p$ most recent gradient estimates $\bm{\epsilon}_\theta$, $p\!=\!\text{min}(N\!-\!n,K)$. This is detailed in \cref{alg:solver}. This formulation allows the solver to learn a specialized aggregation strategy for each distinct timestep $n$, effectively adapting to the non-stationary dynamics of the diffusion process.

\vspace{0.5em}
\noindent
\textbf{Time calibration.} Learning unconstrained weights $\mathbf{w}_n$ introduces a critical challenge: a mismatch between the solver's numerical step size and the diffusion model's internal time schedule. The effective step size, $\tilde{h}_n = \sum_{i=0}^{k-1} w_{n,i} h_n$, dynamically changes at each step. If this is not accounted for, the solver queries the denoising network $\bm{\epsilon}_\theta$ at times that are inconsistent with the true noise level, leading to gradient variance instability and training loss oscillation, as illustrated in \cref{fig_timeshifting}. We introduce the \emph{time shifting} mechanism to manage this mismatch, by reusing the unnormalized part of the learned weights to \textit{explicitly} compute the subsequent timestep $\tilde{t}_{n-1}$:
\vspace{-2mm}
\begin{equation}
    \tilde{t}_{n-1} = t_n+\sum_{i=0}^{k-1} w_{n,i} \cdot (t_{n-1}-t_n).
    \label{time_shifting}
\end{equation}
Moreover, to compensate for remaining temporal mismatch and the exposure bias \cite{exposure_alleviating,exposure_elucidating} introduced by aggregating historical gradients, we adopt a learnable, per-step \emph{time scaling} scalar $s_n$. This scalar is applied directly to the time $t$ fed into the diffusion network: $\bm{\epsilon}_\theta(\mathbf{x}_{t_n}, s_n t_n)$. This two-part time calibration mechanism allows the solver to learn \textit{when} to query the model (\ie, at a slightly different noise level) to obtain the most accurate gradient for few-step sampling. The complete set of learnable parameters for an $N$-step \ours~is thus $\Phi = \{\mathbf{W}, \mathbf{s}\}$, where $\mathbf{W} \in \mathbb{R}^{N \times k}$ and $\mathbf{s} \in \mathbb{R}^N$.



 


    
    
    

\begin{algorithm}[t]
\caption{\texttt{DyWeight} Sampling}
\label{alg:solver}
\begin{algorithmic}[1]
\STATE \textbf{Given:}\\

\hspace{1em} Network $\mathcal{D}_\theta$, learned $\textcolor{paramcolor}{\Phi} = \{\textcolor{paramcolor}{\mathbf{W}}^{N \times K}, \textcolor{paramcolor}{\mathbf{s}}^{N \times 1}\}$,\\

\hspace{1em} Time schedule $\mathcal{T}_N$, order $K$, condition $\mathbf{c}$
 
\STATE \textbf{Initialize:}\\ 
\hspace{1em} $\mathbf{x}_{N}\!\leftarrow\!\mathbf{x}_T$, $t_{N}\!\leftarrow\!T$, buffer $\mathcal{B}\!\leftarrow\!\text{FixedQueue}(K)$

\FOR{$n=N$ to $1$}
    \vspace{1ex}
    \STATE $\mathcal{B}\xleftarrow{\text{buffer}}[\mathbf{x}_n-\mathcal{D}_\theta(\mathbf{x}_{n},\textcolor{paramcolor}{s_n}t_n,\mathbf{c})]/t_n$ 
    \hfill $\triangleright~\textcolor{commentcolor}{\text{Time scaling}}$
    \vspace{0.5ex}
    
    \STATE $\textcolor{paramcolor}{\mathbf{w}_n}^{p \times 1}\!\leftarrow\!\texttt{Get\_available\_weights}(\textcolor{paramcolor}{\mathbf{W}},K,n)$
    \vspace{0.5ex}
    
    \STATE $d_n \leftarrow \sum^{p-1}_{i=0}  \textcolor{paramcolor}{w_{n,i}} \cdot \mathcal{B}[i]$
    \hfill $\triangleright~\textcolor{commentcolor}{\text{Gradients  weighting}}$
    \vspace{0.5ex}
    
    \STATE $\mathbf{x}_{n-1} \leftarrow \mathbf{x}_{n}+ d_n \cdot (t_{n-1}-t_n)$
    \vspace{0.5ex}
    
    \STATE $t_{n-1}\!\leftarrow\!t_n\!+\!\sum^{p-1}_{i=0}\textcolor{paramcolor}{w_{n,i}}\cdot(t_{n-1}\!-\!t_n)$ 
     \!$\triangleright~\textcolor{commentcolor}{\text{Time shifting}}$
\ENDFOR
\STATE \textbf{Return:} $\mathbf{x}_0$
\end{algorithmic}
\end{algorithm}

\vspace{0.5em}
\noindent
\textbf{Optimization.} We optimize the solver parameters $\Phi$ via a teacher-student distillation framework \cite{AMED,LD3,EPD,DLMS,DSS}, as detailed in \cref{alg:training}. The parameters shown in \textcolor{paramcolor}{magenta} represent the learnable parameters. We first pre-compute a high-fidelity teacher trajectory $\mathbf{x}_0^{\text{tea}}$, by running a high-step ($M$ steps) solver (\eg, 35-step iPNDM \cite{iPNDM,PNDM}) from an initial noise sample $\mathbf{x}_T$. The student solver $\texttt{DyWeight}(\Phi)$ then generates its own solution $\mathbf{x}_0^{\text{stu}}$ from the \textit{same} $\mathbf{x}_T$ but with only $N \ll M$ steps. The parameters $\Phi$ are optimized by minimizing the distance between the final generated samples:
\begin{equation}
    \mathcal{L}(\Phi) = \mathbb{E}_{\mathbf{x}_T \sim \mathcal{N}(0, \sigma_T^2 \mathbf{I})} \left[ \text{dist}(\mathbf{x}_0^{\text{stu}}, \mathbf{x}_0^{\text{tea}}) \right].
    \label{eq:loss}
\end{equation}
The distance function $\text{dist}(\cdot, \cdot)$ is adapted to the data modality. For pixel-space datasets \cite{cifar,ffhq,afhqv2,imagenet} and LSUN-Bedroom \cite{lsun}, we measure the squared $\ell_2$ distance within the feature space of the final layer of a pre-trained Inception network \cite{inception}. For other latent-space models \cite{sd1.5,flux,flux_paper}, we compute the $\ell_2$ distance directly in the latent space. This result-supervised \cite{LD3} approach, which only supervises the terminal output, provides better flexibility than path-supervised methods \cite{DLMS,DSS,AMED,EPD}. We consider this is crucial for few-step solvers, as it permits the student to discover a more efficient, non-teacher-like trajectory that may lead to a better final result \cite{LD3}. The complete sampling and training processes are presented in \cref{alg:solver} and \cref{alg:training}, respectively.

\makeatletter
\renewcommand\thesubtable{(\alph{subtable})}
\makeatother

\begin{table*}[t!]
\caption{
Image generation FID$\downarrow$ \cite{fid} results across six datasets, \textbf{pixel space}:
{(a)} CIFAR-10 ($32 \times 32$) \cite{cifar}, 
{(b)} FFHQ ($64 \times 64$) \cite{ffhq},
{(c)} AFHQv2 ($64 \times 64$) \cite{afhqv2},
{(d)} ImageNet ($64 \times 64$) \cite{imagenet}, 
and \textbf{latent space}:
{(e)} LSUN-Bedroom ($256 \times 256$) \cite{lsun},
{(f)} MS-COCO ($512 \times 512$) \cite{mscoco} with Stable-Diffusion-v1.5 (SD-v1.5) \cite{sd1.5}.
We compared our $\ours$~with (1) Training-free multi-step solvers: iPNDM \cite{iPNDM,PNDM}, DPM-Solver++ \cite{dpmpp}, and UniPC \cite{UniPC}, (2) Learnable single-step and multi-step solvers: AMED-Solver and AMED-Plugin \cite{AMED}, EPD-Solver and EPD-Plugin \cite{EPD}, LD3 \cite{LD3} (we report the best FID results achieved by LD3 across different multi-step solvers) and DLMS \cite{DLMS}. The best results are in \textbf{bold}, the second best are \underline{underlined}. See Appendix \ref{sec:supp_quant} for complete results. 
}
\label{tab:main_results}
\vspace{-1.5mm}
{\small 
\captionsetup[subfloat]{labelformat=simple, labelsep=space}
\begin{minipage}[t]{0.48\textwidth}
\fontsize{8}{10}\selectfont
\subfloat[Unconditional generation on \textbf{CIFAR-10} $32 \times 32$ \cite{cifar}]{
\begin{tabular}{llcccc}
\toprule
&\multirow{2}{*}{Method} & \multicolumn{4}{c}{NFE} \\
\cmidrule{3-6} & & 3 & 5 & 7 & 9 \\
\midrule
\multirow{3}{*}{\rotatebox{90}{\fontsize{6.0pt}{7.3pt}\selectfont Handcrafted}} 
&iPNDM \cite{iPNDM,PNDM} & 24.55 & 7.77 & 4.04 & 2.83 \\
&DPM-Solver++(3M) \cite{dpmpp} & 55.76 & 9.94 & 4.29 & 2.99 \\
&UniPC \cite{UniPC} & 109.60 & 23.98 & 5.83 & 3.21 \\
\midrule
\multirow{7}{*}{\rotatebox{90}{\fontsize{6.3pt}{7.3pt}\selectfont Learnable}}
&AMED-Solver \cite{AMED} & 18.49 & 7.59 & 4.36 & 3.67 \\ 
&AMED-Plugin \cite{AMED} & 10.81 & 6.61 & 3.65 & 2.63 \\ 
&EPD-Solver \cite{EPD} & \underline{10.40} & 4.33 & 2.82 & 2.49 \\
&EPD-Plugin \cite{EPD} & 10.54 & 4.47 & 3.27 & 2.42 \\
&LD3 \cite{LD3} & 16.52 & 6.39 & 2.98 & 2.51 \\
&DLMS \cite{DLMS} & - & \underline{3.23} & 2.53 & 2.37 \\ 
&S4S-Alt \cite{s4s} & 16.95 & 3.73 & \underline{2.52} & \underline{2.31} \\ 
&\cellcolor{lightCyan}$\ours$~(ours) & \cellcolor{lightCyan}\textbf{8.16}  & \cellcolor{lightCyan}\textbf{3.02}  & \cellcolor{lightCyan}\textbf{2.40} & \cellcolor{lightCyan}\textbf{2.13}\\
\bottomrule
\end{tabular}
}
\vspace{0.5em} 
\subfloat[Unconditional generation on \textbf{FFHQ} $64 \times 64$ \cite{ffhq}]{
\begin{tabular}{llcccc}
\toprule
 &\multirow{2}{*}{Method} & \multicolumn{4}{c}{NFE} \\
\cmidrule{3-6} & & 3 & 5 & 7 & 9 \\
\midrule
\multirow{3}{*}{\rotatebox{90}{\fontsize{6.0pt}{7.3pt}\selectfont Handcrafted}} & iPNDM \cite{iPNDM,PNDM} & 27.72 & 13.80 & 7.16 & 4.98 \\
&DPM-Solver++(3M) \cite{dpmpp} & 66.07 & 13.47 & 6.20 & 4.77 \\
&UniPC \cite{UniPC} & 86.43 & 21.40 & 7.44 & 4.47 \\
\midrule
\multirow{7}{*}{\rotatebox{90}{\fontsize{6.0pt}{7.3pt}\selectfont Learnable}} 
&AMED-Solver \cite{AMED} & 47.31 & 14.80 & 8.82 & 6.31 \\ 
&AMED-Plugin \cite{AMED} & 26.87 & 12.49 & 6.64 & 4.24 \\ 
&EPD-Solver \cite{EPD} & 21.74 & 7.84 & 4.81 & 3.82 \\
&EPD-Plugin \cite{EPD} & \underline{19.02} & 7.97 & 5.09 & 3.53 \\
&LD3 \cite{LD3} & 23.86 & 10.36 & 4.38 & \underline{2.94} \\
&DLMS \cite{DLMS} & - & 6.85 & 5.16 & 4.23 \\ 
&S4S-Alt \cite{s4s} & 19.86 & \underline{6.25} & \underline{3.45} & 3.00 \\ 
&\cellcolor{lightCyan}$\ours$~(ours) & \cellcolor{lightCyan}\textbf{16.78}  & \cellcolor{lightCyan}\textbf{5.85}  & \cellcolor{lightCyan}\textbf{3.39} & \cellcolor{lightCyan}\textbf{2.77}\\
\bottomrule
\end{tabular}
}
\vspace{0.5em}

\subfloat[Unconditional generation on \textbf{AFHQ} $64 \times 64$ \cite{afhqv2}]{
\begin{tabular}{llcccc}
\toprule
 &\multirow{2}{*}{Method} & \multicolumn{4}{c}{NFE} \\
\cmidrule{3-6} & & 3 & 5 & 7 & 9 \\
\midrule
\multirow{3}{*}{\rotatebox{90}{\fontsize{6.0pt}{7.3pt}\selectfont Handcrafted}} & iPNDM \cite{iPNDM,PNDM} & 15.53 & 5.58 & 3.19 & 2.48 \\
&DPM-Solver++(3M) \cite{dpmpp} & 35.05 & 10.63 & 4.47 & 3.44 \\
&UniPC \cite{UniPC} & 60.89 & 13.01 & 5.07 & 4.46 \\
\midrule
\multirow{3}{*}{\rotatebox{90}{\fontsize{6.3pt}{7.3pt}\selectfont Learnable}} 
&LD3 \cite{LD3} & 17.94 & 6.09 & 2.97 & 2.27 \\
&S4S-Alt \cite{s4s} & \underline{14.71} & \underline{3.89} & \underline{2.56} & \underline{2.18} \\ 

&\cellcolor{lightCyan}$\ours$~(ours) & \cellcolor{lightCyan}\textbf{9.16}  & \cellcolor{lightCyan}\textbf{3.20} & \cellcolor{lightCyan}\textbf{2.42} & \cellcolor{lightCyan}\textbf{2.13}\\
\bottomrule
\end{tabular}
}
\hspace{-3ex}

\end{minipage}
\hfill
\begin{minipage}[t]{0.48\textwidth}
\centering
  \fontsize{8}{10}\selectfont

\subfloat[Conditional generation on \textbf{ImageNet} $64 \times 64$ \cite{imagenet}]{
\begin{tabular}{llcccc}
\toprule
 &\multirow{2}{*}{Method} & \multicolumn{4}{c}{NFE} \\
\cmidrule{3-6} & & 3 & 5 & 7 & 9 \\
\midrule
\multirow{3}{*}{\rotatebox{90}{\fontsize{6.0pt}{7.3pt}\selectfont Handcrafted}} & iPNDM \cite{iPNDM,PNDM} & 34.81 & 15.54 & 8.64 & 5.64 \\
&DPM-Solver++(3M) \cite{dpmpp} & 65.19 & 16.87 & 8.68 & 6.25 \\
&UniPC \cite{UniPC} & 91.38 & 24.36 & 9.57 & 6.34 \\
\midrule
\multirow{7}{*}{\rotatebox{90}{\fontsize{6.0pt}{7.3pt}\selectfont Learnable}} 
&AMED-Solver \cite{AMED} & 38.10 & 10.74 & 6.66 & 5.44 \\ 
&AMED-Plugin \cite{AMED} & 28.06 & 13.83 & 7.81 & 5.60 \\ 
&EPD-Solver \cite{EPD} & \underline{18.28} & \underline{6.35} & 5.26 & 4.27 \\
&EPD-Plugin \cite{EPD} & 19.89 & 8.17 & \underline{4.81} & \underline{4.02} \\
&LD3 \cite{LD3} & 27.82 & 11.55 & 5.63 & 4.71 \\
&DLMS \cite{DLMS} & - & 7.16 & 6.31 & 4.57 \\ 

&\cellcolor{lightCyan}$\ours$~(ours) & \cellcolor{lightCyan}\textbf{17.37}  & \cellcolor{lightCyan} \textbf{6.30}  & \cellcolor{lightCyan}\textbf{4.55} & \cellcolor{lightCyan}\textbf{3.82}\\
\bottomrule
\end{tabular}
}
\vspace{1.68em}

\subfloat[Unconditional generation on \textbf{LSUN Bedroom} $256 \times 256$ \cite{lsun}]{
\begin{tabular}{llcccc}
\toprule
 &\multirow{2}{*}{Method} & \multicolumn{4}{c}{NFE} \\
\cmidrule{3-6} & & 3 & 5 & 7 & 9 \\
\midrule
\multirow{3}{*}{\rotatebox{90}{\fontsize{6.0pt}{7.3pt}\selectfont Handcrafted}} & iPNDM \cite{iPNDM,PNDM} & 43.31 & 18.46 & 11.03 & 7.89 \\
&DPM-Solver++(3M) \cite{dpmpp} & 111.90 & 18.44 & 5.18 & 3.77 \\
&UniPC \cite{UniPC} & 112.30 & 13.76 & 4.52 & 3.72 \\
\midrule
\multirow{7}{*}{\rotatebox{90}{\fontsize{6.0pt}{7.3pt}\selectfont Learnable}} 
&AMED-Solver \cite{AMED} & 58.21 & 13.20 & 7.10 & 5.65 \\ 
&AMED-Plugin \cite{AMED} & 101.5 & 25.68 & 8.63 & 7.82 \\ 
&EPD-Solver \cite{EPD} & \underline{13.21} & 7.52 & 5.97 & 5.01 \\
&EPD-Plugin \cite{EPD} & 14.12 & 8.26 & 5.24 & 4.51 \\
&LD3 \cite{LD3} & 14.62 & 5.93 & 4.16 & 3.98 \\
&DLMS \cite{DLMS} & - & \underline{5.44}  & \underline{3.99}  & \underline{3.70} \\ 
&S4S-Alt \cite{s4s} & 37.65 & 13.03 & 10.03 & - \\ 
&\cellcolor{lightCyan}$\ours$~(ours) & \cellcolor{lightCyan}\textbf{9.82}  & \cellcolor{lightCyan}\textbf{4.97}  & \cellcolor{lightCyan}\textbf{3.87} & \cellcolor{lightCyan}\textbf{3.45}\\
\bottomrule
\end{tabular}
\label{tab:lsun_fid}
}

\vspace{0.5em}
\subfloat[Conditional generation on \textbf{MS-COCO} $512 \times 512$ \cite{mscoco}]{
\begin{tabular}{llcccc}
\toprule
 &\multirow{2}{*}{Method} & \multicolumn{4}{c}{NFE(1 step = 2 NFE)} \\
\cmidrule{3-6} & & 8 & 12 & 16 & 20 \\
\midrule
\multirow{3}{*}{\rotatebox{90}{\fontsize{6.0pt}{7.3pt}\selectfont Handcrafted}}
&iPNDM(2M) \cite{iPNDM,PNDM}& 19.65 & 14.38 & 14.03 & 13.75 \\
&DPM-Solver++(2M) \cite{dpmpp} & 21.33 & 15.15 & \underline{12.33} & \underline{12.04} \\
&UniPC(2M) \cite{UniPC} & 32.01 & 15.81 & 14.44 & 14.36 \\
\midrule
\multirow{3}{*}{\rotatebox{90}{\fontsize{6.0pt}{7.3pt}\selectfont Learnable}} 
&AMED-Plugin \cite{AMED} & 18.92 & 14.84 & 13.96 & 13.24 \\ 
&EPD-Plugin \cite{EPD} & \underline{16.46} & \underline{13.14} & 12.52 & 12.17 \\
&\cellcolor{lightCyan}$\ours$~(ours) & \cellcolor{lightCyan}\textbf{14.92} & \cellcolor{lightCyan}\textbf{11.82} & \cellcolor{lightCyan}\textbf{11.75} & \cellcolor{lightCyan}\textbf{11.54} \\
\bottomrule
\end{tabular}
 \label{tab:sd_fid}
}
\end{minipage}}
\end{table*}
\section{Experiments}
\subsection{Setup}

\noindent\textbf{Pretrained models.} We comprehensively evaluate our method across diverse benchmarks. For pixel-space generation, we use the official EDM \cite{EDM} checkpoints on CIFAR-10 (32$\times$32) \cite{cifar}, ImageNet (64$\times$64) \cite{imagenet}, FFHQ (64$\times$64) \cite{ffhq}, and AFHQv2 (64$\times$64) \cite{afhqv2}. 
For latent-space generation, we use Stable Diffusion v1.5 \cite{sd1.5} on LSUN-Bedroom \cite{lsun} and MS-COCO \cite{mscoco}, as well as the recent state-of-the-art text to image generation model FLUX.1-dev (Flux) \cite{flux,flux_paper}.

\noindent
\textbf{Baseline methods.}  We compare \ours~against two categories of diffusion solvers: (1) \textit{Handcrafted} multi-step solvers: iPNDM \cite{PNDM,iPNDM}, DPM-Solver++ \cite{dpmpp}, and UniPC \cite{UniPC}. We follow recommended time schedules: logSNR for DPM-Solver++ and UniPC \cite{dpmpp,UniPC}, and polynomial ($\rho=7$) \cite{EDM} for iPNDM \cite{PNDM,iPNDM}. (2) \textit{Learnable} solvers: AMED-Solver/Plugin \cite{AMED}, EPD-Solver/Plugin \cite{EPD}, LD3 \cite{LD3}, DLMS \cite{DLMS} and S4S-Alt \cite{s4s}, with results from their original papers. For LD3, we report the best FID across its base solvers; for DLMS, 3-NFE is omitted as not reported. Our method and learnable baselines use the Analytical First Step (AFS) \cite{afs} on pixel-space datasets so handcrafted solvers are also reported with AFS for fair comparison. On latent-space, AFS is used on LSUN-Bedroom \cite{lsun} but not on SD-v1.5 \cite{sd1.5} or Flux \cite{flux}.

\noindent
\textbf{Training details.} We optimize \ours~solver parameters for each NFE using a teacher–student distillation framework (\cref{alg:training}). A 35-NFE iPNDM \cite{PNDM,iPNDM} serves as the teacher to generate 10k teacher-student pairs. Parameters are optimized with Adam \cite{opt_adam} and a cosine-annealing LR, minimizing the end-point distance in \eqref{eq:loss}. Training is efficient: for a 35-NFE teacher, \ours~takes $\sim$3 minutes on CIFAR-10 \cite{cifar} (2$\times$ RTX 4090), $\sim$5 minutes on ImageNet-64 \cite{imagenet} (8$\times$ RTX 4090) and $\sim$9 minutes on LSUN-Bedroom \cite{lsun} (8$\times$ H100), \textit{including the time required to generate teacher samples}. As only solver parameters are trained, \ours~introduces no inference latency.

\noindent
\textbf{Evaluation metrics.} For pixel space datasets \cite{cifar,ffhq,afhqv2,imagenet} and LSUN-Bedroom \cite{lsun}, we report quantitative results using the Fréchet Inception Distance (FID) \cite{fid} computed on 50k generated samples. For text-to-image models\cite{sd1.5,flux}, we report both FID \cite{fid} and CLIP score \cite{clip} on 30K samples \cite{fid_sd1.5} from the MS-COCO \cite{mscoco} validation split to assess quality and prompt-image alignment.

\subsection{Main results}
\begin{table}[t]
    \caption{Performance comparison on FLUX.1-dev~\cite{flux} for text-to-image generation, evaluated on MS-COCO (512$\times$512)~\cite{mscoco}. \ours~consistently outperforms competitive solvers in both FID~\cite{fid} and CLIP score~\cite{clip}.}
    \label{tab:flux_main_comparison}
    \centering
      \fontsize{8}{10}\selectfont
      \setlength{\tabcolsep}{4pt} 
    \begin{tabular}{clccc}
    \toprule
    Metrics & Method & NFE=3 & NFE=5 & NFE=7 \\
    \midrule
    \multirow{3}{*}{\parbox[c]{1.5cm}{\centering FID$\downarrow$}} 
    & DPM-Solver++(2M) \cite{dpmpp} & 63.64 & 21.81 & 22.18 \\
    & iPNDM(2M) \cite{PNDM, iPNDM}     & 48.46 & 25.03 &  22.22 \\
    & \cellcolor{lightCyan}$\ours$(ours) & \cellcolor{lightCyan}\textbf{42.69} & \cellcolor{lightCyan}\textbf{21.02} & \cellcolor{lightCyan}\textbf{20.69} \\
    \midrule
    \multirow{3}{*}{\parbox[c]{1.5cm}{\centering CLIP$\uparrow$}} 
    & DPM-Solver++(2M) \cite{dpmpp} & 24.40 & 26.38 & 26.20 \\
    & iPNDM(2M) \cite{PNDM, iPNDM}     & 24.83 & 26.37 &  26.17 \\
    & \cellcolor{lightCyan}$\ours$(ours) & \cellcolor{lightCyan}\textbf{25.43} & \cellcolor{lightCyan}\textbf{26.73} & \cellcolor{lightCyan}\textbf{26.49} \\
    \bottomrule
    \end{tabular}
\end{table}

\begin{table}[t]
    \caption{Performance comparison on FLUX.1-dev~\cite{flux} for text-to-image generation, evaluated on Drawbench~\cite{drawbench}. \ours~consistently outperforms competitive solvers in CLIP score~\cite{clip}, HPSv2.1~\cite{hpsv2} and ImageReward~\cite{imagereward}.}
    \label{tab:flux_drawbench_comparison}
    \centering
      \fontsize{8}{10}\selectfont
      \setlength{\tabcolsep}{4pt} 
    \begin{tabular}{clccc}
    \toprule
    Metrics & Method & NFE=5 & NFE=7 & NFE=9 \\
    \midrule
    \multirow{3}{*}{\parbox[c]{1.5cm}{\centering CLIP$\uparrow$}} 
    & DPM-Solver++(2M) \cite{dpmpp} & 26.95 & 27.35 & 27.36 \\
    & iPNDM(2M) \cite{PNDM, iPNDM}     & 26.95 & 27.10 &  27.14 \\
    & \cellcolor{lightCyan}$\ours$(ours) & \cellcolor{lightCyan}\textbf{27.48} & \cellcolor{lightCyan}\textbf{27.75} & \cellcolor{lightCyan}\textbf{27.63} \\
    \midrule
    \multirow{3}{*}{\parbox[c]{1.5cm}{\centering HPSv2.1$\uparrow$}} 
    & DPM-Solver++(2M) \cite{dpmpp} & 0.271 & 0.286 & 0.291 \\
    & iPNDM(2M) \cite{PNDM, iPNDM}     & 0.273 & 0.286 & 0.291 \\
    & \cellcolor{lightCyan}$\ours$(ours) & \cellcolor{lightCyan}\textbf{0.279} & \cellcolor{lightCyan}\textbf{0.288} & \cellcolor{lightCyan}\textbf{0.293} \\
    \midrule
    \multirow{3}{*}{\parbox[c]{1.5cm}{\centering ImageReward$\uparrow$}} 
    & DPM-Solver++(2M) \cite{dpmpp} & 0.637 & 0.863 & 0.916 \\
    & iPNDM(2M) \cite{PNDM, iPNDM}     & 0.664 & 0.852 &  0.909 \\
    & \cellcolor{lightCyan}$\ours$(ours) & \cellcolor{lightCyan}\textbf{0.856} & \cellcolor{lightCyan}\textbf{0.921} & \cellcolor{lightCyan}\textbf{0.928} \\
    \bottomrule
    \end{tabular}
\end{table}
\begin{table}[t]
  \caption{CLIP-Score$\uparrow$ \cite{clip} results on Stable-Diffusion-v1.5~\cite{sd1.5}.}
  \label{tab:sd-clip-score}
  \centering
      \fontsize{8}{10}\selectfont
      \setlength{\tabcolsep}{7pt} 
    \begin{tabular}{lcccc}
      \toprule
      \multirow{2}{*}{Method} & \multicolumn{4}{c}{NFE (1 Step = 2 NFEs with CFG \cite{cfg})} \\
      \cmidrule{2-5}
      & 8 & 12 & 16 & 20 \\
      \midrule
      iPNDM(2M) \cite{iPNDM,PNDM}& 28.33 & \underline{29.79} & \underline{30.15} & \textbf{30.30} \\
      DPM-Solver++(2M) \cite{dpmpp} & 28.50 & 28.65 & 29.65 & 30.02 \\
      UniPC(2M) \cite{UniPC} & 26.44 & 29.05 & 29.79 & \underline{30.08} \\
      \rowcolor{lightCyan}
$\ours$ (ours) & \textbf{28.56} & \textbf{29.81} & \textbf{30.17} & 30.05 \\
      \bottomrule
    \end{tabular}
\end{table}
We present the primary quantitative comparisons in \cref{tab:main_results}. \ours~consistently establishes a new state-of-the-art for few-step sampling, achieving the best performance on all datasets and settings.

\begin{figure}[t]
\centering
    \includegraphics[width=\linewidth]{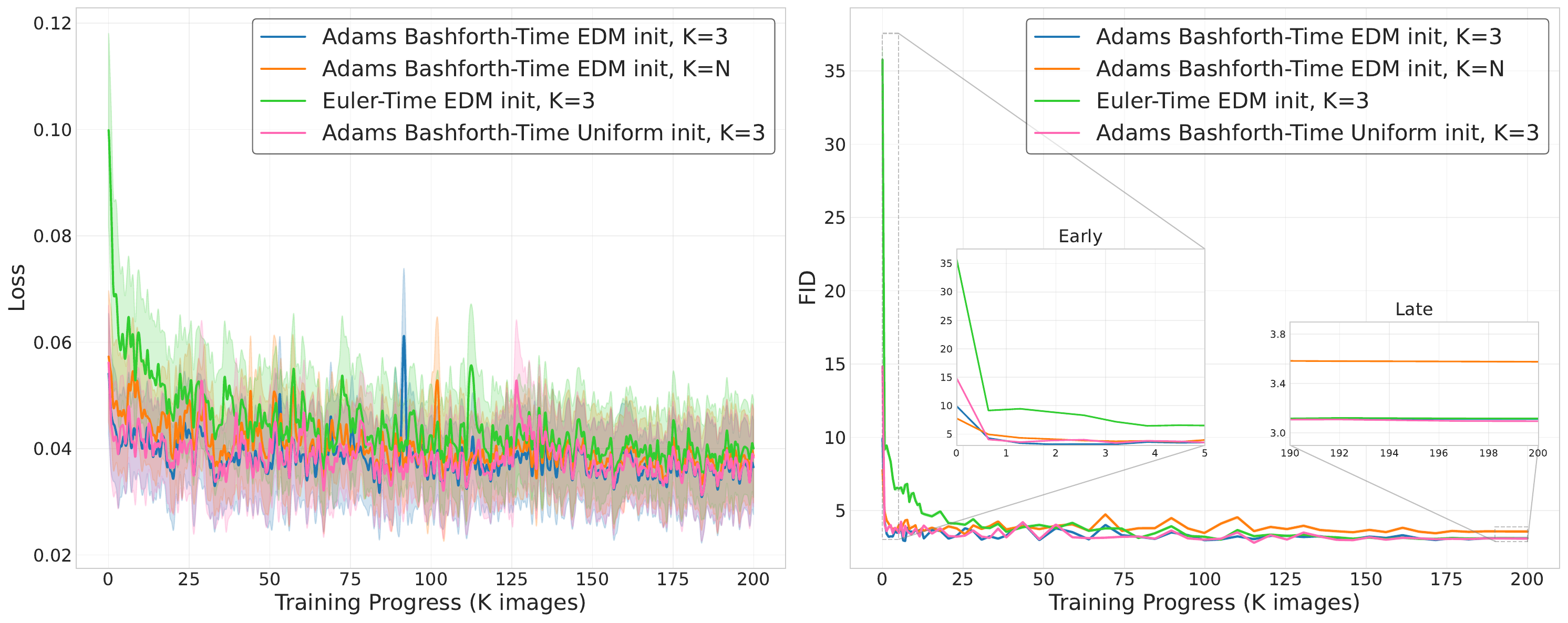}
    \vspace{-5mm}
    \caption{\textbf{Training convergence on CIFAR-10 (200K training images).} We track training loss (left) and generation FID (right) against optimization progress under various solver configurations. As shown, our framework exhibits extreme sample efficiency, achieving massive performance gains with merely $\sim \mathbf{600}$ images.}
    \label{fig:convergence}
    \vspace{-4mm}
\end{figure}

\noindent
\textbf{Pixel-space generation.} On CIFAR-10 \cite{cifar} at NFE=3, \ours~achieves an FID of 8.16, a 21.5\% improvement over the next-best learnable method EPD-Solver \cite{EPD} (10.40), and dramatically better than the widely-used handcrafted iPNDM \cite{PNDM,iPNDM} (24.55). This significant quality advantage holds for higher resolution generation: on AFHQv2 \cite{afhqv2} at NFE=3 (FID 9.16 vs. 14.71 for S4S-Alt \cite{s4s}) and FFHQ \cite{ffhq} at NFE=5 (FID 5.85 vs. 6.85 for DLMS \cite{DLMS}).

\noindent
\textbf{Latent-space generation.} \ours~also excels on latent diffusion models \cite{sd1.5,flux}. On LSUN-Bedroom \cite{lsun} at NFE=5, \ours~(4.97 FID) again outperforms all competitors. On SD-v1.5 \cite{sd1.5} (\cref{tab:sd_fid,tab:sd-clip-score}), while \ours~achieves superior FID scores, its CLIP score at 20 NFEs is slightly lower than baselines. This highlights a well-documented trade-off between perceptual quality (FID) and prompt alignment (CLIP), where optimizing for one can slightly degrade the other \cite{clipvsfid_1,clipvsfid_2}. However, on the state-of-the-art Flux model \cite{flux,flux_paper} (\cref{tab:flux_main_comparison}), \ours~achieves consistent improvement across both metrics, outperforming baselines in FID and CLIP scores at all tested NFEs. Moreover, to evaluate the robustness of our learned solver against prompt distribution shifts, we further test \ours~on the challenging DrawBench \cite{drawbench} prompt suite. As shown in \cref{tab:flux_drawbench_comparison}, \ours~generalizes effectively to these unseen, complex prompts, consistently outperforming baselines in human preference metrics such as HPSv2.1 \cite{hpsv2} and ImageReward \cite{imagereward}. This demonstrates that our framework captures generalized solver dynamics rather than simply overfitting to the training prompt distribution.

\noindent
\textbf{Qualitative results.} \cref{fig:teaser} provides a qualitative comparison on the Flux \cite{flux} benchmark at 7 NFEs. We compare \ours~(bottom row) against DPM-Solver++ \cite{dpmpp} (top row) and iPNDM \cite{PNDM,iPNDM} (middle row). Our method generates images with superior visual quality, structural coherence (\eg, the legible ``DyWeight'' text and the corgi's face), and consistency with complex prompts. See Appendix \ref{sec:supp_qualitative_results} for additional examples and results on other datasets.

\subsection{Ablation studies}
\begin{figure*}[ht]
\centering

\subfloat[FID results on CIFAR-10 \cite{cifar}.]{
    \includegraphics[width=0.27\linewidth]{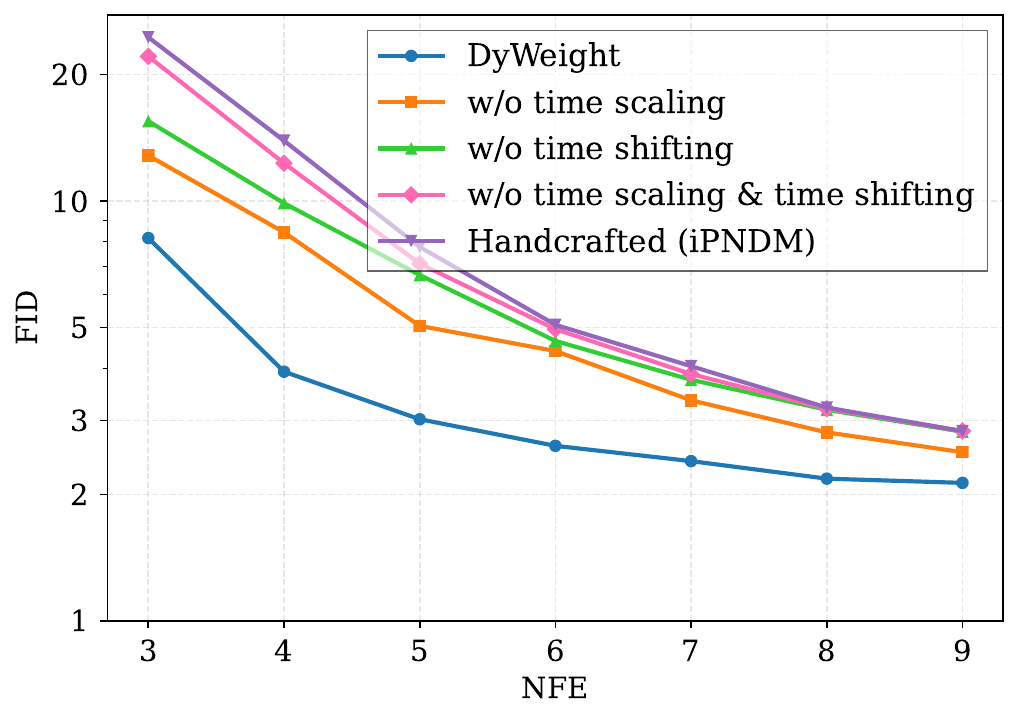}
    \label{fig:ab_cifar}
    } 
\hspace{1.5em}
\subfloat[FID results on FFHQ \cite{ffhq}.]{
    \includegraphics[width=0.27\linewidth]{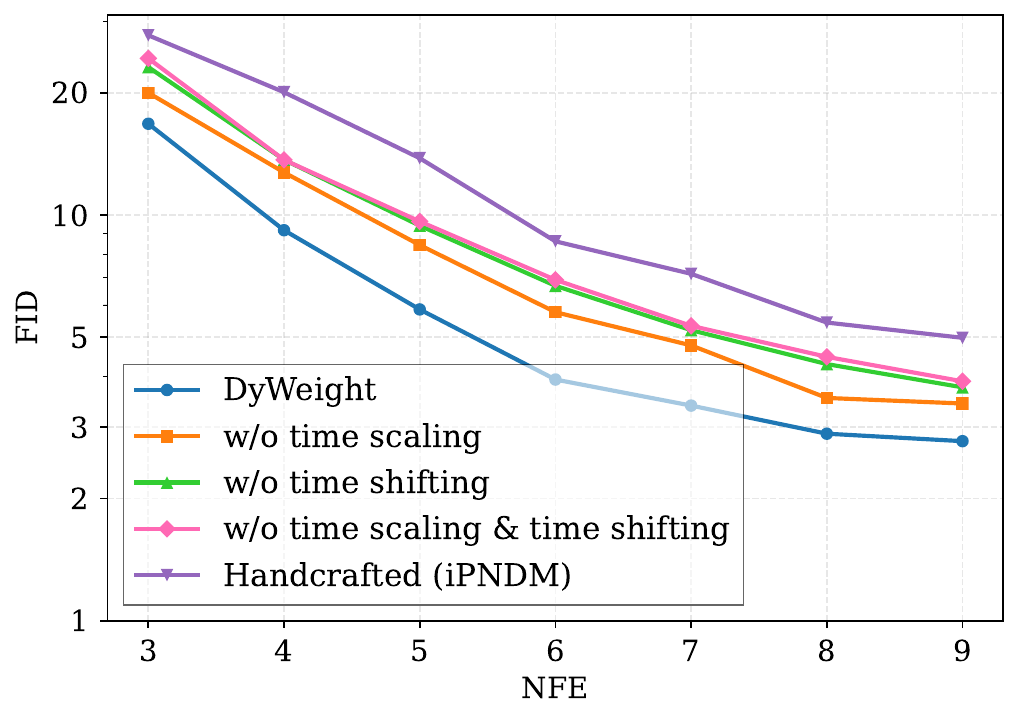}
    \label{fig:ab_ffhq}
    }
\hspace{1.5em}
\subfloat[FID results on CIFAR-10 \cite{cifar}.]{
    \includegraphics[width=0.27\linewidth]{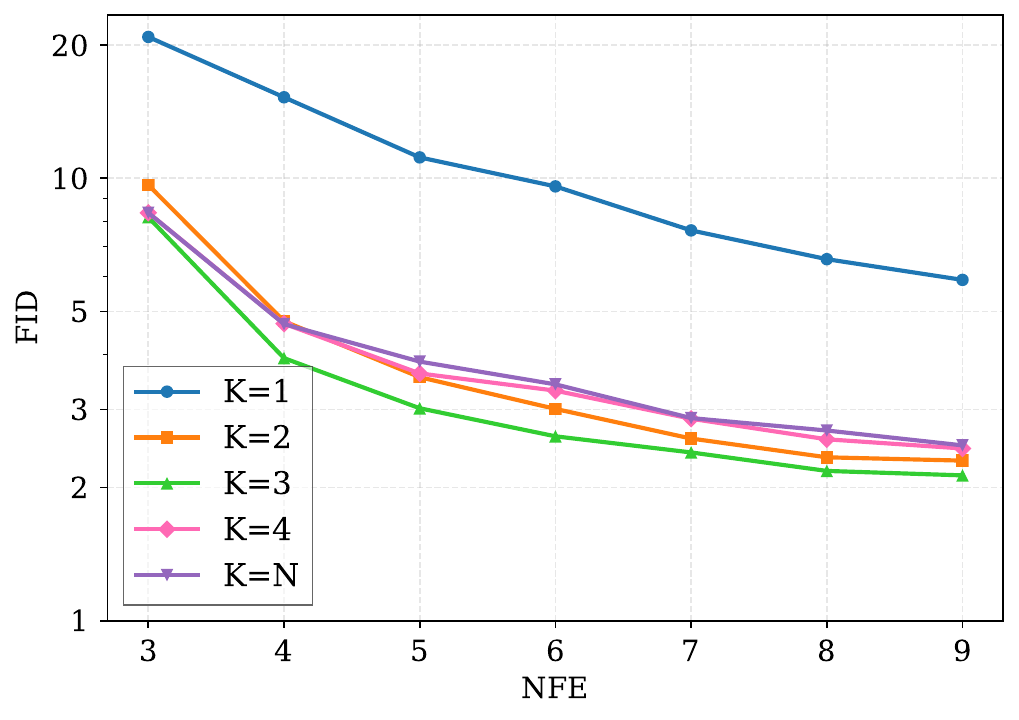}
    \label{fig:ab_window_cifar}
    }
    \vspace{-2mm}
    \caption{Ablation studies. (a, b) Impact of time calibration components on CIFAR-10 \cite{cifar} and FFHQ \cite{ffhq}. Both time shifting and time scaling are essential for optimal performance. (c) Sensitivity to multi-step order $K$ on CIFAR-10 \cite{cifar}, showing $K=3$ is optimal.}
    \vspace{-4mm}
\end{figure*}

\textbf{Effect of time calibration.} In \cref{fig:ab_cifar,fig:ab_ffhq}, we analyze our time calibration mechanism. Disabling time scaling or time shifting degrades performance, and disabling both causes a substantial drop. This confirms that adaptively learning \textit{both} the gradient weights and the temporal alignment is critical for achieving stable and accurate few-step sampling.

\noindent
\textbf{Sensitivity to solver order $K$.}~\cref{fig:ab_window_cifar} examines the sensitivity to the multi-step order $K$ (number of historical gradients). Performance improves from $K\!=\!1$ to $K\!=\!3$, with the best and most stable results at $K\!=\!3$. Notably, using available historical gradients ($K=N$) degrades performance. This suggests that a history of 3 or 4 steps provides an optimal balance between information gain and the inclusion of stale, less-relevant gradients. It also indicates that naively using all history, as explored in works like DSS \cite{DSS} which frame this as a parameter search problem, may be suboptimal.

\noindent
\textbf{Training convergence.} We analyze the optimization dynamics of \ours~in \cref{fig:convergence}. Proper initialization provides a strong inductive bias that significantly enhances optimization efficiency. Specifically, starting with higher-order classical coefficients (\eg, Adams-Bashforth \cite{adam}) yields a superior initial state, allowing both the training loss and generation FID to converge extremely rapidly, achieving substantial performance gains with merely $\sim$\textbf{600} training images. To systematically investigate this, \cref{tab:supp_init_comparison} in Appendix further details how different initialization strategies impact the final generative quality when optimized on 10K images. Furthermore, regarding the gradient buffer size (order), while utilizing all available history ($K=N$) eventually converges to a similar loss, it yields slightly worse final fidelity. This indicates that long-term coefficients, even when decayed to near-zero, still accumulate noise from stale gradients, validating our design choice to truncate historical gradients for optimal stability.

\noindent
\textbf{Robustness to teacher solver.}~\cref{tab:abl-teacher-solver} examines the sensitivity of \ours~to the teacher solver. The student achieves similar final performance regardless of teacher type or step setting, validating our end-point supervision and showing that \ours~learns an intrinsically efficient trajectory rather than simply mimicking a specific teacher.

\section{Conclusion}
\begin{table}[t]
  \caption{Robustness to teacher solver on CIFAR-10 \cite{cifar}. Numbers in parentheses denote teacher steps. Student performance is stable across different high-fidelity teacher solvers and step counts.}
  \label{tab:abl-teacher-solver}
  \centering
      \fontsize{8}{10}\selectfont
      \setlength{\tabcolsep}{8pt} 
    \begin{tabular}{lcccc}
      \toprule
      \multirow{2}{*}{Teacher Solver ($M$ steps)} & \multicolumn{4}{c}{NFE} \\
      \cmidrule{2-5}
      & 3 & 5 & 7 & 9 \\
      \midrule
      iPNDM (18) \cite{iPNDM,PNDM}& 9.06 & 3.33 & 2.47 & 2.28 \\
      iPNDM (35) \cite{iPNDM,PNDM}& \textbf{8.16} & \textbf{3.02} & 2.40 & 2.13 \\
      iPNDM (50) \cite{iPNDM,PNDM}& 8.70 & 3.23 & 2.39 & 2.29 \\
      DPM-Solver-2 (35) \cite{dpmsolver} & 8.77 & 3.17 & 2.44 & 2.24 \\
      Heun (35) \cite{EDM} & 8.80 & 3.09 & \textbf{2.28} & \textbf{2.12} \\
      \bottomrule
    \end{tabular}
    \vspace{-3mm}
\end{table}
We introduce \ours, a learning-based multi-step solver designed for efficient few-step diffusion sampling. Conventional solvers rely on static, handcrafted coefficients that are suboptimal for the highly non-stationary diffusion dynamics. To overcome this, \ours~employs unconstrained, time-varying parameters to implicitly couple gradient aggregation with dynamic step-size modulation. By relaxing the constraint that weights must sum to one, the solver adaptively re-weights historical gradients while simultaneously performing time calibration (shifting and scaling) to maintain temporal alignment with the model's internal dynamics. Trained via a lightweight teacher-student distillation framework, our method efficiently captures the evolving denoising behavior without the optimization bottlenecks of decoupled schedules, and without modifying the underlying diffusion model. Extensive experiments across diverse datasets demonstrate that \ours~achieves superior visual fidelity with minimal NFEs, consistently outperforming prior methods and establishing a new state-of-the-art in the low-NFE regime.

{
    \small
    \bibliographystyle{ieeenat_fullname}
    \bibliography{cite}
}


\appendix
\newcommand{\AppendixPrefix}{A}

\clearpage

\setcounter{page}{1}
\setcounter{figure}{0}
\setcounter{table}{0}
\setcounter{equation}{0}

\renewcommand{\thefigure}{A.\arabic{figure}}
\renewcommand{\thetable}{A.\arabic{table}}
\renewcommand{\theequation}{A.\arabic{equation}}

\startcontents[supps]
{
    \hypersetup{linkcolor=black}

    \titlecontents{lsection}[0pt]
        {\vspace{3pt}\bfseries} 
        {\thecontentslabel\hspace{1em}}
        {}
        {\hfill\thecontentspage}

    \titlecontents{lsubsection}[1.5em]
        {\normalfont}
        {\thecontentslabel\hspace{1em}}
        {} 
        {\titlerule*[0.5pc]{.}\thecontentspage}

    \printcontents[supps]{l}{1}{\section*{Appendix}}
}

\vspace{1em}
\section{Derivation of general multi-step solvers}
\label{sec:appendix_derivation}

In this section, we explicitly derive the coefficients for the generalized Adams-Bashforth solvers adapted for non-uniform time steps. We show that the classical coefficients \cite{numerical_analysis,numerical_ode} are special cases of our general formulation.

\subsection{Problem formulation}
The diffusion sampling update at step $n$ requires solving the integral:
\begin{equation}
   \mathbf{x}_{t_{n-1}} 
    =\mathbf{x}_{t_n} + \int_{t_n}^{t_{n-1}} \bm{\epsilon}_\theta(\mathbf{x}_t,t) \mathrm{d}t.
\end{equation}
Let $h_n = t_{n-1} - t_n$ be the current step size. We utilize historical gradients at time steps $\{t_{n}, t_{n+1}, \dots, t_{n+K-1}\}$. We define the time interval between the current time $t_n$ and historical time $t_{n+i}$ as $\tau_i = t_n - t_{n+i}$ (where $\tau_0=0$).
To simplify integration, we apply a change of variables $u = t - t_n$, mapping the integration interval $[t_n, t_{n-1}]$ to $[0, h_n]$.

\subsection{\texorpdfstring{Second-order general solver ($K=2$)}{Second-Order General Solver (K=2)}}
We approximate $\epsilon_\theta$ using a linear Lagrange polynomial constructed from points $\{t_n, t_{n+1}\}$. The basis polynomials are:
\begin{equation}
    L_0(u) = \frac{u + \tau_1}{\tau_1}, \quad L_1(u) = \frac{u}{-\tau_1}.
\end{equation}
Integrating these over $[0, h_n]$ yields the weights:
\begin{align}
    \beta_0^{(2)} &= \frac{1}{h_n} \int_0^{h_n} \frac{u + \tau_1}{\tau_1} \mathrm{d}u = 1 + \frac{h_n}{2\tau_1}, \\
    \beta_1^{(2)} &= \frac{1}{h_n} \int_0^{h_n} \frac{-u}{\tau_1} \mathrm{d}u = -\frac{h_n}{2\tau_1}.
\end{align}
The update rule is:
\begin{equation}
    x_{t_{n-1}} = x_{t_n} + h_n \left( \beta_0^{(2)} \epsilon_n + \beta_1^{(2)} \epsilon_{n+1} \right).
\end{equation}
After substitution, it matches the formulation of DPM-Solver++(2M) \cite{dpmpp}.

\noindent
\textbf{Verification.} For uniform steps, $h_n = \tau_1 = h$.
\begin{itemize}
    \item $\beta_0^{(2)} = 1 + 1/2 = 3/2$,
    \item $\beta_1^{(2)} = -1/2$.
\end{itemize}
These match the standard Adams-Bashforth (AB-2) coefficients.

\subsection{\texorpdfstring{Third-order general solver ($K=3$)}{Third-order general solver (K=3)}}
We use three points $\{t_n, t_{n+1}, t_{n+2}\}$ with intervals $\tau_1, \tau_2$. The quadratic Lagrange basis for the current step $\epsilon_n$ is:
\begin{equation}
\medmath{
    L_0(u) = \frac{(u+\tau_1)(u+\tau_2)}{\tau_1 \tau_2} = \frac{u^2 + (\tau_1+\tau_2)u + \tau_1\tau_2}{\tau_1\tau_2}.}
\end{equation}
Integrating $\int_0^{h_n} L_0(u) \mathrm{d}u$ yields $\beta_0^{(3)}$:
\begin{equation}
    \beta_0^{(3)} = 1 + \frac{h_n}{2} \frac{\tau_1 + \tau_2}{\tau_1 \tau_2} + \frac{h_n^2}{3 \tau_1 \tau_2}.
\end{equation}
For historical steps $i \in \{1, 2\}$, the basis polynomials are:
\begin{equation}
    L_i(u) = \frac{u(u+\tau_{3-i})}{\tau_i(\tau_i-\tau_{3-i})}.
\end{equation}
Integration yields:
\begin{equation}
    \beta_i^{(3)} = \frac{1}{\tau_i(\tau_i - \tau_{3-i})} \left( \frac{h_n^2}{3} + \frac{h_n \tau_{3-i}}{2} \right).
\end{equation}

\noindent
\textbf{Verification.} For uniform steps ($\tau_1=h, \tau_2=2h$):
\begin{equation}
\begin{aligned}
    \beta_0^{(3)} &= 1 \!+\! \frac{h(3h)}{2(2h^2)} \!+\! \frac{h^2}{3(2h^2)} = 1 \!+\! \frac{3}{4}\! +\! \frac{1}{6} = \frac{23}{12}, \\
    \beta_1^{(3)} &= \frac{h^2/3 + h(2h)/2}{h(-h)} = \frac{4h^2/3}{-h^2} = -\frac{16}{12}, \\
    \beta_2^{(3)} &= \frac{h^2/3 + h(h)/2}{2h(h)} = \frac{5h^2/6}{2h^2} = \frac{5}{12}.
\end{aligned}
\end{equation}
These exactly recover the AB-3 coefficients.

\subsection{\texorpdfstring{Fourth-order general solver ($K=4$)}{Fourth-order general solver (K=4)}}
For the fourth-order solver, we integrate the cubic Lagrange polynomial constructed from $\{t_n, t_{n+1}, t_{n+2}, t_{n+3}\}$. The general update rule is:
\begin{equation}
    x_{t_{n-1}} = x_{t_n} + h_n \sum_{i=0}^{3} \beta_i^{(4)} \epsilon_{n+i}.
\end{equation}
The explicit coefficients derived from integration are coefficient for current step ($\epsilon_n$):
\begin{equation}
    \medmath{
    \beta_0^{(4)} = 1 + \frac{h_n}{2} \!\sum_{j=1}^3 \frac{1}{\tau_j} + \frac{h_n^2}{3} \!\!\sum_{1 \le j < k \le 3} \! \frac{1}{\tau_j \tau_k} + \frac{h_n^3}{4 \prod_{j=1}^3 \tau_j},
    }
\end{equation}
and coefficients for historical steps ($\epsilon_{n+i}, i \in \{1,2,3\}$):
\begin{equation}
    \beta_i^{(4)} = \frac{\frac{h_n^3}{4} + \frac{h_n^2}{3} S_1^{(i)} + \frac{h_n}{2} S_2^{(i)}}{\prod_{j \neq i} (\tau_j - \tau_i)}.
\end{equation}
Where $S_1^{(i)} = \sum_{j \ne i} \tau_j$ and $S_2^{(i)} = \prod_{j \ne i,~j \ne 0} \tau_j$ represent the sum and product of the other time intervals, respectively.

\noindent
\textbf{Verification.} We verify $\beta_0^{(4)}$ under uniform steps where $\tau_1=h, \tau_2=2h, \tau_3=3h$:
\begin{align}
    & \text{Sum terms:}  \quad \sum \frac{1}{\tau} = \frac{1}{h}\left(1+\frac{1}{2}+\frac{1}{3}\right) = \frac{11}{6h}, \nonumber \\
    & \text{Pair terms:}  \quad \sum \frac{1}{\tau\tau} = \frac{1}{h^2}\left(\frac{1}{2}+\frac{1}{3}+\frac{1}{6}\right) = \frac{1}{h^2}, \nonumber \\
    & \text{Prod term:}  \quad \prod \tau = 6h^3. \nonumber
\end{align}
Substituting these back:
\begin{equation}
    \beta_0^{(4)} = 1 + \frac{h}{2}\left(\frac{11}{6h}\right) + \frac{h^2}{3}\left(\frac{1}{h^2}\right) + \frac{h^3}{4(6h^3)} = \frac{55}{24}.
\end{equation}
Similarly, calculating the historical weights yields $\beta_1^{(4)}=-\frac{59}{24}$, $\beta_2^{(4)}=\frac{37}{24}$, and $\beta_3^{(4)}=-\frac{9}{24}$. This confirms that our derivation generalizes the standard AB-4 method.

\vspace{0.5em}
\noindent
\textbf{Summary.} The derivation above explicitly demonstrates that the coefficients of classical solvers (\eg, iPNDM \cite{PNDM,iPNDM}) are predicated on the strict assumption of linear, uniform time schedules (where $\tau_j = j \cdot h$). However, practical diffusion sampling often utilizes non-linear schedules \cite{EDM,score_based} to maximize image quality, rendering these fixed coefficients mathematically inaccurate. Furthermore, even under the ideal condition of uniform steps, our numerical analysis in \cref{sec:numerical} reveals that low-order polynomial approximations are suboptimal when the denoising function exhibits high complexity, especially in few-step settings. As the function complexity increases and the number of steps decreases, the truncation error of these static solvers escalates significantly. This theoretical misalignment underscores the necessity of our proposed \ours: by optimizing the coefficients via learning, the solver can adaptively compensate for both the non-uniformity of few-step schedules and the complex, non-stationary dynamics of the diffusion process, yielding a more accurate sampling trajectory.

\begin{figure*}[t]
\centering
    \includegraphics[width=\linewidth]{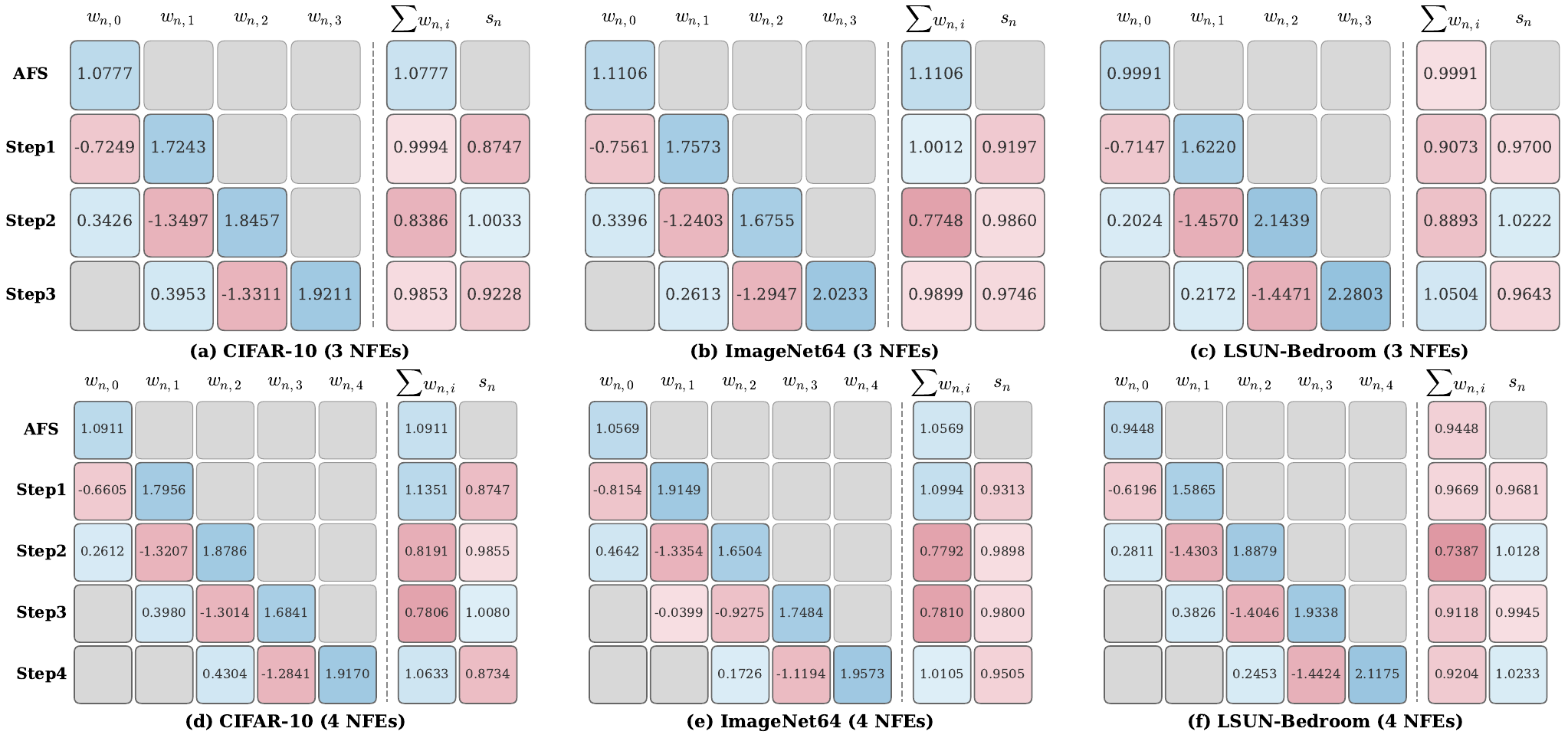}
    \caption{\textbf{Visualization of the learned coefficients across different datasets at 3 and 4 NFEs.} Each row represents the learned parameters for a specific denoising step, starting with the Analytical First Step (AFS) \cite{afs} in the first row. The columns explicitly detail the unconstrained historical gradient weights ($w_{n,i}$), followed by their unnormalized sum ($\sum w_{n,i}$) and the time scaling factor ($s_n$).}
    \label{fig_supp_weights}
\end{figure*}

\begin{table}[t]
\centering
\caption{\textbf{Cross-dataset evaluation of learned coefficients.} We evaluate the transferability of 3-NFE parameters trained on one dataset (columns, denoted by ``C.-'') across different evaluation distributions (rows). Native coefficients generally provide the most stable performance (diagonal).}
\label{tab:suppl_cross_dataset}
\footnotesize 
\setlength{\tabcolsep}{2.5pt} 
\begin{tabular}{l c c c c c}
\toprule
Dataset / Coeff. & C.-CIFAR & C.-ImgNet & C.-FFHQ & C.-LSUN & iPNDM \\
\midrule
CIFAR-10 \cite{cifar} & \textbf{8.16} & 14.04 & 16.41 & 14.86 & 24.55 \\
ImageNet \cite{imagenet} & 25.18 & \textbf{17.37} & 21.20 & 20.11 & 34.81 \\
FFHQ \cite{ffhq} & 33.89 & \textbf{16.64} & 16.78 & 21.49 & 27.72 \\
LSUN-Bed \cite{lsun} & 56.52 & 27.58 & 54.53 & \textbf{12.65} & 34.81 \\
\bottomrule
\end{tabular}
\end{table}

\section{Visualization and transferability of learned coefficients}
\textbf{Coefficients visualization.} To understand the behavior of \ours, we visualize the learned weights across different datasets and step settings in \cref{fig_supp_weights}. The unconstrained historical gradient weights ($w_{n,i}$) deviate significantly from standard handcrafted coefficients. This demonstrates that optimal solver configurations are highly dependent on the target data distribution and inference budget, proving static coefficients are sub-optimal. Furthermore, the unnormalized weight sum ($\sum w_{n,i}$) implicitly performs dynamic time shifting, while $s_n$ acts as a time scaling factor. Together, they elegantly couple gradient aggregation and schedule alignment.

\vspace{0.5em}
\noindent
\textbf{Cross-dataset transferability.} As demonstrated in \cref{tab:flux_drawbench_comparison}, the \ours~coefficients trained on MS-COCO \cite{mscoco} exhibit robust generalization when applied to the unseen and challenging DrawBench \cite{drawbench} prompt suite using Flux \cite{flux,flux_paper}. To delve deeper into this cross-domain transferability, we conduct comprehensive evaluations across various datasets, as detailed in \cref{tab:suppl_cross_dataset}. While native coefficients trained directly on the target distribution generally provide a highly optimal result, the transferred parameters reveal notable generalization patterns tied to dataset resolution and complexity. Specifically, coefficients distilled from high-resolution distributions (\eg, ImageNet \cite{imagenet}) exhibit exceptional transferability, not only showing strong performance on latent-space LSUN-Bedroom \cite{lsun} but also slightly outperforming native coefficients on FFHQ \cite{ffhq}. Conversely, transferring from lower-resolution to more complex ones is inherently more challenging. However, while severe domain shifts (\eg, from pixel-space CIFAR-10 \cite{cifar} to latent-space) inevitably cause expected performance drops, the swapped coefficients still frequently outperform the handcrafted iPNDM \cite{iPNDM,PNDM} baseline. Collectively, these findings validate that \ours~captures intrinsic and transferable solver dynamics.

\section{Impact of supervision objectives}
\label{sec:supp_supervision}

We rely on \textit{endpoint supervision} (supervising only the final generated sample $\mathbf{x}_0$) to train \ours. However, recent learning-based solvers such as AMED \cite{AMED}, DLMS \cite{DLMS}, EPD \cite{EPD} and DSS \cite{DSS} use \textit{path supervision}, forcing the student to match the teacher's trajectory at every intermediate step. In this section, we verify the superiority of our choice.

\vspace{0.5em}
\noindent
\textbf{Path supervision setup.} To implement path supervision on \ours, we require the high-fidelity teacher to provide ground-truth guidance for every intermediate step of the student. Since the student takes large strides (few steps), we execute the teacher solver for $M$ smaller sub-steps between each student interval $[t_n, t_{n-1}]$ to generate a precise target.
Formally, let the student's trajectory be $\{\mathbf{x}_{t_N}, \dots, \mathbf{x}_{t_0}\}$. For a student step from $t_n$ to $t_{n-1}$, the teacher generates a corresponding high-precision target $\mathbf{x}_{t_{n-1}}^\text{tea}$. The path supervision loss is defined as the cumulative error across all steps:
\begin{equation}
    \mathcal{L}_\text{path} = \sum_{n=1}^{N} \text{dist}(\mathbf{x}_{t_{n-1}}^\text{stu}, \mathbf{x}_{t_{n-1}}^\text{tea}).
    \label{eq:path_loss}
\end{equation}
Here, the metric $\text{dist}(\cdot)$ varies by step: for all intermediate steps ($n > 1$), it computes the $L_2$ loss to enforce trajectory alignment; for the final step ($n=1$), it employs the perceptual distance (\eg, Inception feature distance \cite{inception}) consistent with our main objective in \cref{eq:loss}.
In our experiments, we set the interval sub-steps $M=5$. Since the trajectory is generated sequentially, parameters at each step influence all subsequent states and are thus updated by back-propagating gradients accumulated from all following loss terms.

\vspace{0.5em}
\noindent
\textbf{Results and analysis.} As shown in \cref{tab:supp_supervision_comparison}, \textit{endpoint supervision} consistently outperforms \textit{path supervision} by a significant margin. On CIFAR-10 \cite{cifar} at 3 NFEs, \textit{endpoint supervision} achieves an FID of 8.16, whereas \textit{path supervision} degrades to 11.74. A similar trend is observed on ImageNet-64 \cite{imagenet}. We argue that the inferiority of path supervision in the few-step regime stems from \textit{over-constraint} and \textit{limited flexibility}.
\begin{itemize}
    \item Trajectory flexibility vs. imitation: \textit{Path supervision} forces the student to strictly ``imitate'' the teacher's high-precision trajectory. However, a few-step student (\eg, 3 steps) fundamentally lacks the capacity to physically follow the exact path of a high-step teacher, as the optimal trajectory for a large stride is often different from the accumulation of many small strides.
    \item Goal-oriented optimization: \textit{Endpoint supervision} relaxes these intermediate constraints. It treats the solver as a function approximator that only needs to map the initial noise $\mathbf{x}_T$ to the correct data $\mathbf{x}_0$. This grants the solver the freedom to explore ``shortcuts'' or alternative trajectories that may deviate from the teacher in the middle but land more accurately on the data manifold at the end.
\end{itemize}
This experiment not only justifies our design choice but also highlights a key distinction between \ours~and concurrent works like DLMS \cite{DLMS} and DSS \cite{DSS}: for extreme few-step sampling, \textit{where} you end up matters more than \textit{how} you get there.

\begin{table}[t]
  \caption{\textbf{Comparison of supervision strategies for training.} We compare trajectory-based (\textit{path}) supervision against terminal-state (\textit{endpoint}) supervision . \textit{Path} supervision computes the loss at every intermediate steps, forcing the student to strictly mimic the teacher's trajectory. In contrast, \textit{endpoint} supervision only supervises the final output, consistently yielding better results.}
  \label{tab:supp_supervision_comparison}
  \centering
  \fontsize{9}{10}\selectfont
  \setlength{\tabcolsep}{8pt} 
  \begin{tabular}{lcccc}
    \toprule
    \multirow{2}{*}{Supervision strategy} & \multicolumn{4}{c}{NFE} \\
    \cmidrule(lr){2-5}
    & 3 & 5 & 7 & 9 \\
    \midrule
    
    \rowcolor{gray!25} 
    \multicolumn{5}{l}{\textbf{CIFAR-10} \cite{cifar}} \\
    \midrule
    iPNDM \cite{PNDM,iPNDM}             & 24.55 & 7.77 & 4.04 & 2.83 \\
    \ours-\textit{path}     & 11.74 & 4.44 & 2.91 & 2.38 \\
    \ours-\textit{endpoint} & \textbf{8.16} & \textbf{3.02} & \textbf{2.40} & \textbf{2.13} \\
    
    \midrule
    
    \rowcolor{gray!25} 
    \multicolumn{5}{l}{\textbf{ImageNet-64} \cite{imagenet}} \\
    \midrule
    iPNDM \cite{PNDM,iPNDM}             & 34.81 & 15.54 & 8.64 & 5.64 \\
    \ours-\textit{path}     & 19.68 & 8.79 & 5.81 & 4.55 \\
    \ours-\textit{endpoint} & \textbf{17.37} & \textbf{6.30} & \textbf{4.55} & \textbf{3.82} \\
    
    \bottomrule
  \end{tabular}
\end{table}

\section{Experimental settings and implementation details}
\subsection{Controlled numerical study configuration}
To rigorously validate our hypothesis that handcrafted coefficients are suboptimal for high-complexity dynamics under few-step settings (\cref{sec:numerical}), we designed a ``white-box'' numerical testbed. This section details the construction of the synthetic problem, the solver parameterization, and the optimization procedure.

\vspace{0.5em}
\noindent
\textbf{Problem construction.} We construct a high-dimensional Ordinary Differential Equation (ODE) system $\frac{\mathrm{d}\mathbf{y}}{\mathrm{d}t} = \mathbf{f}(t, \mathbf{y})$ defined on the time interval $t \in [0, 1]$ with state dimension $D=50$. The ground-truth trajectory $\textbf{y}_\text{gt}(t)$ is defined as a polynomial of degree $K$:
\begin{equation}
    \textbf{y}_\text{gt}(t) = \mathbf{C} \cdot [1, t, t^2, \dots, t^K]^\top,
\end{equation}

where the coefficient matrix $\mathbf{C} \in \mathbb{R}^{D \times (K+1)}$ is sampled from a Gaussian distribution $\mathcal{N}(0, 4\mathbf{I})$. To mimic the complex feature interactions in the diffusion U-Net \cite{sd1.5,unet} and DiT \cite{flux,dit}, we introduce a stiff coupling term in the drift function:
\begin{equation}
    \mathbf{f}(t, \mathbf{y}) = \mathbf{y}'_\text{gt}(t) + \mathbf{A}\big[\mathbf{y}(t) - \mathbf{y}_\text{gt}(t)\big].
\end{equation}
Here, $\mathbf{A} \!\in \!\mathbb{R}^{D \times D}$ is a random coupling matrix initialized with entries sampled from $\mathcal{N}(0, 0.1/\sqrt{D})$ and shifted by $-0.5\mathbf{I}$ to ensure system stability. This formulation ensures that $\mathbf{y}_\text{gt}(t)$ is the exact solution while introducing cross-dimensional dependencies that challenge the numerical solver.

\vspace{0.5em}
\noindent
\textbf{Solver implementation.} We evaluate Adams-Bashforth (AB) \cite{numerical_analysis,numerical_ode,adam} solvers of orders $O \in \{1, 2, 3, 4\}$. 
\begin{itemize}
    \item \textbf{Standard solver}: Uses the classical, time-invariant coefficients derived from Lagrange polynomial integration on uniform grids \cite{numerical_analysis,numerical_ode}.
    \item \textbf{Optimized solver}: We parameterize the solver with time-varying learnable coefficients. For a total of $S$ steps, the solver learns a weight matrix $\mathbf{W} \in \mathbb{R}^{(S - O + 1) \times O}$. Unlike the standard solver which applies the same coefficients at every step, the optimized solver utilizes a specific set of weights for each integration step $n$, allowing it to find a more optimal path.
    \item \textbf{Initialization and start-up}: To ensure a fair comparison and isolate the solver's integration error from start-up inaccuracies, both standard and optimized solvers are initialized with the exact ground-truth values for the first $O$ history steps. Furthermore, the learnable weights of the optimized solver are initialized with the standard AB coefficients to provide a valid numerical baseline.
\end{itemize}

\begin{figure*}[t]
    \centering
    \begin{tabular}{cccc}
        \vspace{1mm} \hspace{-4mm} \subfloat[$S=6$]{\includegraphics[width=0.24\linewidth]{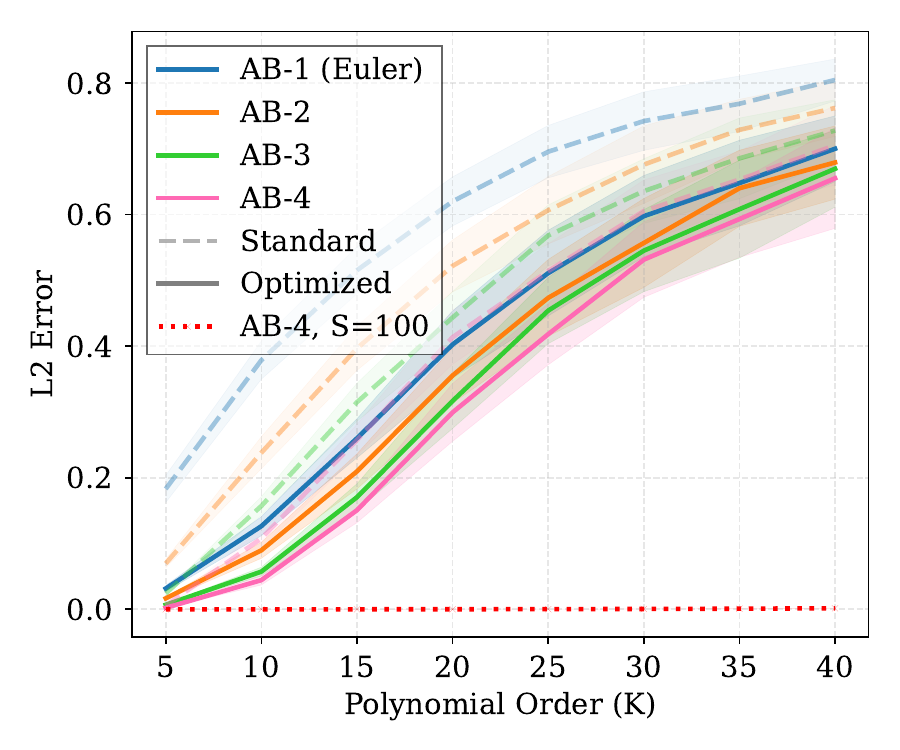}} & \hspace{-4mm}
        \subfloat[$S=8$]{\includegraphics[width=0.24\linewidth]{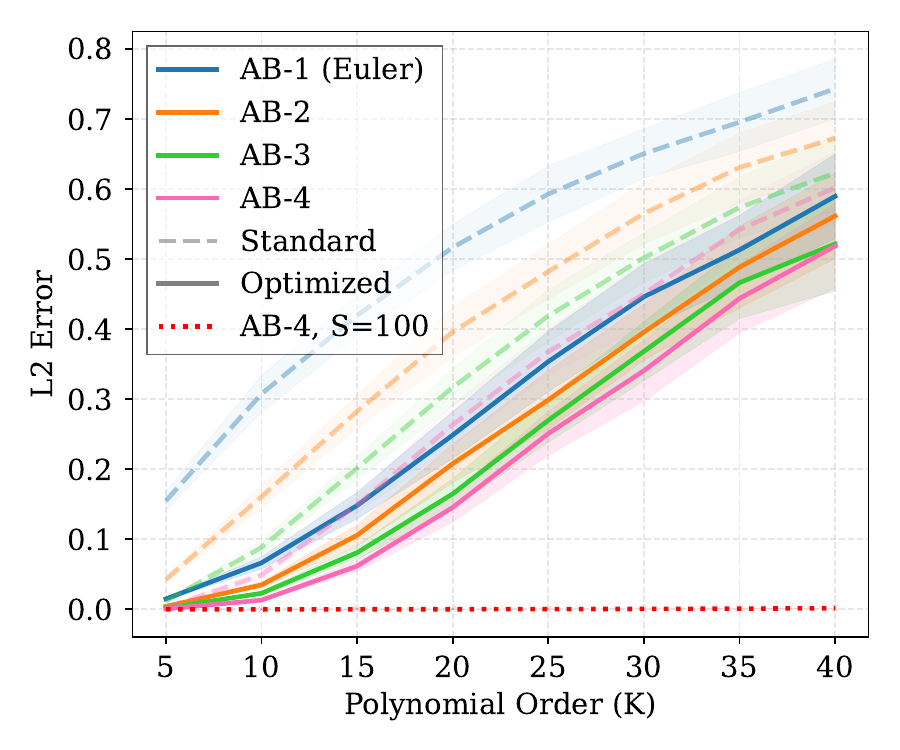}} & \hspace{-4mm}
        \subfloat[$S=10$]{\includegraphics[width=0.24\linewidth]{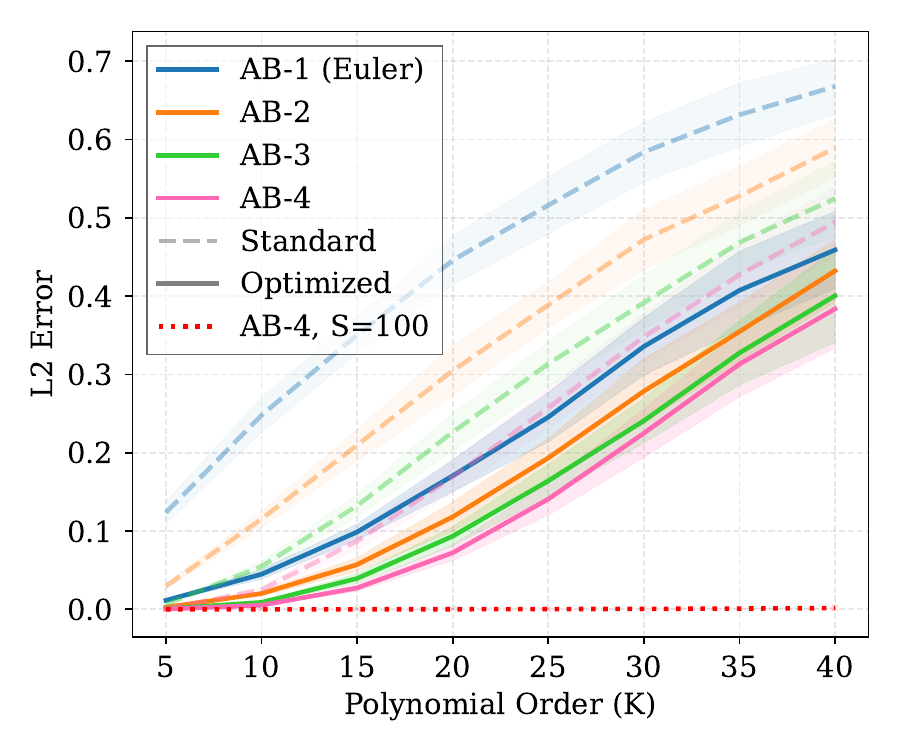}} & \hspace{-4mm}
        \subfloat[$S=12$]{\includegraphics[width=0.24\linewidth]{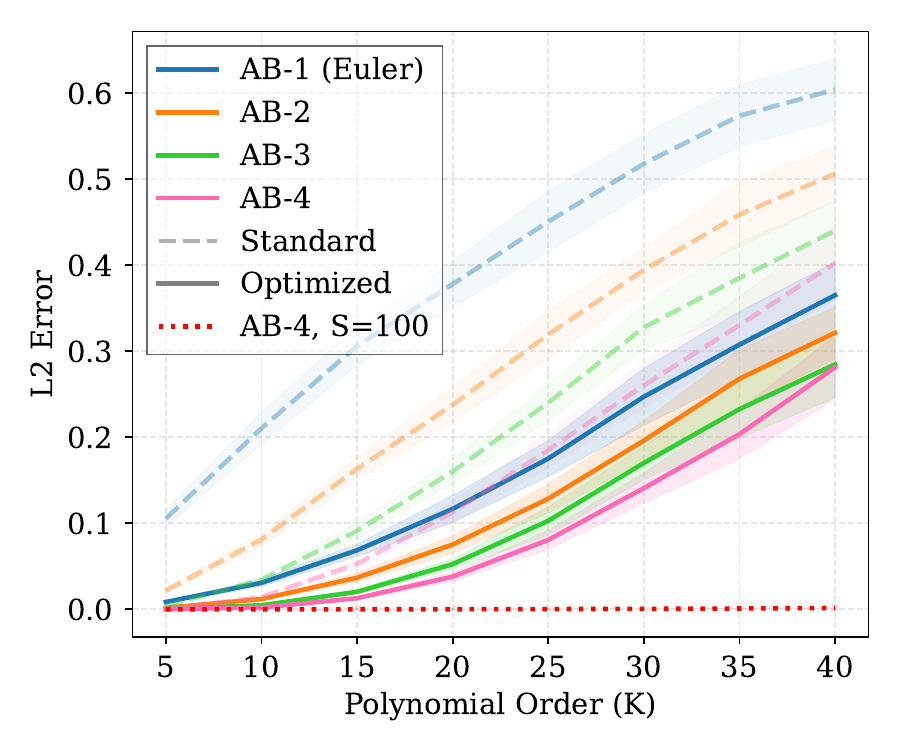}} \\

        \vspace{3mm} \hspace{-4mm}\subfloat[$S=14$]{\includegraphics[width=0.24\linewidth]{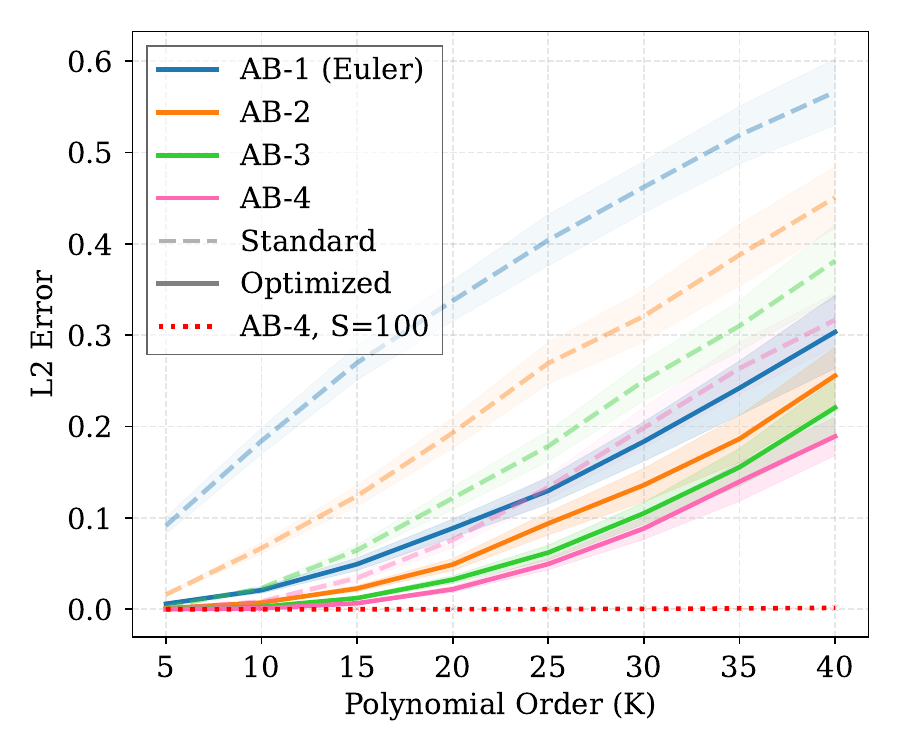}} & \hspace{-4mm}
        \subfloat[$S=16$]{\includegraphics[width=0.24\linewidth]{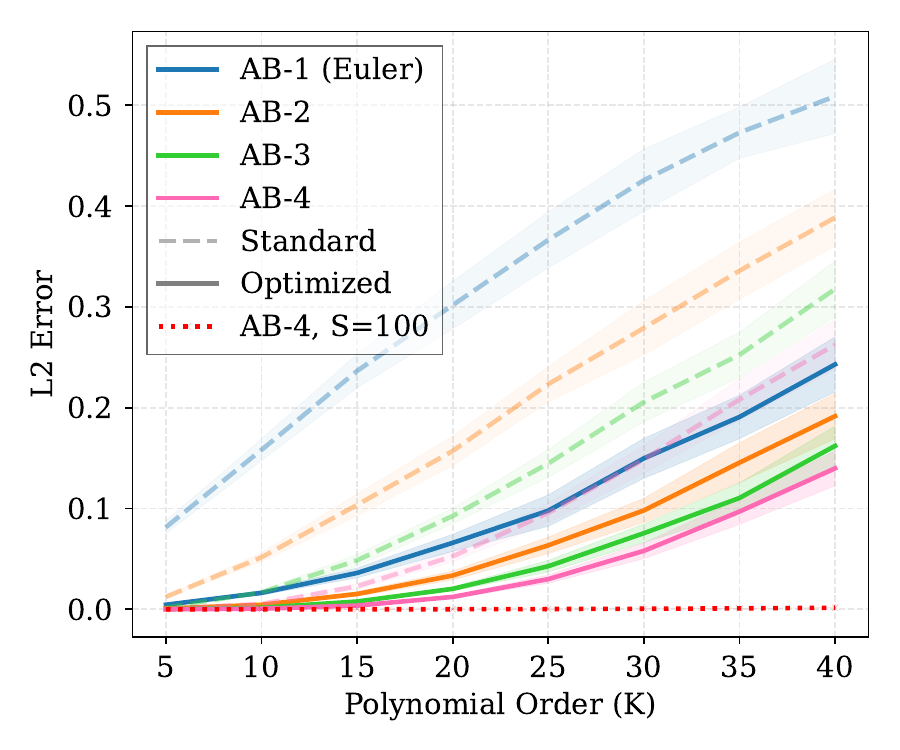}} & \hspace{-4mm}
        \subfloat[$S=18$]{\includegraphics[width=0.24\linewidth]{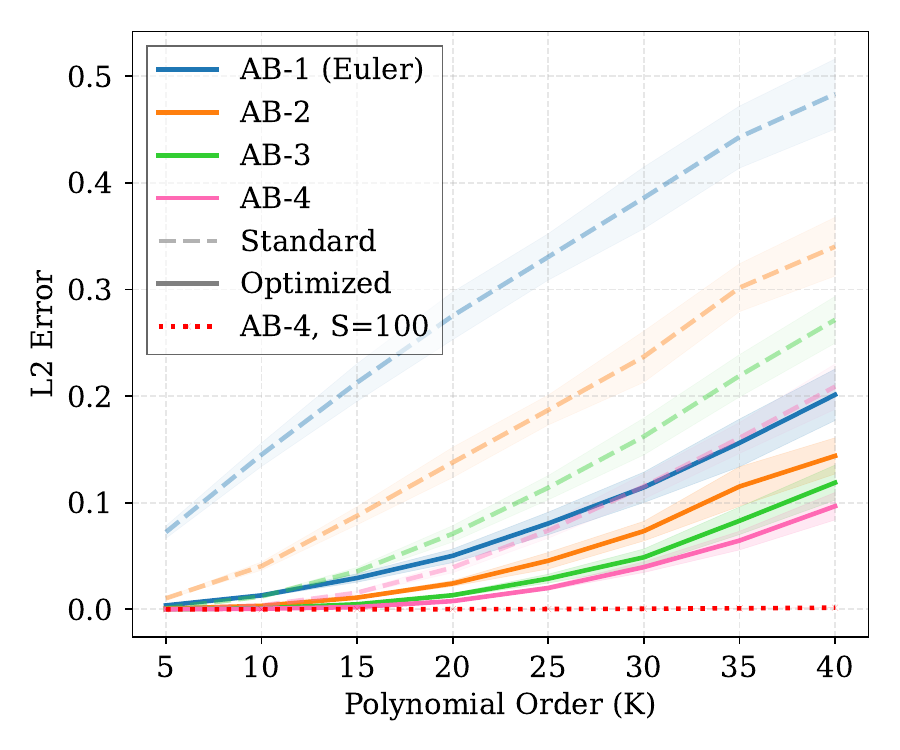}} & \hspace{-4mm}
        \subfloat[$S=20$]{\includegraphics[width=0.24\linewidth]{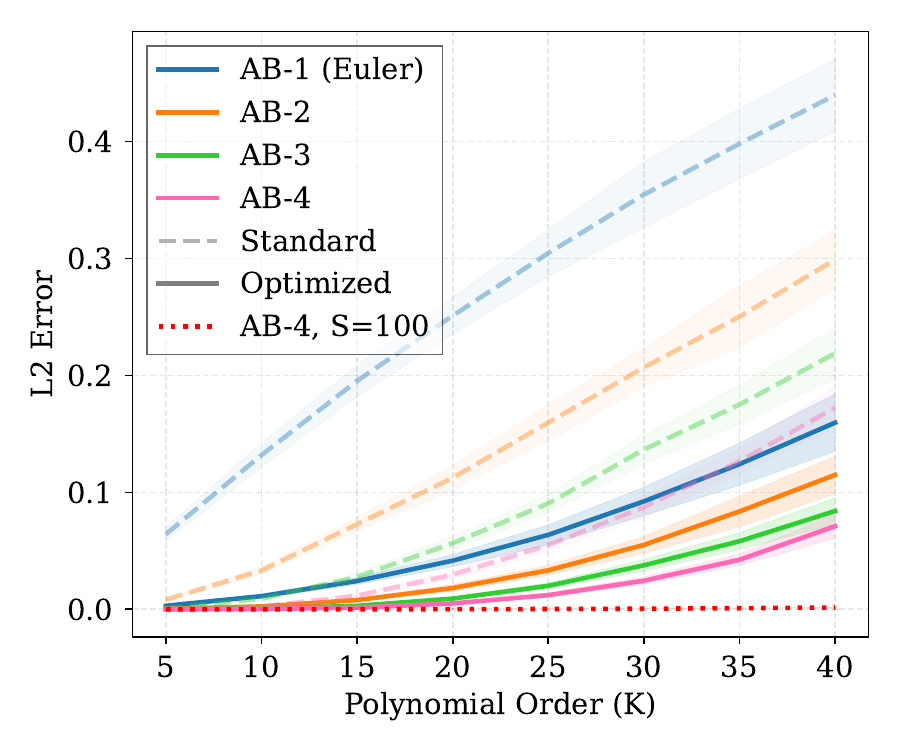}} \\

        \vspace{3mm} \hspace{-4mm} \subfloat[$K=5$]{\includegraphics[width=0.24\linewidth]{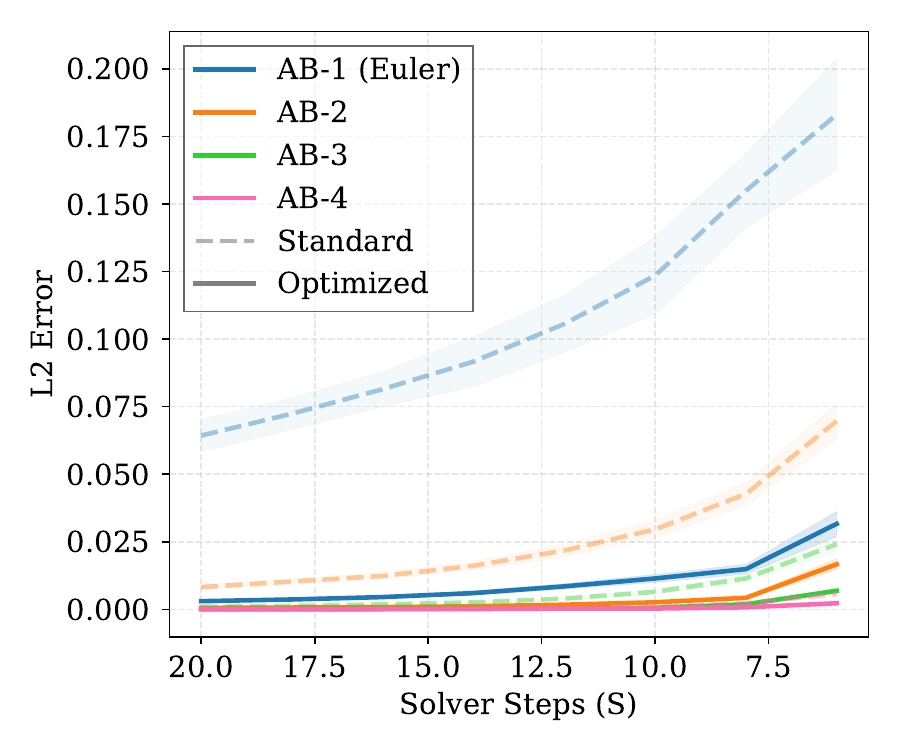}} & \hspace{-4mm}
        \subfloat[$K=10$]{\includegraphics[width=0.24\linewidth]{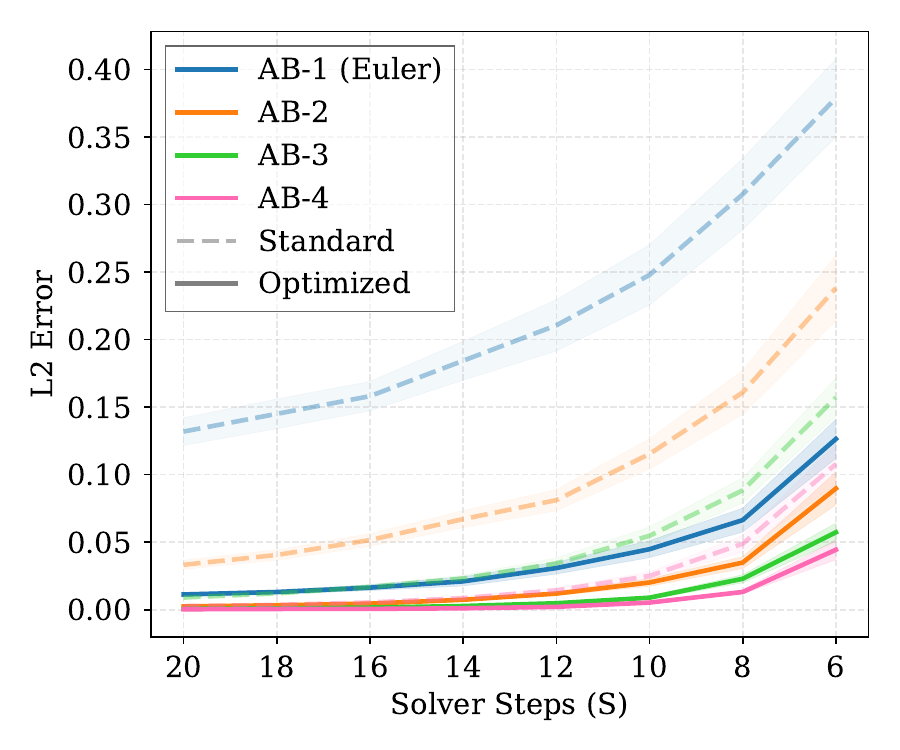}} & \hspace{-4mm}
        \subfloat[$K=15$]{\includegraphics[width=0.24\linewidth]{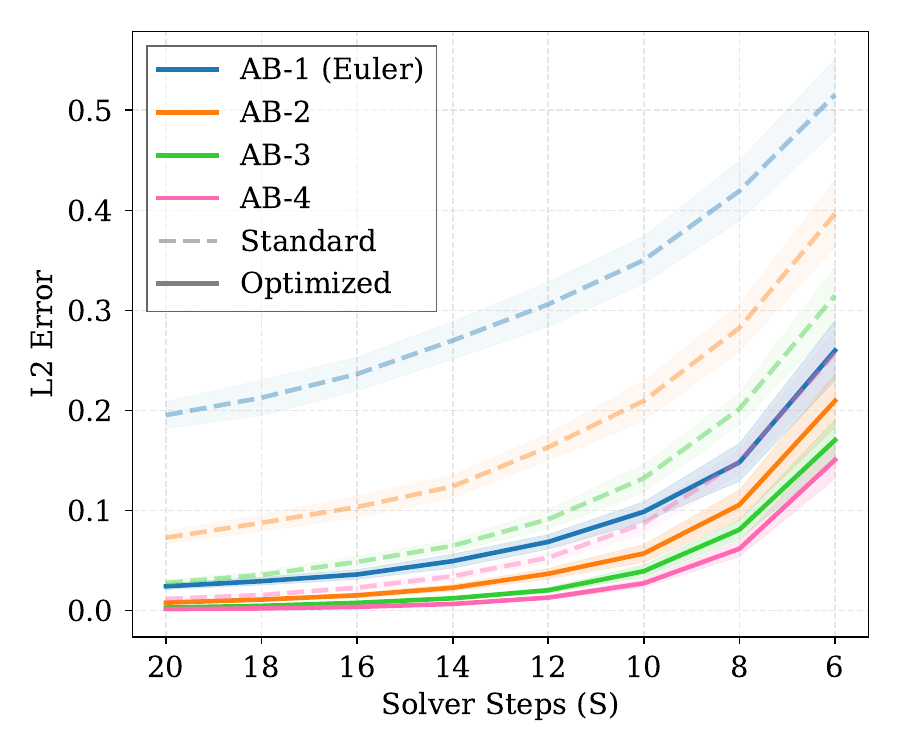}} & \hspace{-4mm}
        \subfloat[$K=20$]{\includegraphics[width=0.24\linewidth]{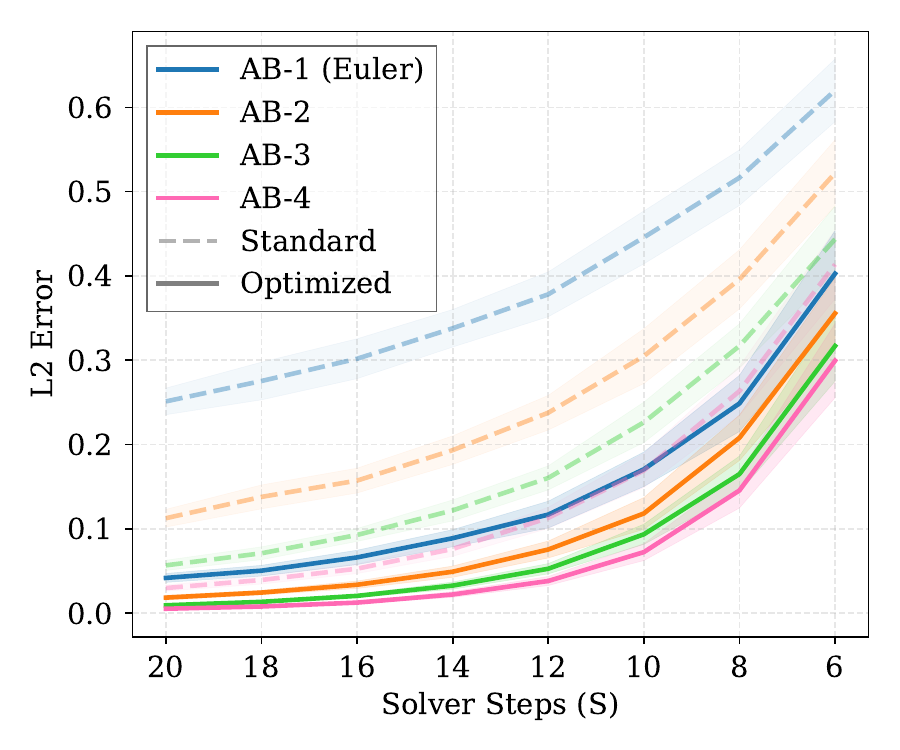}} \\

        \vspace{3mm} \hspace{-4mm} \subfloat[$K=25$]{\includegraphics[width=0.24\linewidth]{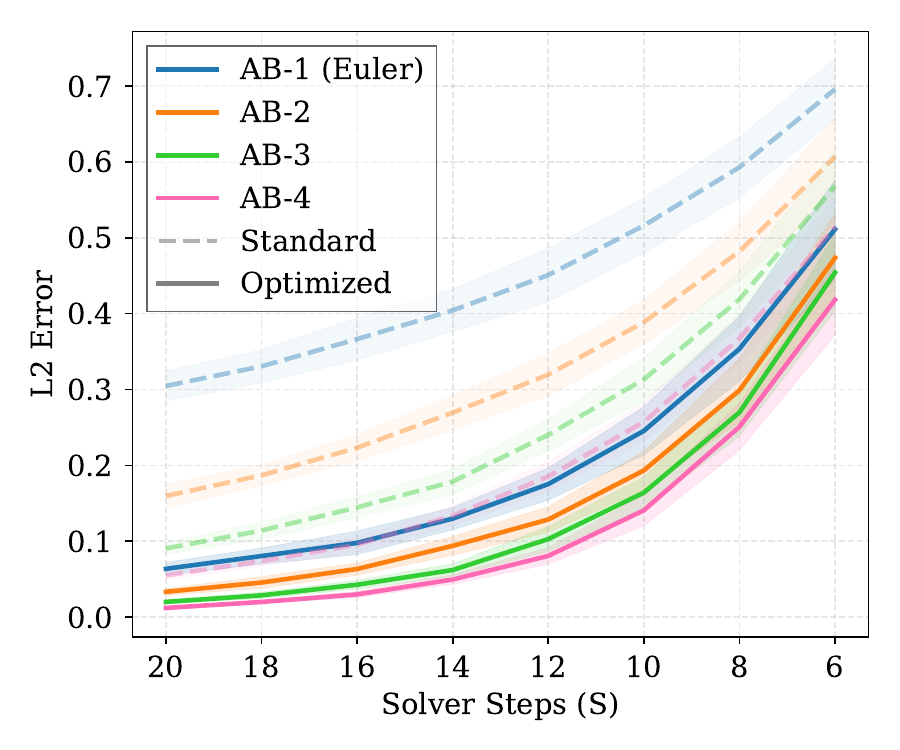}} & \hspace{-4mm}
        \subfloat[$K=30$]{\includegraphics[width=0.24\linewidth]{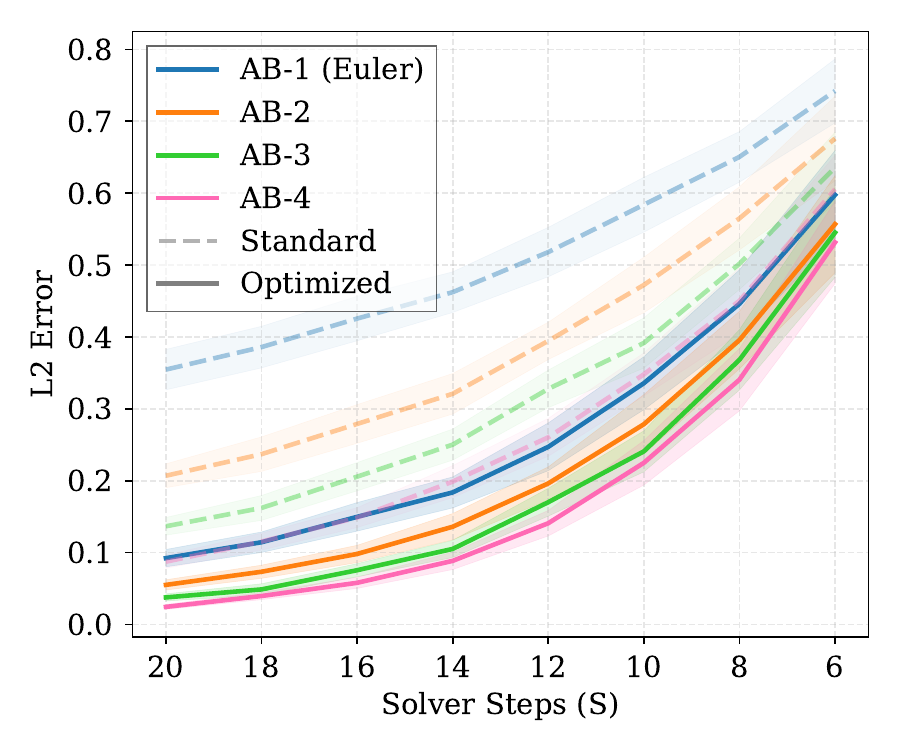}} & \hspace{-4mm}
        \subfloat[$K=35$]{\includegraphics[width=0.24\linewidth]{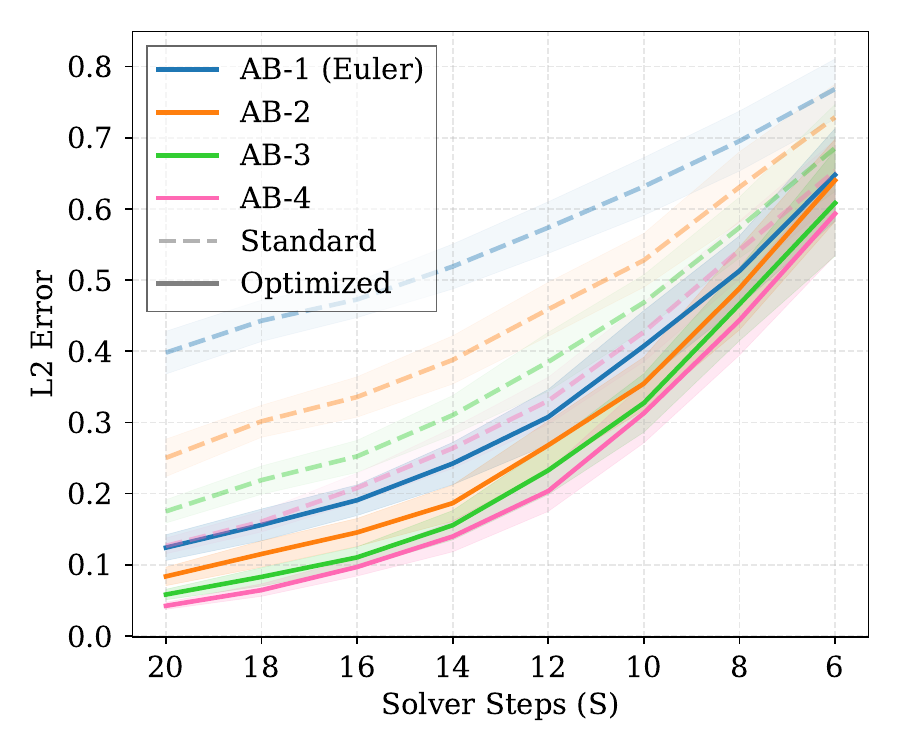}} & \hspace{-4mm}
        \subfloat[$K=40$]{\includegraphics[width=0.24\linewidth]{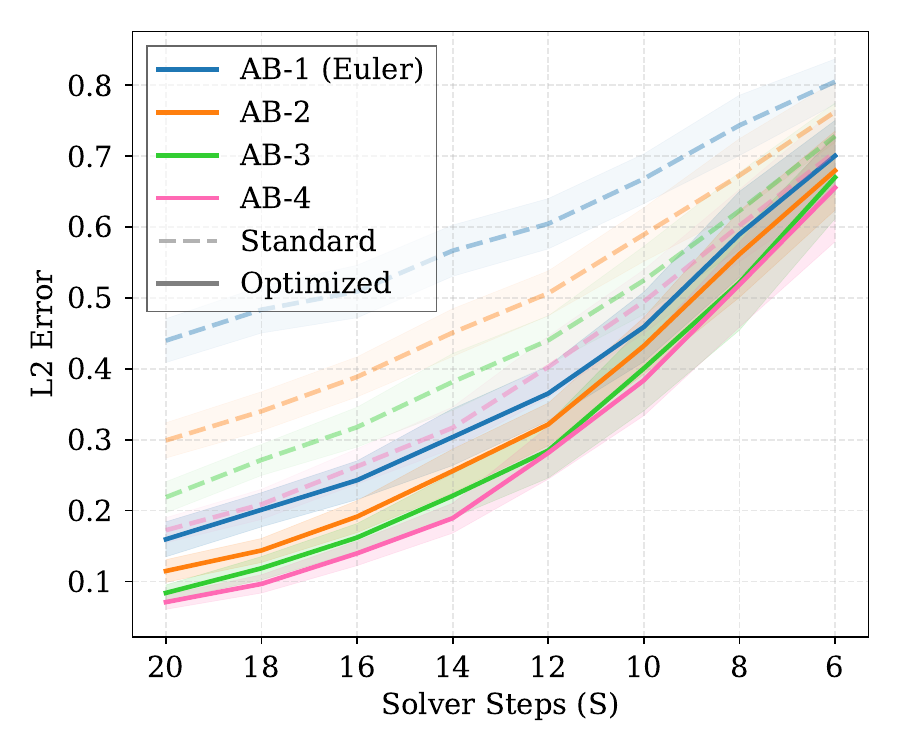}} \\
    \end{tabular}
    \caption{\textbf{Complete results of the numerical study.} L2 error under different complexity and steps settings. (a) -- (h) Error vs. Complexity: We fix the solver steps ($S$) and increase the problem complexity (polynomial order $K$). (i) -- (p) Error vs. Steps: We fix the complexity ($K$) and decrease the number of solver steps ($S$).}
    \label{fig:supp_num}
\end{figure*}

\vspace{0.5em}
\noindent
\textbf{Optimization and evaluation.} The optimized solvers are trained to minimize the Mean Squared Error (MSE) between the predicted terminal state $\mathbf{y}_\text{pred}(1)$ and the ground truth $\mathbf{y}_\text{gt}(1)$ (\textit{endpoint supervision}). We use the Adam optimizer \cite{opt_adam} with a learning rate of $1\times 10^{-3}$ for 2,000 iterations.
For the comprehensive evaluation shown in \cref{fig:supp_num}, we traverse a dense grid of configurations: polynomial orders $K \in \{5, 10, \dots, 40\}$ and step counts $S \in \{6, 8, \dots, 20\}$. For each configuration $(K, S)$, we perform 50 independent runs with randomly sampled problems and report the mean Relative $L_2$ Error, defined as $||\mathbf{y}_\text{pred}(1) - \mathbf{y}_\text{gt}(1)||_2 / ||\mathbf{y}_\text{gt}(1)||_2$. As illustrated in \cref{fig:supp_num}, the optimized solvers consistently achieve lower relative errors compared to their standard counterparts across all tested configurations.

\subsection{DyWeight experimental details}
In this section, we detail the experimental settings to ensure reproducibility. We describe the baseline configurations and initialization strategies, followed by the specific optimization setup and computational efficiency analysis of our method.

\vspace{0.5em}
\noindent
\textbf{Baseline Configurations.} We compare our method against solvers including iPNDM \cite{PNDM,iPNDM}, DPM-Solver++ \cite{dpmpp}, UniPC \cite{UniPC}, AMED \cite{AMED}, EPD \cite{EPD}, LD3 \cite{LD3}, DLMS \cite{DLMS} and S4S-Alt \cite{s4s}. A specific note regarding AMED \cite{AMED} and EPD \cite{EPD}: these methods typically utilize an Analytical First Step (AFS) \cite{afs} followed by two-stage (\eg, midpoint) updates, which naturally result in an odd number of NFEs per inference (\eg, $2k - 1_{\text{AFS}}$). Consequently, their results at even NFEs (\eg, 4, 6, 8) are reported without AFS \cite{afs} then derived from suboptimal configurations, often leading to performance degradation as noted in their original reports. For fair comparison, we report their best-performing odd-NFE results in the main paper, while the supplementary provides full results (\cref{sec:supp_quant}) including AMED’s even-step evaluations.

\vspace{0.5em}
\noindent
\textbf{Initialization strategies.} Since \ours~dynamically adjusts the effective step size via learned gradient weights rather than explicitly predicting time values (like DLMS \cite{DLMS}) or schedules (like LD3 \cite{LD3}), proper initialization of the time schedule and weights is crucial for stability.
\begin{itemize}
    \item \textbf{Gradient weights initialization}: We initialize the learnable weight matrix $\textbf{W}$ using the coefficients of the standard Adams-Bashforth (AB) \cite{numerical_analysis,numerical_ode,PNDM,iPNDM} method. We also explored alternative strategies, including Euler initialization \cite{numerical_analysis,numerical_ode,DDIM} (setting the current step weight to 1 and history weights to 0) and Uniform initialization (averaging weights as $1/N_{\text{avail}}$). As analyzed in \cref{tab:supp_init_comparison}, AB initialization \cite{numerical_analysis,numerical_ode} consistently yields the best final performance.
    \item \textbf{Time schedule initialization}: \ours~operates on a base time schedule which is then dynamically modulated by the learned weights (\textit{time shifting}). In pixel-space models \cite{cifar,ffhq,afhqv2,imagenet}, we adopt the polynomial time schedule with $\rho=7$, consistent with the EDM \cite{EDM} default configuration. In latent-space models \cite{lsun,sd1.5,mscoco}, we utilize a discrete uniform schedule (linear spacing in $\sigma$ steps, $\rho=1$). Note that this differs from the Time Uniform schedule used in some papers with a DDPM \cite{DDPM} backbone. As detailed in \cref{tab:supp_init_comparison}, the choice of initialization schedule remains a relevant factor. While \ours~achieves comparably superior performance using either EDM \cite{EDM} or LogSNR \cite{dpmsolver,UniPC} schedules, initializing with a Time Uniform schedule leads to noticeable performance degradation.
\end{itemize}

\begin{table}[t]
  \caption{\textbf{Comparison of different initialization strategies for} \ours\textbf{.} We report the FID \cite{fid} scores of different initialization methods across varying NFEs and datasets, all optimized using 10K training pairs. \textbf{Top:} Comparison of gradient weights initialization strategies. \textit{Adams-Bashforth} \cite{numerical_analysis,numerical_ode} initializes weights using standard coefficients (iPNDM \cite{PNDM,iPNDM}); \textit{Euler} \cite{numerical_analysis,numerical_ode} sets the current step weight to 1 and others to 0; \textit{Uniform} distributes weights evenly. \textbf{Bottom:} Comparison of time schedule initialization methods. We set $\rho=7$ for the EDM (Polynomial) \cite{EDM} schedule and $\rho=1$ for the Time Uniform schedule.}
  \label{tab:supp_init_comparison}
  \centering
  \fontsize{8}{10}\selectfont
  \setlength{\tabcolsep}{5pt} 
  
  \begin{tabular}{lcccccc}
    \toprule
    \multirow{2}{*}{Dataset / NFE} & \multicolumn{3}{c}{CIFAR-10 \cite{cifar}} & \multicolumn{3}{c}{AFHQv2 \cite{afhqv2}} \\
    \cmidrule(lr){2-4} \cmidrule(lr){5-7}
     & 4 & 6 & 8 & 4 & 6 & 8 \\
    \midrule
    
    \rowcolor{gray!25} 
    \multicolumn{7}{l}{\textbf{Weights initialization}} \\
    \midrule 
    
    Adams-Bashforth \cite{numerical_analysis,numerical_ode} & \textbf{3.92} & \textbf{2.61} & \textbf{2.18} & \textbf{5.38} & \textbf{2.66} & \textbf{2.23} \\
    Euler \cite{numerical_analysis,numerical_ode}     & 6.46 & 3.92 & 2.80 & 8.26 & 4.19 & 2.77 \\
    Uniform   & 9.40 & 5.40 & 4.39 & 10.25 & 7.86 & 5.76 \\
    
    \midrule
    
    \rowcolor{gray!25} 
    \multicolumn{7}{l}{\textbf{Time schedule initialization}} \\
    \midrule 
    
    EDM \cite{EDM}   & \textbf{3.92} & \textbf{2.61} & \textbf{2.18} & \textbf{5.38} & \textbf{2.66} & \textbf{2.23} \\
    Time Uniform & 6.73 & 5.15 & 4.66 & 6.67 & 3.89 & 3.57 \\
    LogSNR \cite{dpmsolver,UniPC}     & 3.98 & 2.67 & 2.22 & 5.46 & 3.23 & 2.34 \\
    
    \bottomrule
  \end{tabular}
\end{table}

\vspace{0.5em}
\noindent
\textbf{Optimization setup.} We train \ours~using a teacher-student distillation framework with \textit{endpoint supervision}. For \textbf{teacher solver}, We employ a 35-NFE iPNDM \cite{PNDM,iPNDM} solver as the teacher for pixel-space \cite{cifar,ffhq,afhqv2,imagenet} and LSUN-Bedroom \cite{lsun} benchmarks, and a 50-NFE iPNDM (2M) \cite{PNDM,iPNDM} for Stable Diffusion v1.5 \cite{sd1.5}. As demonstrated in \cref{tab:abl-teacher-solver}, our method is agnostic to the specific teacher trajectory. The student only requires a high-quality terminal target for supervision. For \textbf{loss function}, we minimize the squared $\ell_2$ distance within the feature space of the final layer of a pre-trained Inception network \cite{inception} for pixel-space \cite{cifar,ffhq,afhqv2,imagenet} and LSUN-Bedroom \cite{lsun} datasets, and minimize the standard MSE ($\ell_2$) loss for Stable Diffusion v1.5 \cite{sd1.5} and FLUX.1-dev \cite{flux,flux_paper}. For \textbf{hyperparameters}, we use the Adam optimizer \cite{opt_adam} with $\beta_1=0.9, \beta_2=0.999, \epsilon=10^{-8}$. The learning rate (LR) follows a cosine annealing schedule with a linear warmup. The LR ranges from $1\times10^{-3}$ to $5\times10^{-2}$ depending on the student step count (lower LR for higher steps). We set the linear warmup ratio to 0.1 with a start factor of 0.1, and the cosine decay minimum ratio to 0.01. For \textbf{data}, we generate 10K teacher-student pairs for pixel-space \cite{cifar,ffhq,afhqv2,imagenet} and LSUN-Bedroom \cite{lsun} datasets, and 5K pairs for SD v1.5 \cite{sd1.5} and FLUX.1-dev \cite{flux,flux_paper}. We empirically observe that our optimization requires minimal data to converge. Benefiting from a strong inductive bias via proper initialization, the training loss undergoes a substantial reduction in the initial phase of optimization, as depicted in \cref{fig:convergence}. As a result, further expanding the dataset to 20K or 30K samples yields comparable results (\cref{tab:abl-training-size}), confirming that the solver effectively reaches optimal convergence with only 10K samples. This demonstrates the remarkable data efficiency of our method.

\begin{table}[t]
  \caption{Impact of distillation dataset size (number of teacher samples) on CIFAR-10 \cite{cifar} and ImageNet \cite{imagenet}. Performance saturates at 10K samples, demonstrating data efficiency.}
  \label{tab:abl-training-size}
  \centering
      \fontsize{8}{10}\selectfont
      \setlength{\tabcolsep}{4pt} 
    \begin{tabular}{lcccccc}
      \toprule
      \multirow{2}{*}{Size} & \multicolumn{3}{c}{CIFAR-10 \cite{cifar}} & \multicolumn{3}{c}{ImageNet \cite{imagenet}}\\
      \cmidrule(lr){2-4}\cmidrule(){5-7}& NFE=4 & NFE=6 & NFE=8 & NFE=4 & NFE=6 & NFE=8 \\
      \midrule
      10K& \textbf{3.92} & 2.63 & 2.26 & \textbf{11.48} & 6.15 & 4.69 \\
      20K& 4.09 & \textbf{2.61} & \textbf{2.18} & 12.94 & \textbf{5.80} & 5.02 \\
      30K& 4.35 & 2.68 & 2.41 & 13.32 & 6.16 & \textbf{4.68} \\
      \bottomrule
    \end{tabular}
\end{table}

\vspace{0.5em}
\noindent
\textbf{Computational efficiency.} The training cost of our method is minimal, primarily determined by the NFE of the teacher solver and the dataset size. The student NFE has a secondary effect related to the gradient backpropagation depth. We benchmark the training time as follows (including the time required to generate teacher samples):
\begin{itemize}
    \item \textbf{CIFAR-10} \cite{cifar} (10K pairs): On 2$\times$ RTX 4090 GPUs, training takes approximately 2 minutes for an 18-NFE teacher, 3 minutes for a 35-NFE teacher, and 4 minutes for a 50-NFE teacher.
    \item \textbf{ImageNet-64} \cite{imagenet} (10K pairs): On 8$\times$ RTX 4090 GPUs with a 35-NFE teacher, training requires approximately 5 minutes.
    \item \textbf{LSUN-Bedroom} \cite{lsun} (10K pairs): On 8$\times$ H100 GPUs with a 35-NFE teacher, training takes approximately 9 minutes.
    \item \textbf{Stable Diffusion v1.5} \cite{sd1.5} (5K pairs): On 8$\times$ H100 GPUs with a 50-NFE teacher, training completes in approximately 35 minutes. 
\end{itemize}
This rapid training capability enables quick deployment of \ours~on new models.

\vspace{0.5em}
\noindent
\textbf{Implementation details on FLUX.1-dev (FLUX)}~\cite{flux,flux_paper}\textbf{.} 
To validate the effectiveness of our method on the \textbf{state-of-the-art} text-to-image generative model, we train \ours\ solver parameters on FLUX \cite{flux,flux_paper} for each target step count using the teacher--student distillation framework described in \cref{sec:method}. The official FLUX~\cite{flux,flux_paper} Euler sampler with its 28-step configuration is used as the teacher. Following the recommended time schedule, both teacher and student solvers construct their time schedule by first uniformly partitioning the diffusion interval according to the desired number of steps, followed by the deterministic time-shifting adjustment used in the official sampler.
We randomly sample 5K prompts from the MS-COCO~\cite{mscoco} validation set to build the training pairs, generating one teacher trajectory per prompt. Student solvers are trained for 3, 5, and 7 steps. Training is conducted on 8$\times$ H100 GPUs and takes approximately 42 minutes 
(3-step), 50 minutes (5-step), and 55 minutes (7-step). As only solver parameters are optimized and the FLUX~\cite{flux,flux_paper} backbone is kept frozen, the proposed method introduces no additional inference latency. Because FLUX~\cite{flux,flux_paper} integrates classifier-free guidance into its distillation, each sampling step corresponds to one function evaluation (\ie, steps are equivalent to NFEs).

\section{Full results}
\subsection{Quantitative results}
\label{sec:supp_quant}

In this section, we provide the complete quantitative evaluation results comparing \ours~against handcrafted and state-of-the-art learnable solvers. 
\cref{tab:supp_cifar10} to \cref{tab:supp_afhq} report the FID \cite{fid} results on unconditional generation benchmarks, including CIFAR-10 ($32\times32$) \cite{cifar}, FFHQ ($64\times64$) \cite{ffhq}, and AFHQv2 ($64\times64$) \cite{afhqv2}. 
\cref{tab:supp_imagenet} presents the results for conditional generation on ImageNet ($64\times64$) \cite{imagenet}, and \cref{tab:supp_lsun} shows the performance on LSUN-Bedroom ($256\times256$) \cite{lsun}. 
Consistent with the results in the main paper, \ours~demonstrates superior performance and robustness across all datasets and step settings, particularly excelling in the low-NFE regime.

\subsection{Qualitative results}
\label{sec:supp_qualitative_results}
\begin{figure*}[ht]
\centering

\subfloat[iPNDM]{
    \includegraphics[width=0.43\linewidth]{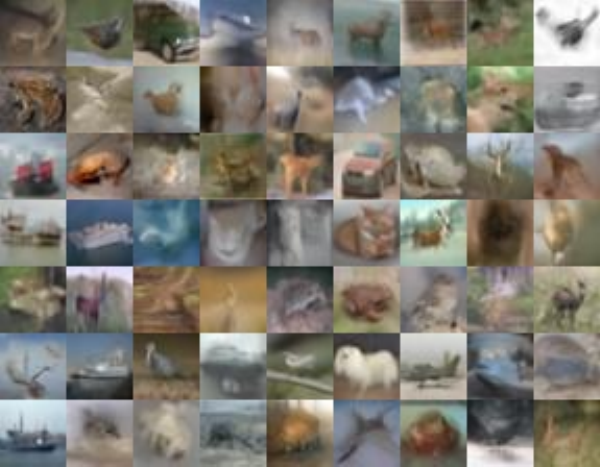}
    \label{fig:supp_cifar_3nfe_ipndm}} 
\hspace{4mm}
\subfloat[\ours]{
    \includegraphics[width=0.43\linewidth]{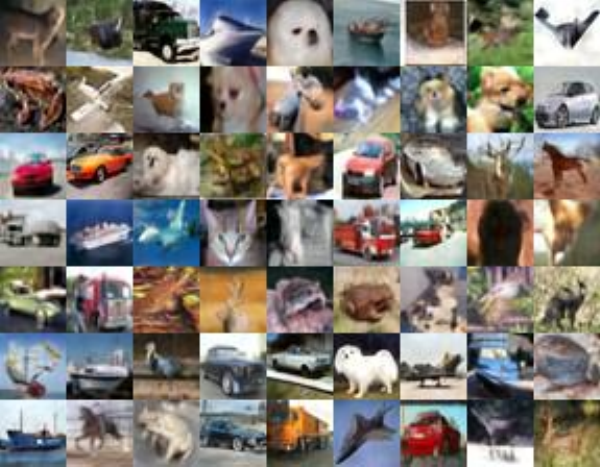}
    \label{fig:supp_cifar_3nfe_dy}
    }
    \vspace{-1mm}
    \caption{Qualitative comparison on CIFAR-10 ($32 \times 32$) \cite{cifar} at 3 NFEs. \ours~exhibits significantly reduced noise and better structure compared to iPNDM \cite{PNDM,iPNDM}.}
    \label{fig:supp_cifar_3nfe}

\end{figure*}

\begin{figure*}[ht]
\centering

\subfloat[iPNDM]{
    \includegraphics[width=0.43\linewidth]{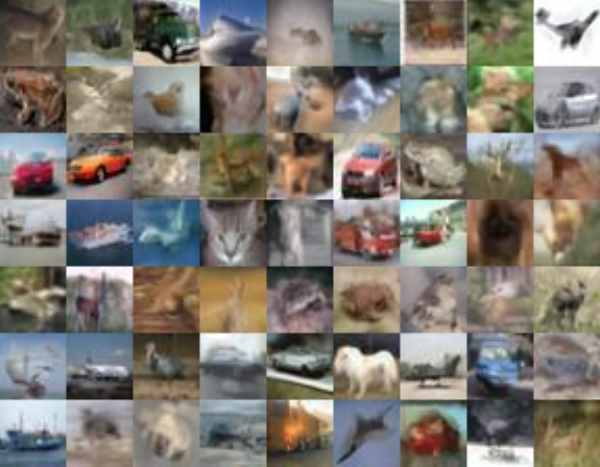}
    \label{fig:supp_cifar_4nfe_ipndm}} 
\hspace{4mm}
\subfloat[\ours]{
    \includegraphics[width=0.43\linewidth]{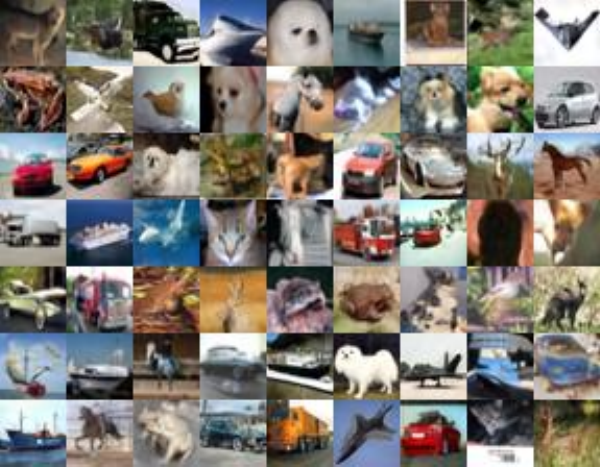}
    \label{fig:supp_cifar_4nfe_dy}
    }
    \vspace{-1mm}
    \caption{Qualitative comparison on CIFAR-10 ($32 \times 32$) \cite{cifar} at 4 NFEs. \ours~exhibits significantly reduced noise and better structure compared to iPNDM \cite{PNDM,iPNDM}.}
    \label{fig:supp_cifar_4nfe}

\end{figure*}

\begin{figure*}[ht]
\centering

\subfloat[iPNDM]{
    \includegraphics[width=0.43\linewidth]{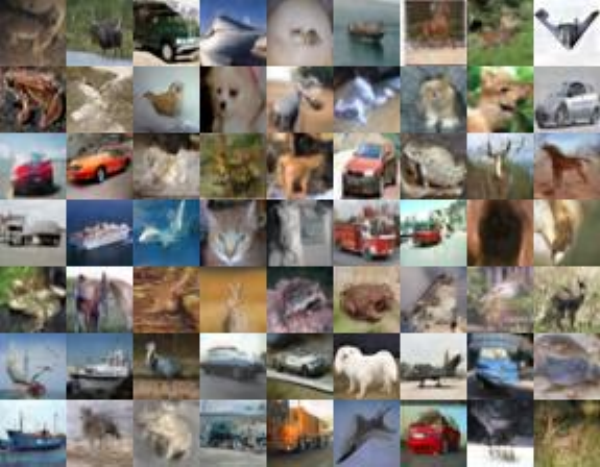}
    \label{fig:supp_cifar_5nfe_ipndm}} 
\hspace{4mm}
\subfloat[\ours]{
    \includegraphics[width=0.43\linewidth]{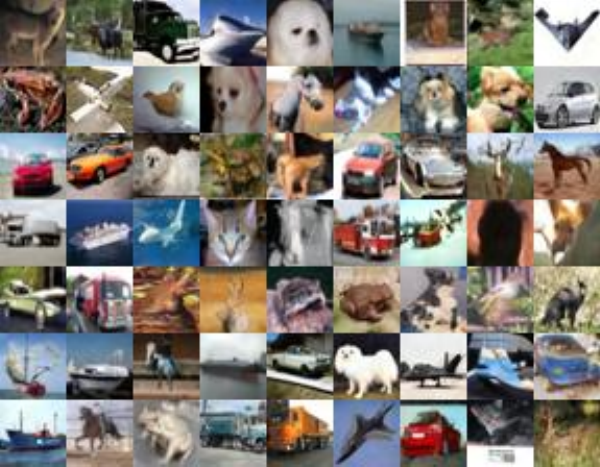}
    \label{fig:supp_cifar_5nfe_dy}
    }
    \vspace{-1mm}
    \caption{Qualitative comparison on CIFAR-10 ($32 \times 32$) \cite{cifar} at 5 NFEs. \ours~exhibits significantly reduced noise and better structure compared to iPNDM \cite{PNDM,iPNDM}.}
    \label{fig:supp_cifar_5nfe}

\end{figure*}

\begin{figure*}[ht]
\centering

\subfloat[iPNDM]{
    \includegraphics[width=0.45\linewidth]{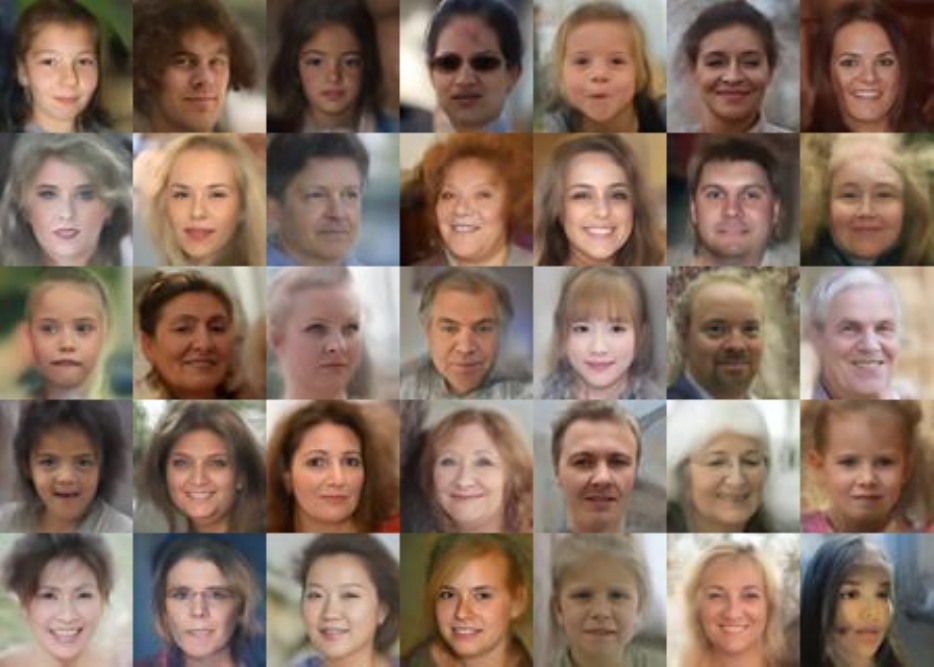}
    \label{fig:supp_ffhq_3nfe_ipndm}} 
\hspace{-0.5mm}
\subfloat[\ours]{
    \includegraphics[width=0.45\linewidth]{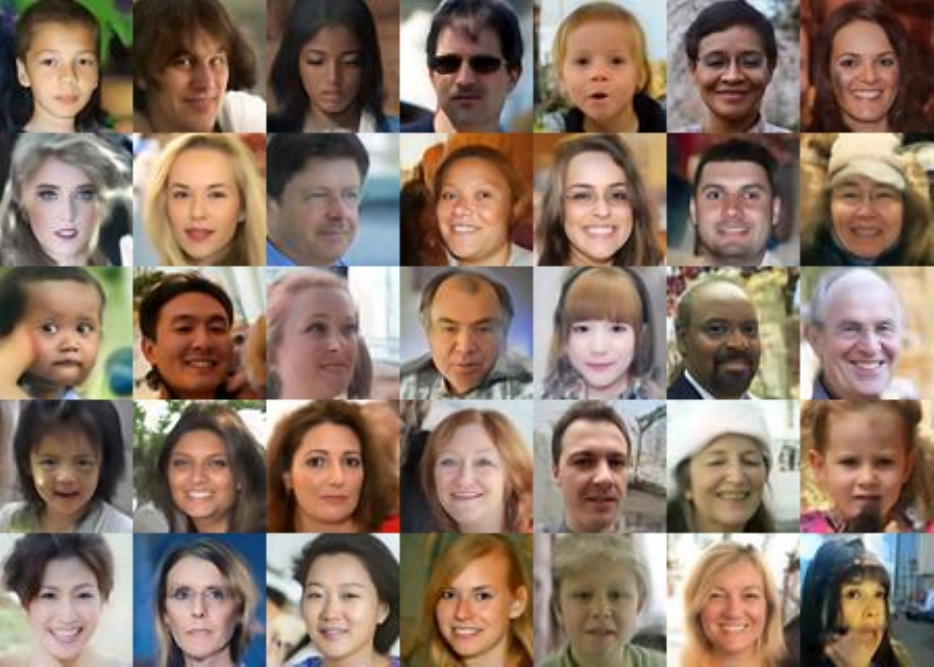}
    \label{fig:supp_ffhq_3nfe_dy}
    }
    \caption{Qualitative comparison on FFHQ ($64 \times 64$) \cite{ffhq} at 3 NFEs. \ours~yields fewer artifacts and clearer facial details compared with iPNDM \cite{PNDM,iPNDM}.}
    \label{fig:supp_ffhq_3nfe}

\end{figure*}

\begin{figure*}[ht]
\centering

\subfloat[iPNDM]{
    \includegraphics[width=0.45\linewidth]{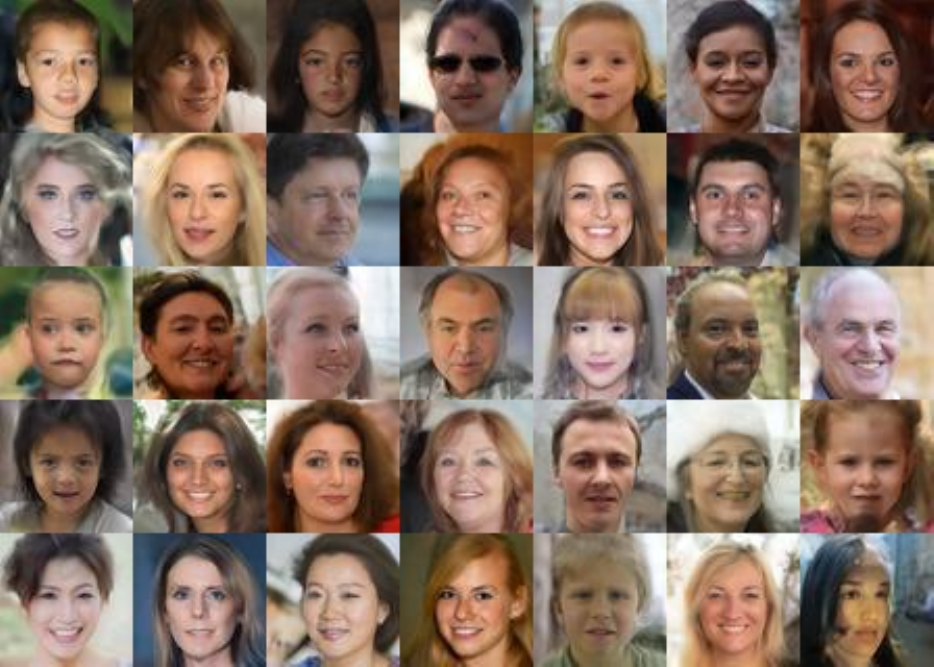}
    \label{fig:supp_ffhq_4nfe_ipndm}} 
\hspace{-0.5mm}
\subfloat[\ours]{
    \includegraphics[width=0.45\linewidth]{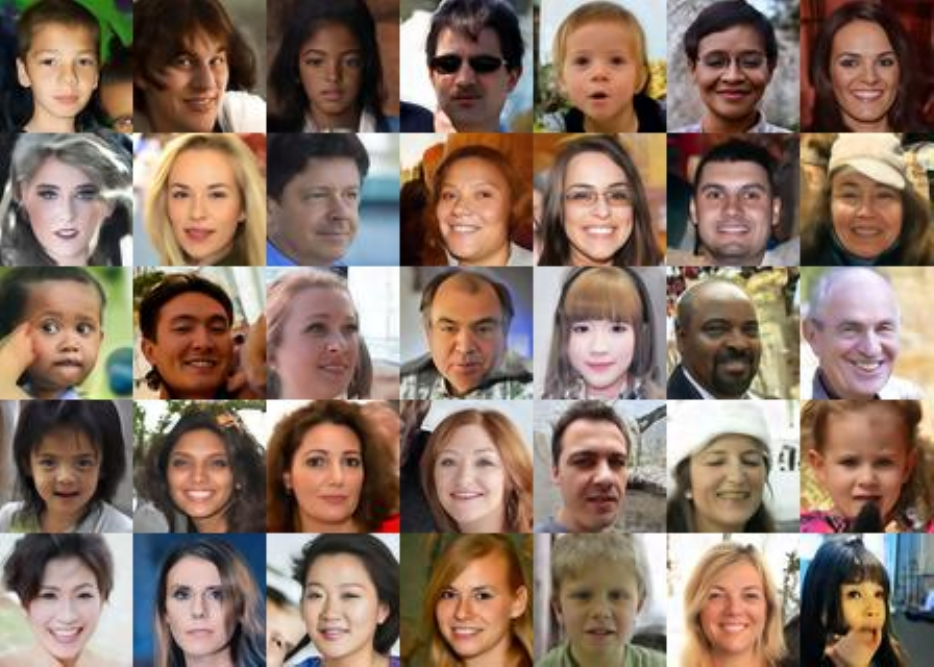}
    \label{fig:supp_ffhq_4nfe_dy}
    }
    \caption{Qualitative comparison on FFHQ ($64 \times 64$) \cite{ffhq} at 4 NFEs. \ours~yields fewer artifacts and clearer facial details compared with iPNDM \cite{PNDM,iPNDM}.}
    \label{fig:supp_ffhq_4nfe}

\end{figure*}

\begin{figure*}[ht]
\centering

\subfloat[iPNDM]{
    \includegraphics[width=0.45\linewidth]{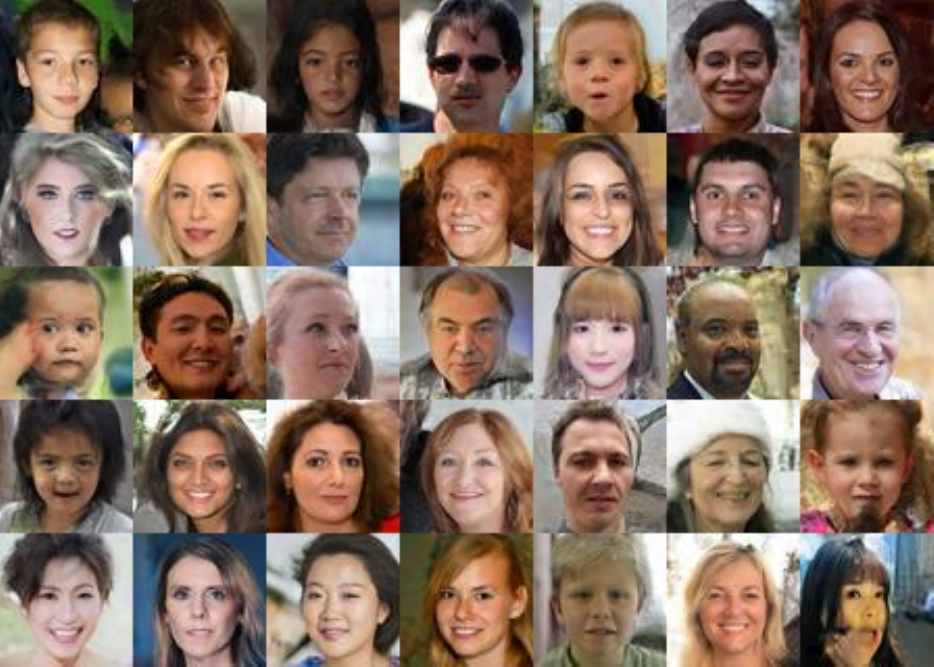}
    \label{fig:supp_ffhq_5nfe_ipndm}} 
\hspace{-0.5mm}
\subfloat[\ours]{
    \includegraphics[width=0.45\linewidth]{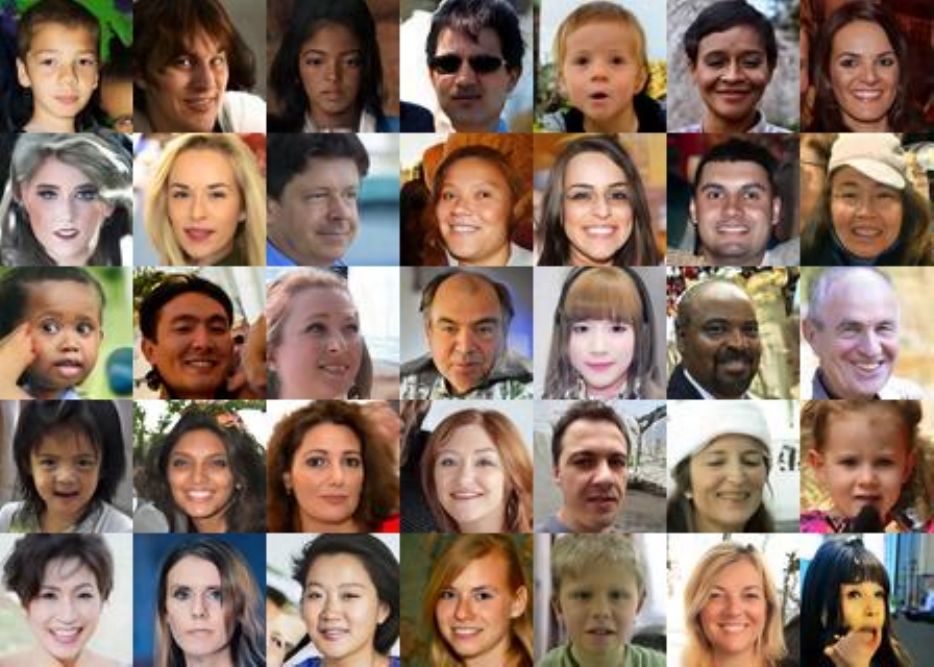}
    \label{fig:supp_ffhq_5nfe_dy}
    }
    \caption{Qualitative comparison on FFHQ ($64 \times 64$) \cite{ffhq} at 5 NFEs. \ours~yields fewer artifacts and clearer facial details compared with iPNDM \cite{PNDM,iPNDM}.}
    \label{fig:supp_ffhq_5nfe}

\end{figure*}

\begin{figure*}[ht]
\centering

\subfloat[iPNDM]{
    \includegraphics[width=0.45\linewidth]{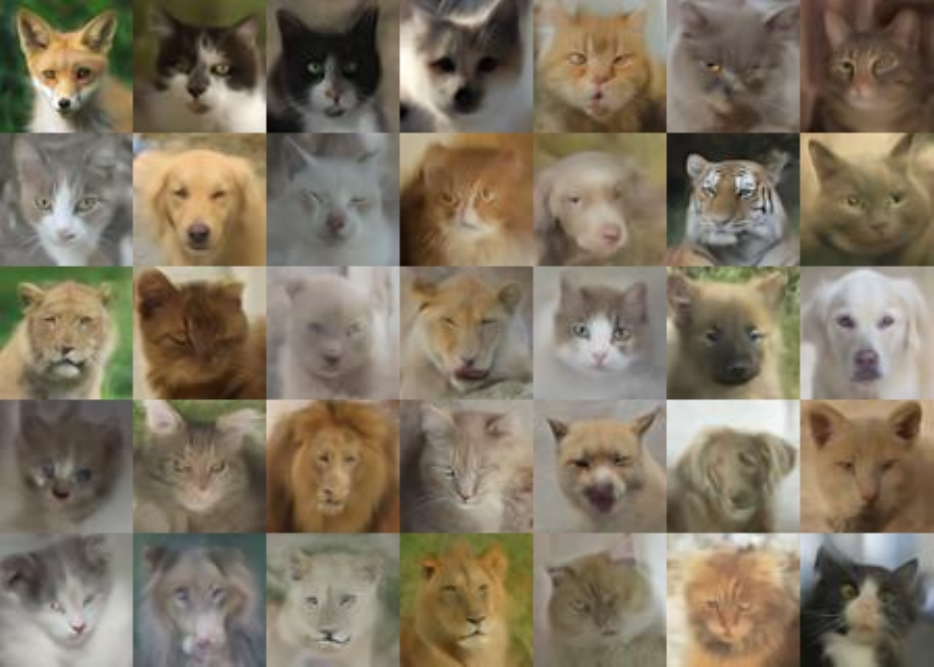}
    \label{fig:supp_afhq_3nfe_ipndm}} 
\hspace{-0.5mm}
\subfloat[\ours]{
    \includegraphics[width=0.45\linewidth]{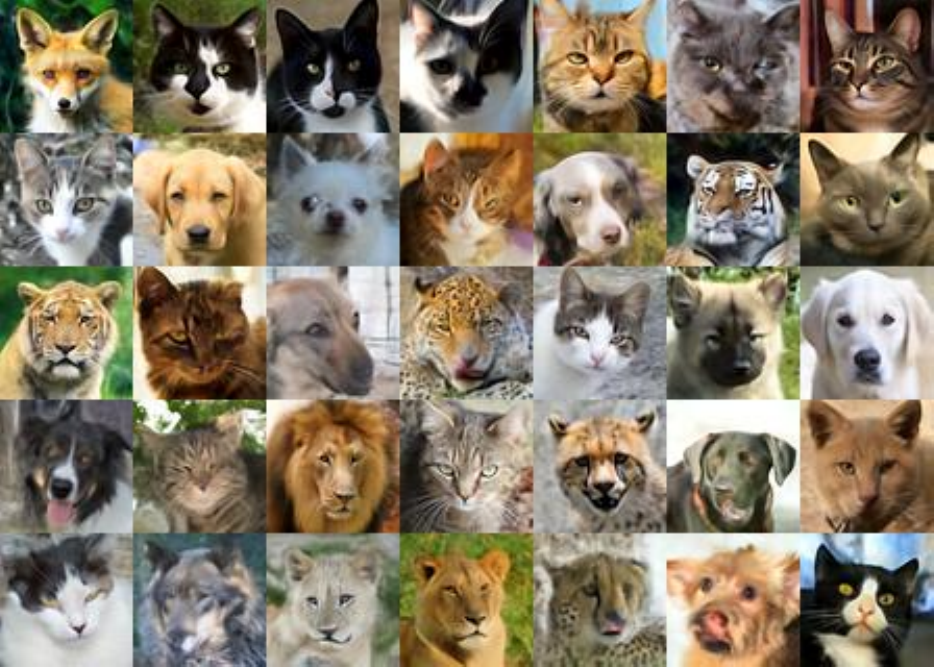}
    \label{fig:supp_afhq_3nfe_dy}
    }
    \caption{Qualitative comparison on AFHQv2 ($64 \times 64$) \cite{afhqv2} at 3 NFEs. \ours~produces animal features with higher fidelity and noticeably less blurring compared with iPNDM \cite{PNDM,iPNDM}.}
    \label{fig:supp_afhq_3nfe}

\end{figure*}

\begin{figure*}[ht]
\centering

\subfloat[iPNDM]{
    \includegraphics[width=0.45\linewidth]{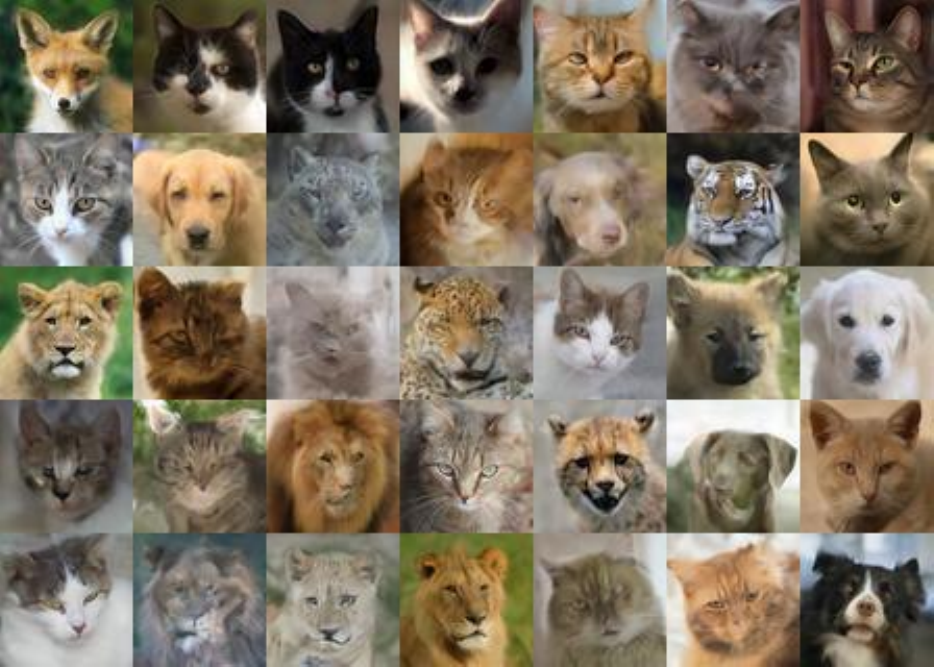}
    \label{fig:supp_afhq_4nfe_ipndm}} 
\hspace{-0.5mm}
\subfloat[\ours]{
    \includegraphics[width=0.45\linewidth]{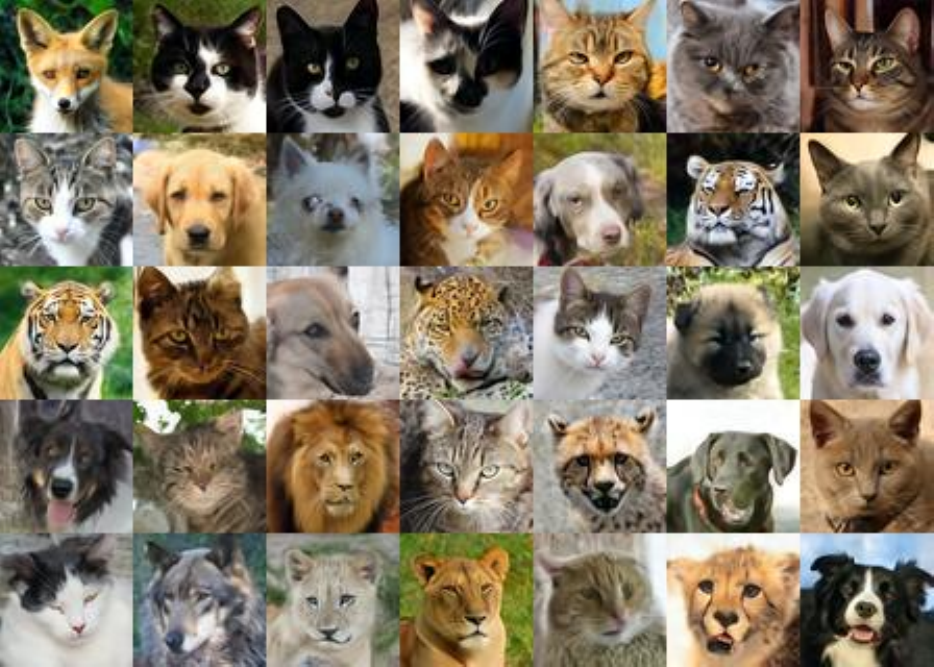}
    \label{fig:supp_afhq_4nfe_dy}
    }
    \caption{Qualitative comparison on AFHQv2 ($64 \times 64$) \cite{afhqv2} at 4 NFEs. \ours~produces animal features with higher fidelity and noticeably less blurring compared with iPNDM \cite{PNDM,iPNDM}.}
    \label{fig:supp_afhq_4nfe}

\end{figure*}

\begin{figure*}[ht]
\centering

\subfloat[iPNDM]{
    \includegraphics[width=0.45\linewidth]{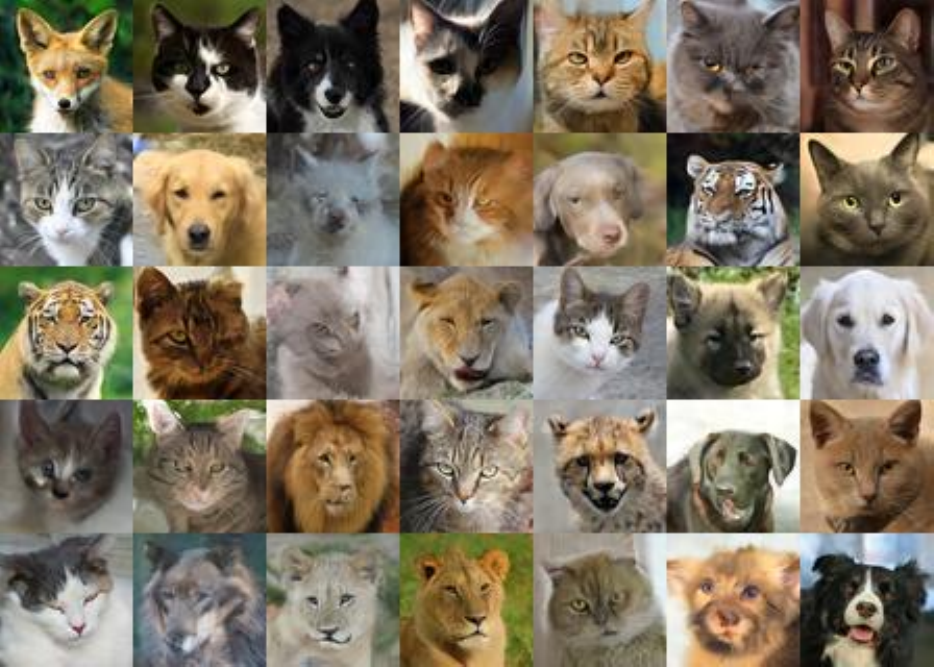}
    \label{fig:supp_afhq_5nfe_ipndm}} 
\hspace{-0.5mm}
\subfloat[\ours]{
    \includegraphics[width=0.45\linewidth]{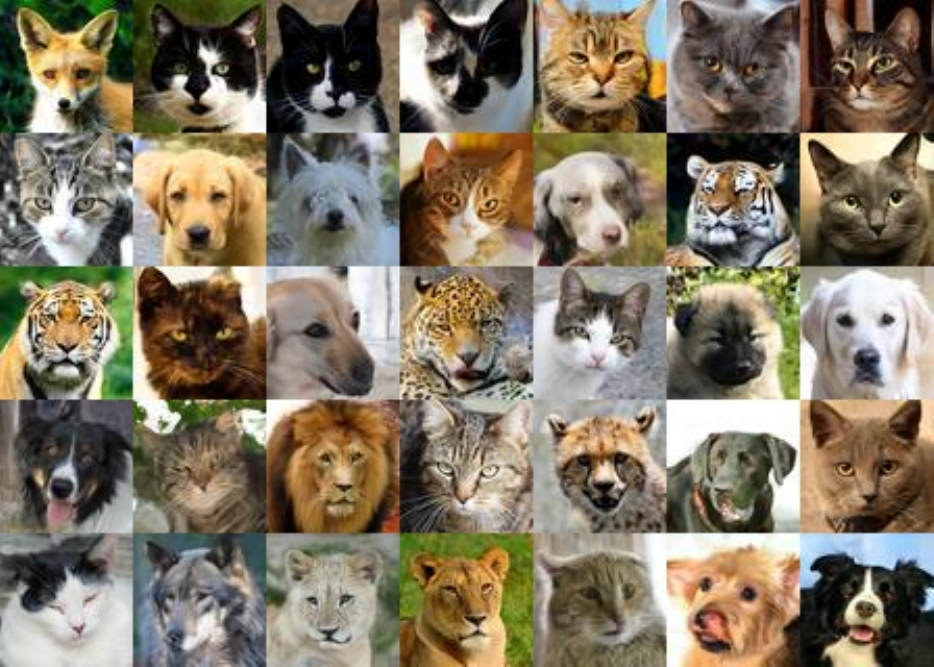}
    \label{fig:supp_afhq_5nfe_dy}
    }
    \caption{Qualitative comparison on AFHQv2 ($64 \times 64$) \cite{afhqv2} at 5 NFEs. \ours~produces animal features with higher fidelity and noticeably less blurring compared with iPNDM \cite{PNDM,iPNDM}.}
    \label{fig:supp_afhq_5nfe}

\end{figure*}

\begin{figure*}[ht]
\centering

\subfloat[iPNDM]{
    \includegraphics[width=0.45\linewidth]{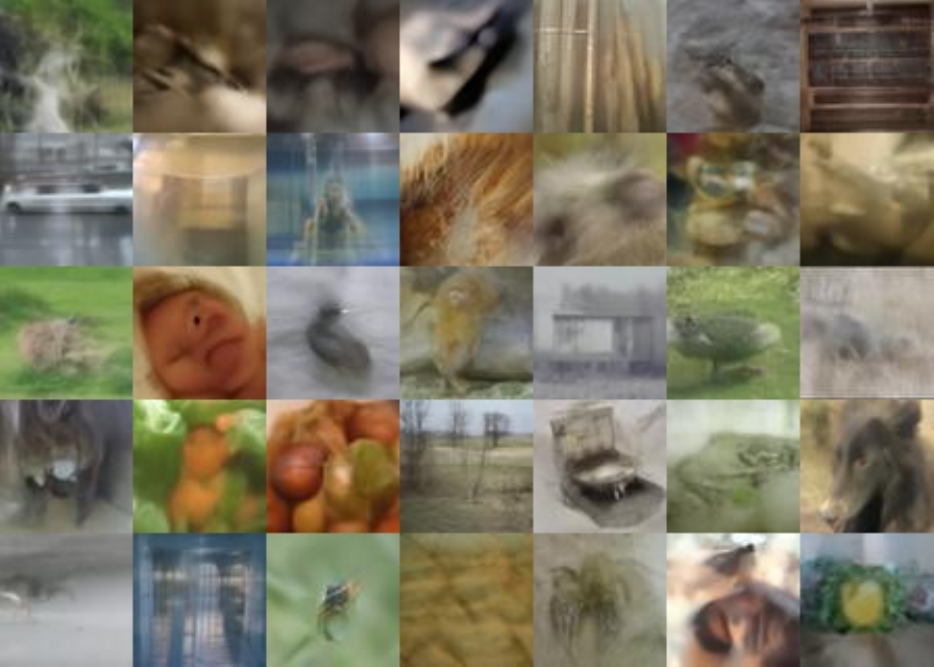}
    \label{fig:supp_imagenet_3nfe_ipndm}} 
\hspace{-0.5mm}
\subfloat[\ours]{
    \includegraphics[width=0.45\linewidth]{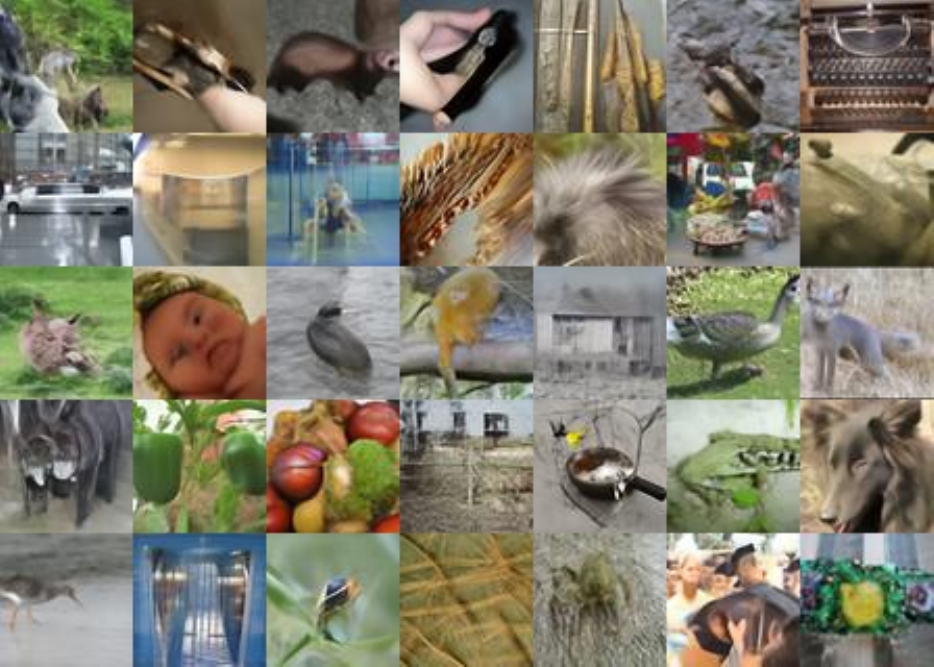}
    \label{fig:supp_imagenet_3nfe_dy}
    }
    \caption{Qualitative comparison on ImageNet ($64 \times 64$) \cite{imagenet} at 3 NFEs. \ours~generates recognizable object classes with reduced noise compared to iPNDM \cite{PNDM,iPNDM}.}
    \label{fig:supp_imagenet_3nfe}

\end{figure*}

\begin{figure*}[ht]
\centering

\subfloat[iPNDM]{
    \includegraphics[width=0.45\linewidth]{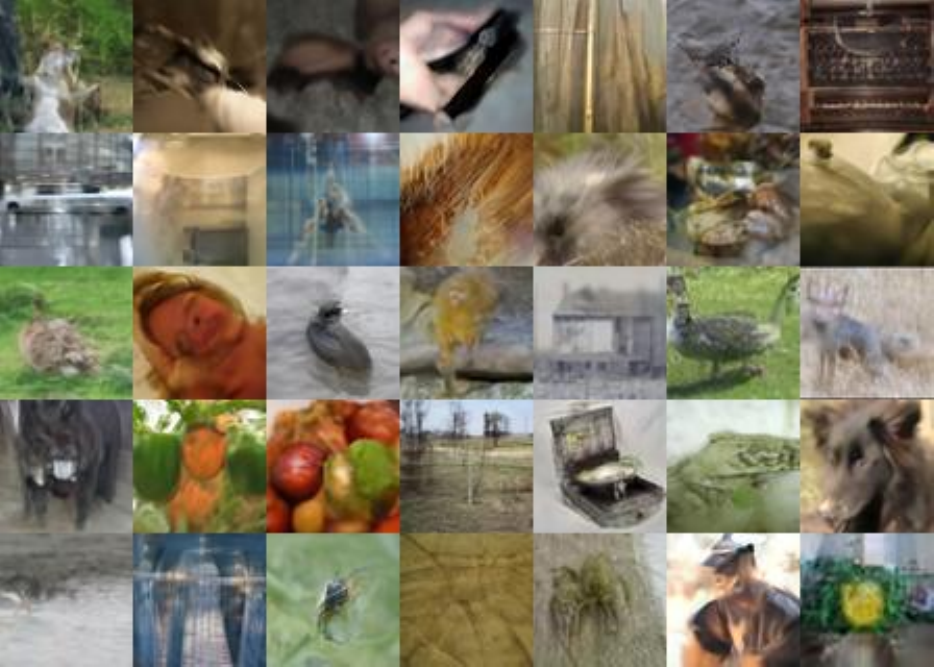}
    \label{fig:supp_imagenet_4nfe_ipndm}} 
\hspace{-0.5mm}
\subfloat[\ours]{
    \includegraphics[width=0.45\linewidth]{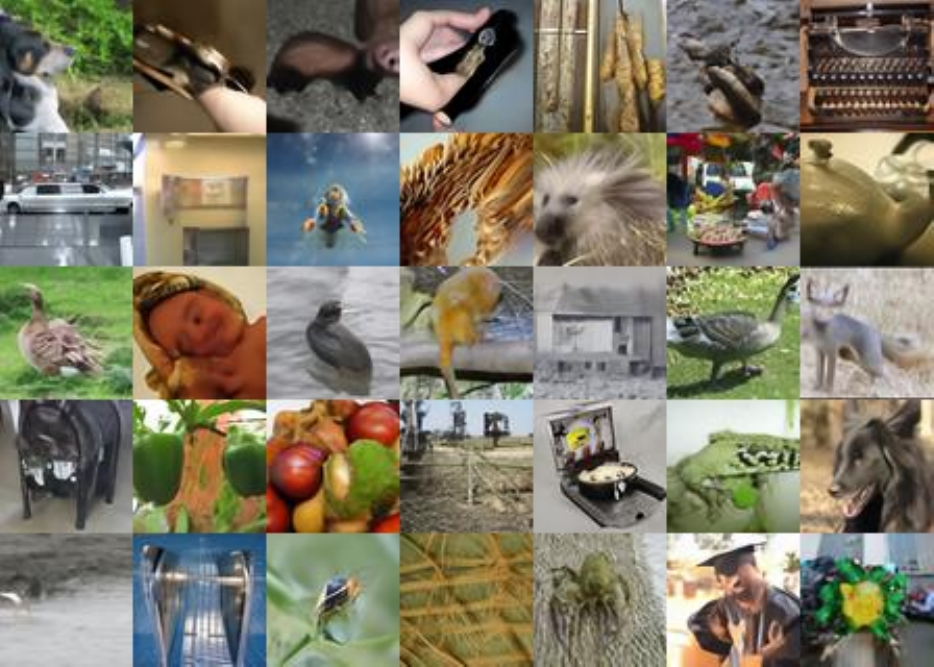}
    \label{fig:supp_imagenet_4nfe_dy}
    }
    \caption{Qualitative comparison on ImageNet ($64 \times 64$) \cite{imagenet} at 4 NFEs. \ours~generates recognizable object classes with reduced noise compared to iPNDM \cite{PNDM,iPNDM}.}
    \label{fig:supp_imagenet_4nfe}

\end{figure*}

\begin{figure*}[ht]
\centering

\subfloat[iPNDM]{
    \includegraphics[width=0.45\linewidth]{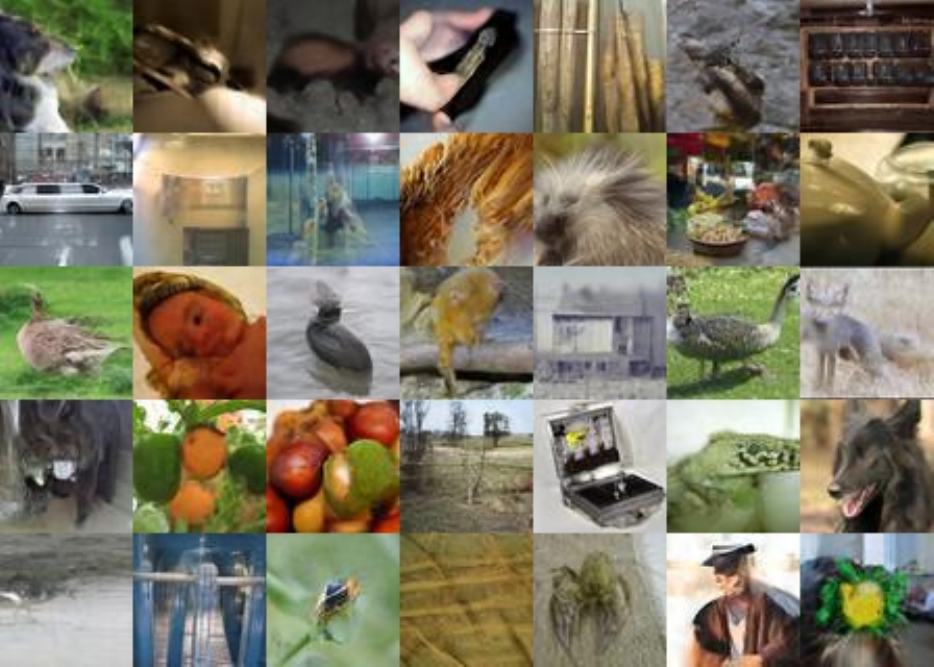}
    \label{fig:supp_imagenet_5nfe_ipndm}} 
\hspace{-0.5mm}
\subfloat[\ours]{
    \includegraphics[width=0.45\linewidth]{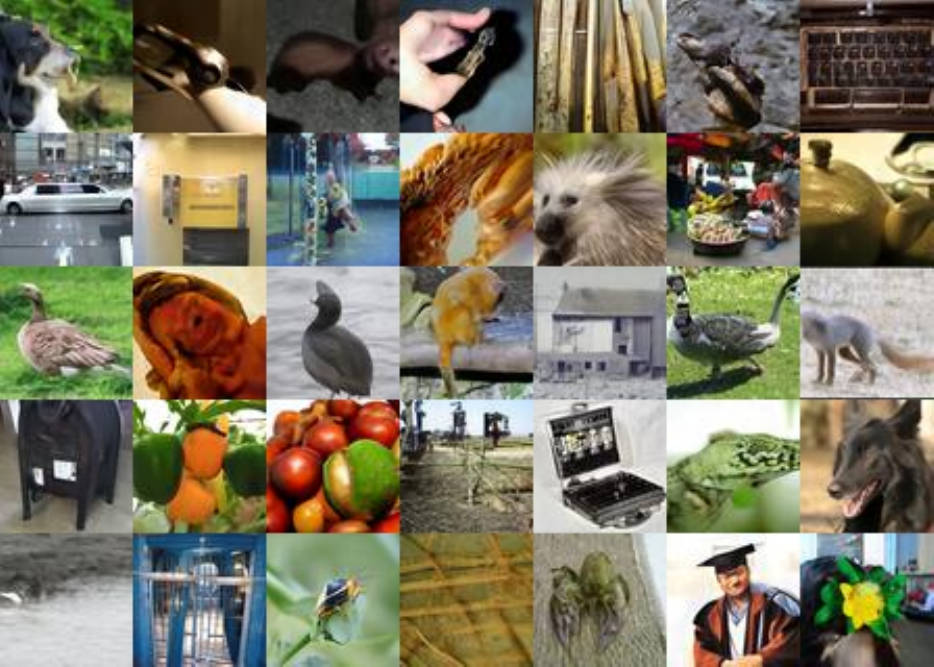}
    \label{fig:supp_imagenet_5nfe_dy}
    }
    \caption{Qualitative comparison on ImageNet ($64 \times 64$) \cite{imagenet} at 5 NFEs. \ours~generates recognizable object classes with reduced noise compared to iPNDM \cite{PNDM,iPNDM}.}
    \label{fig:supp_imagenet_5nfe}

\end{figure*}

\begin{figure*}[ht]
\centering

\subfloat[iPNDM]{
    \includegraphics[width=0.45\linewidth]{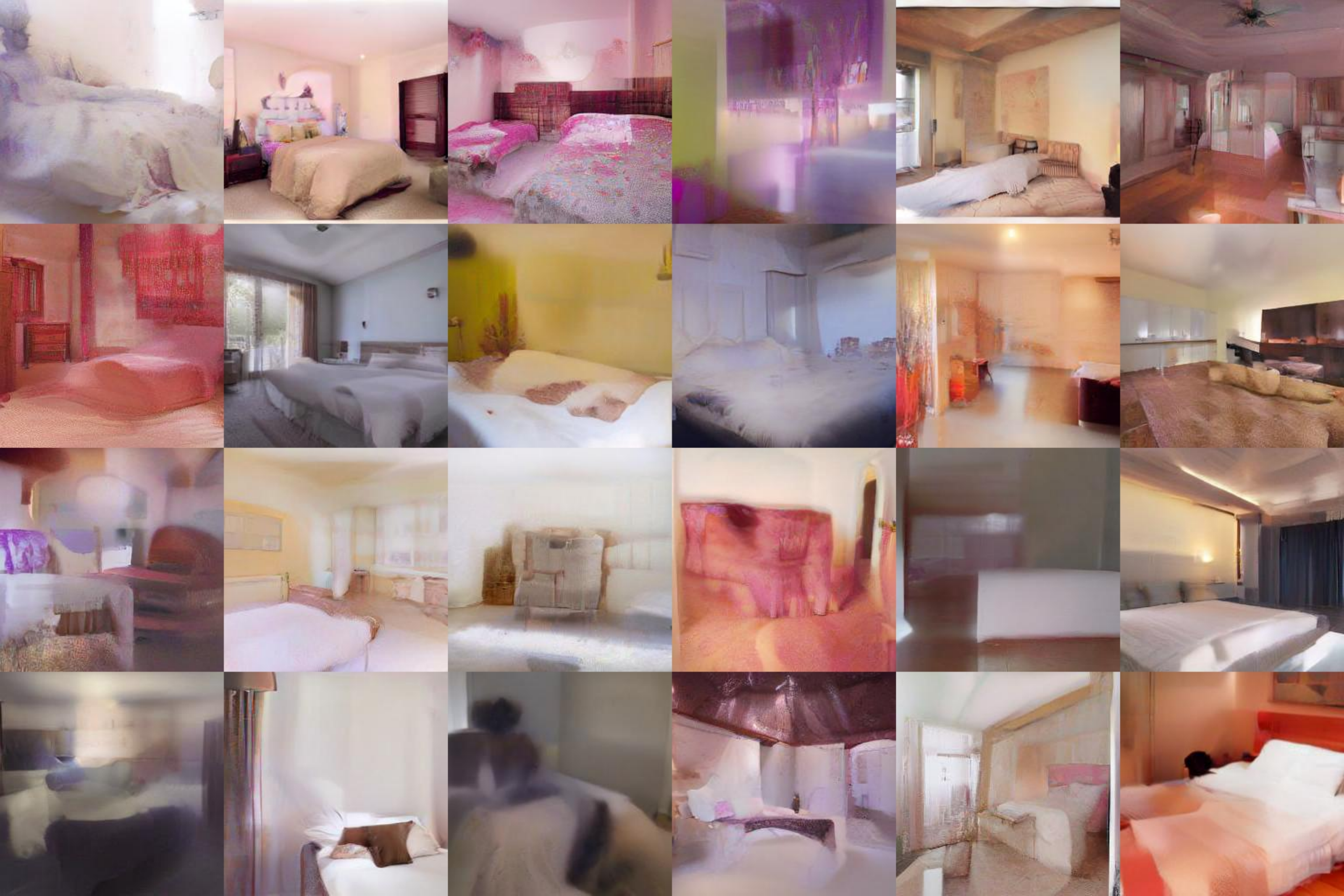}
    \label{fig:supp_lsun_3nfe_ipndm}} 
\hspace{-0.5mm}
\subfloat[\ours]{
    \includegraphics[width=0.45\linewidth]{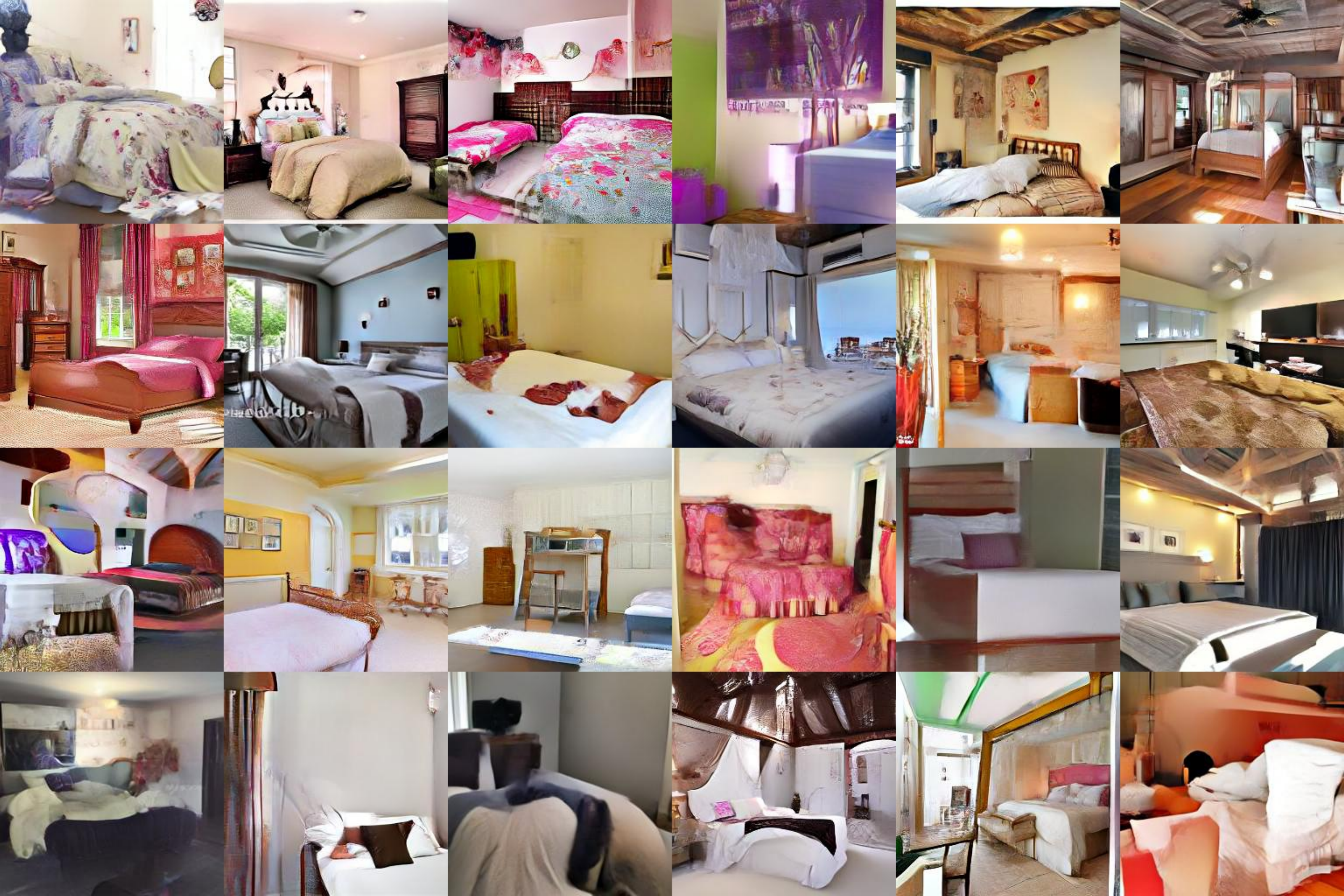}
    \label{fig:supp_lsun_3nfe_dy}
    }
    \caption{Qualitative comparison on LSUN-Bedroom ($256 \times 256$) \cite{lsun} at 3 NFEs. \ours~preserves global room layout and structure, reduces blur, and enhances object details compared with iPNDM \cite{PNDM,iPNDM}.}
    \label{fig:supp_lsun_3nfe}

\end{figure*}

\begin{figure*}[ht]
\centering

\subfloat[iPNDM]{
    \includegraphics[width=0.45\linewidth]{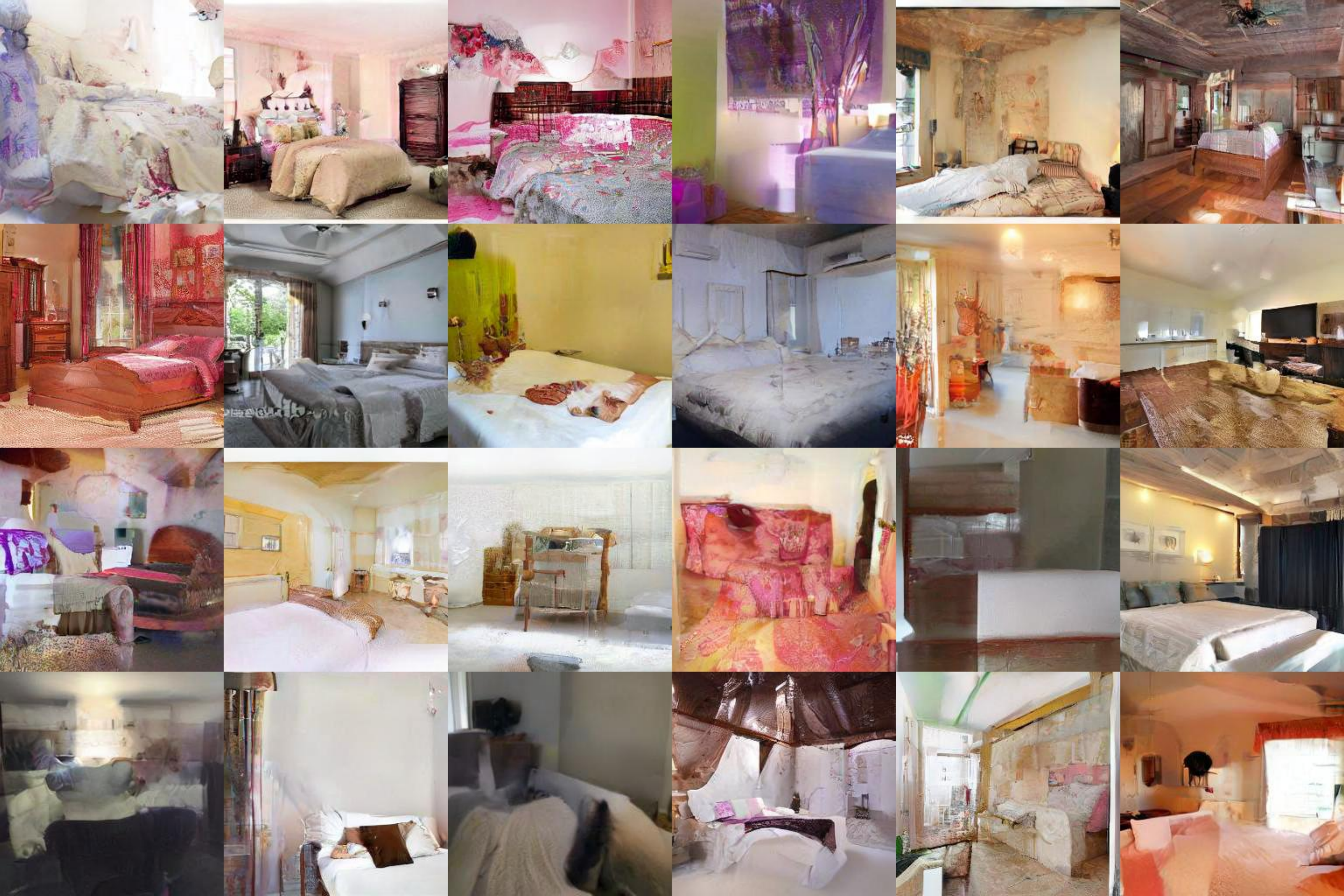}
    \label{fig:supp_lsun_4nfe_ipndm}} 
\hspace{-0.5mm}
\subfloat[\ours]{
    \includegraphics[width=0.45\linewidth]{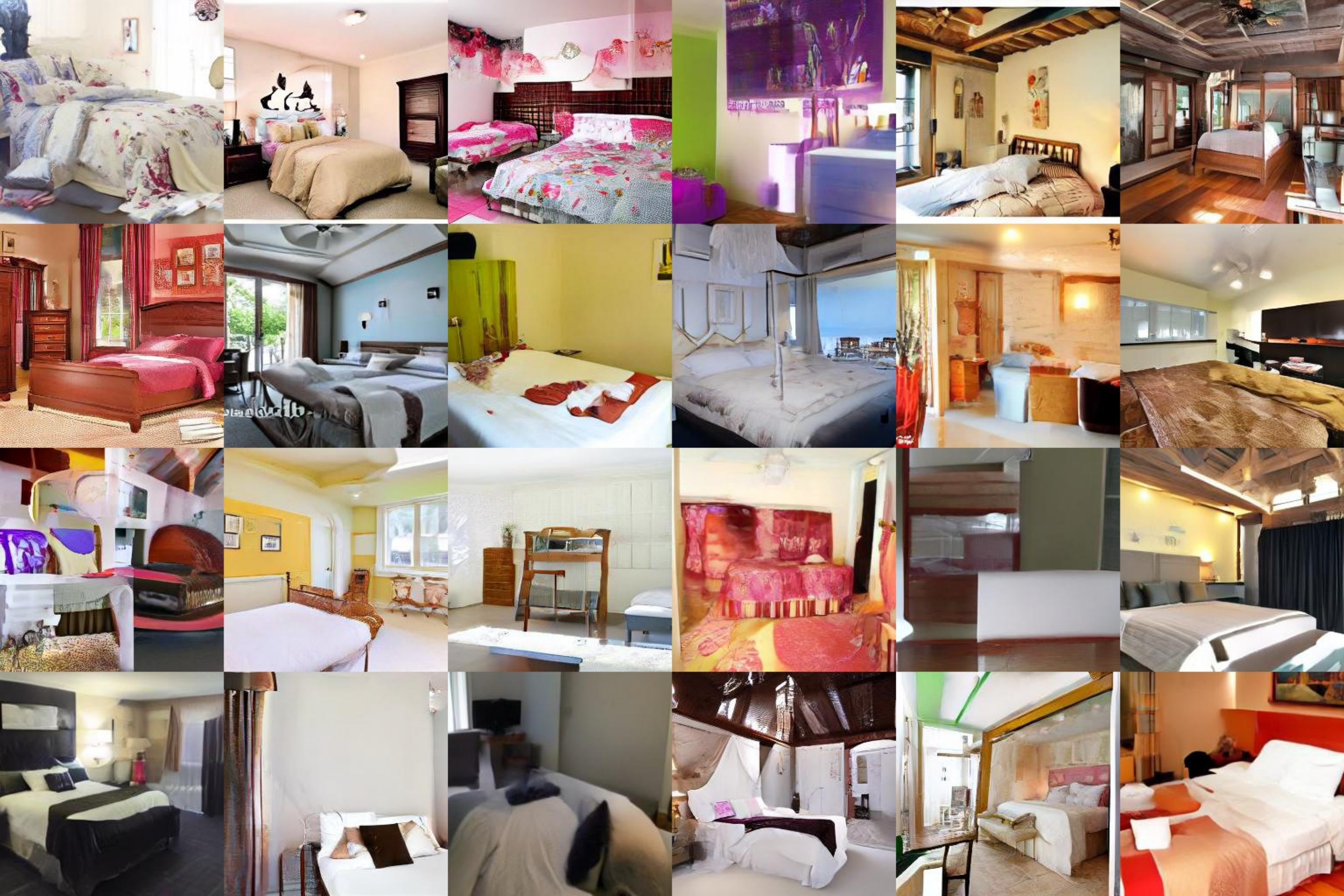}
    \label{fig:supp_lsun_4nfe_dy}
    }
    \caption{Qualitative comparison on LSUN-Bedroom ($256 \times 256$) \cite{lsun} at 4 NFEs. \ours~preserves global room layout and structure, reduces blur, and enhances object details compared with iPNDM \cite{PNDM,iPNDM}.}
    \label{fig:supp_lsun_4nfe}

\end{figure*}

\begin{figure*}[ht]
\centering

\subfloat[iPNDM]{
    \includegraphics[width=0.45\linewidth]{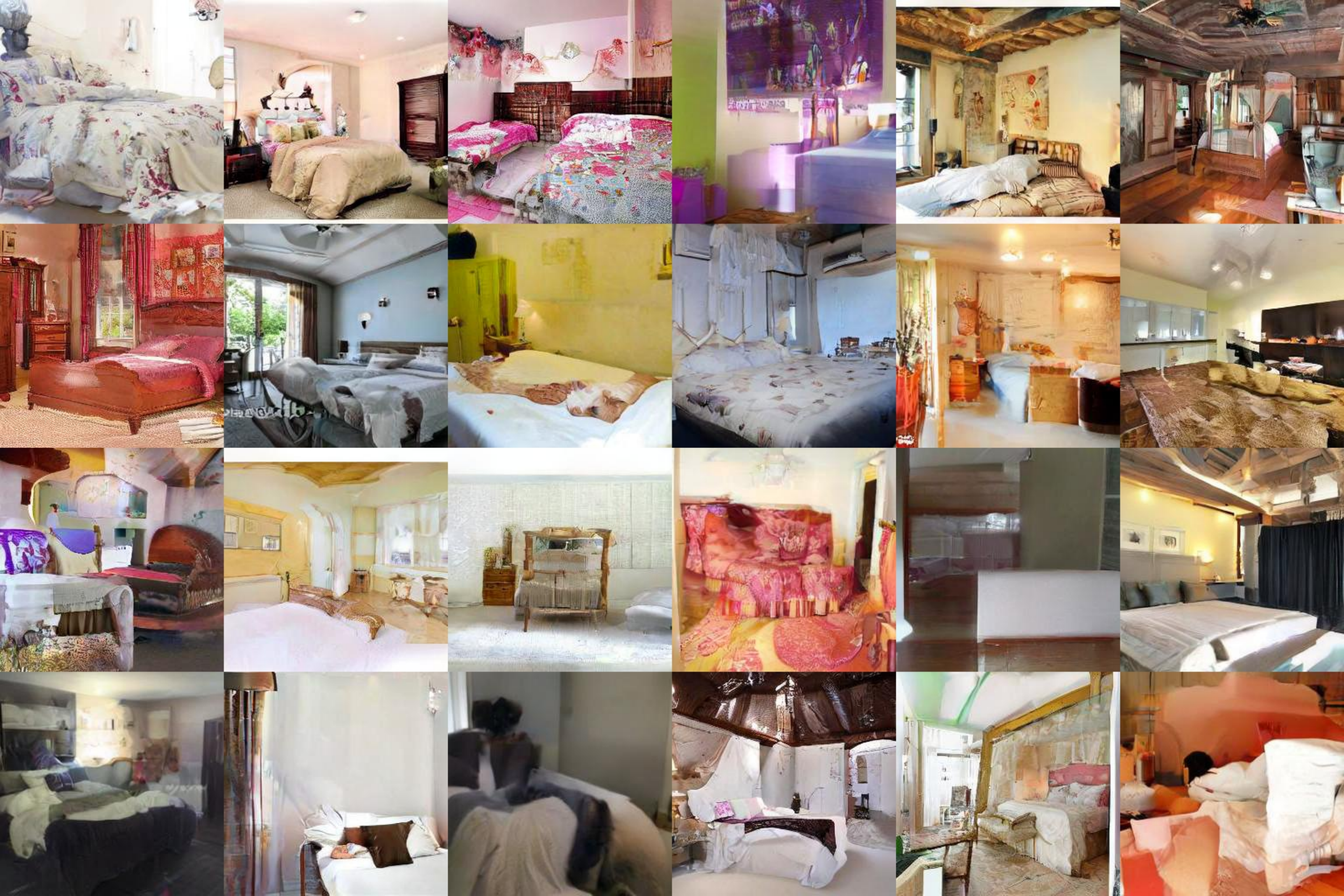}
    \label{fig:supp_lsun_5nfe_ipndm}} 
\hspace{-0.5mm}
\subfloat[\ours]{
    \includegraphics[width=0.45\linewidth]{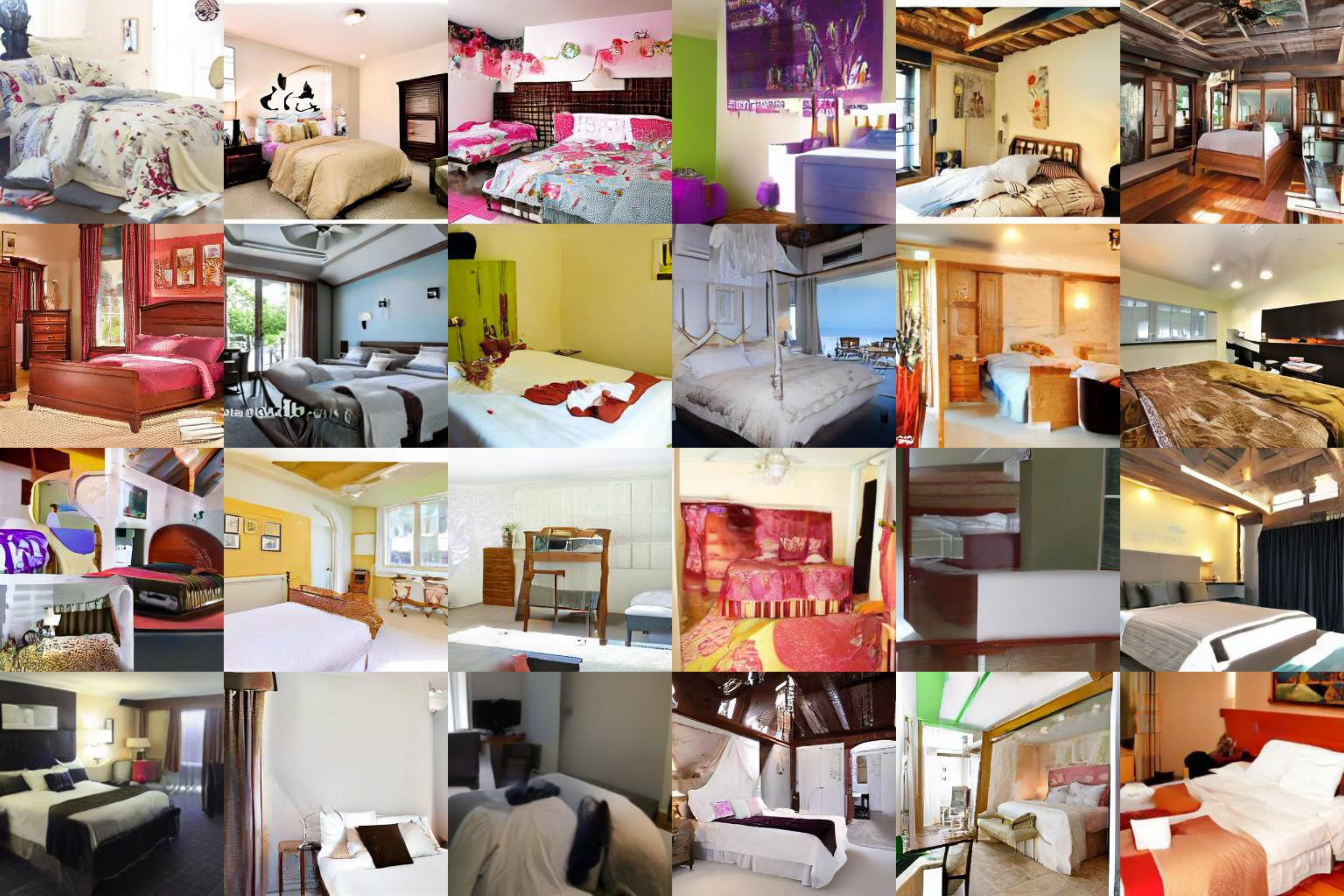}
    \label{fig:supp_lsun_5nfe_dy}
    }
    \caption{Qualitative comparison on LSUN-Bedroom ($256 \times 256$) \cite{lsun} at 5 NFEs. \ours~preserves global room layout and structure, reduces blur, and enhances object details compared with iPNDM \cite{PNDM,iPNDM}.}
    \label{fig:supp_lsun_5nfe}

\end{figure*}

\begin{figure*}[ht]
\centering

\subfloat[iPNDM]{
    \includegraphics[width=0.31\linewidth]{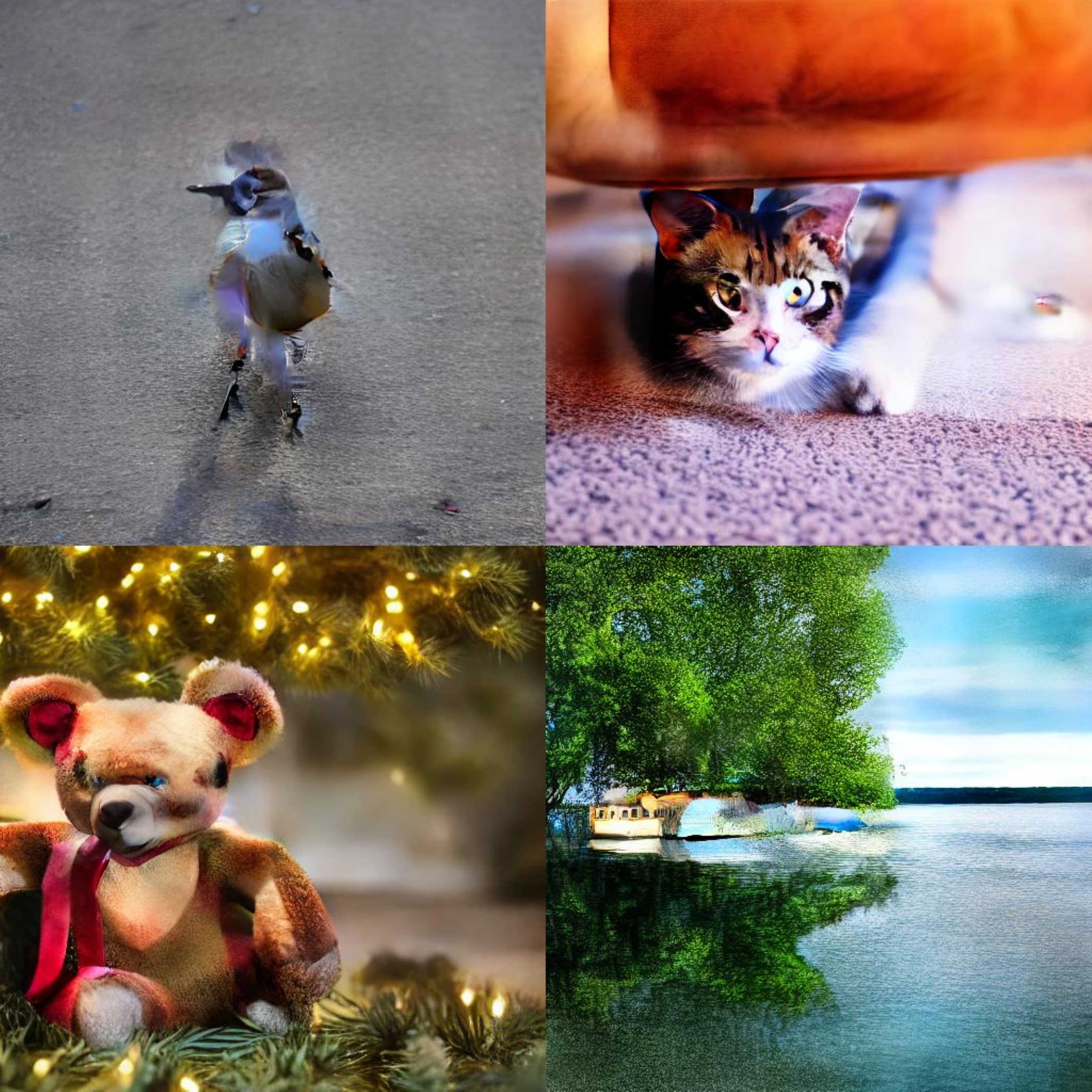}} 
\hspace{18mm}
\subfloat[\ours]{
    \includegraphics[width=0.31\linewidth]{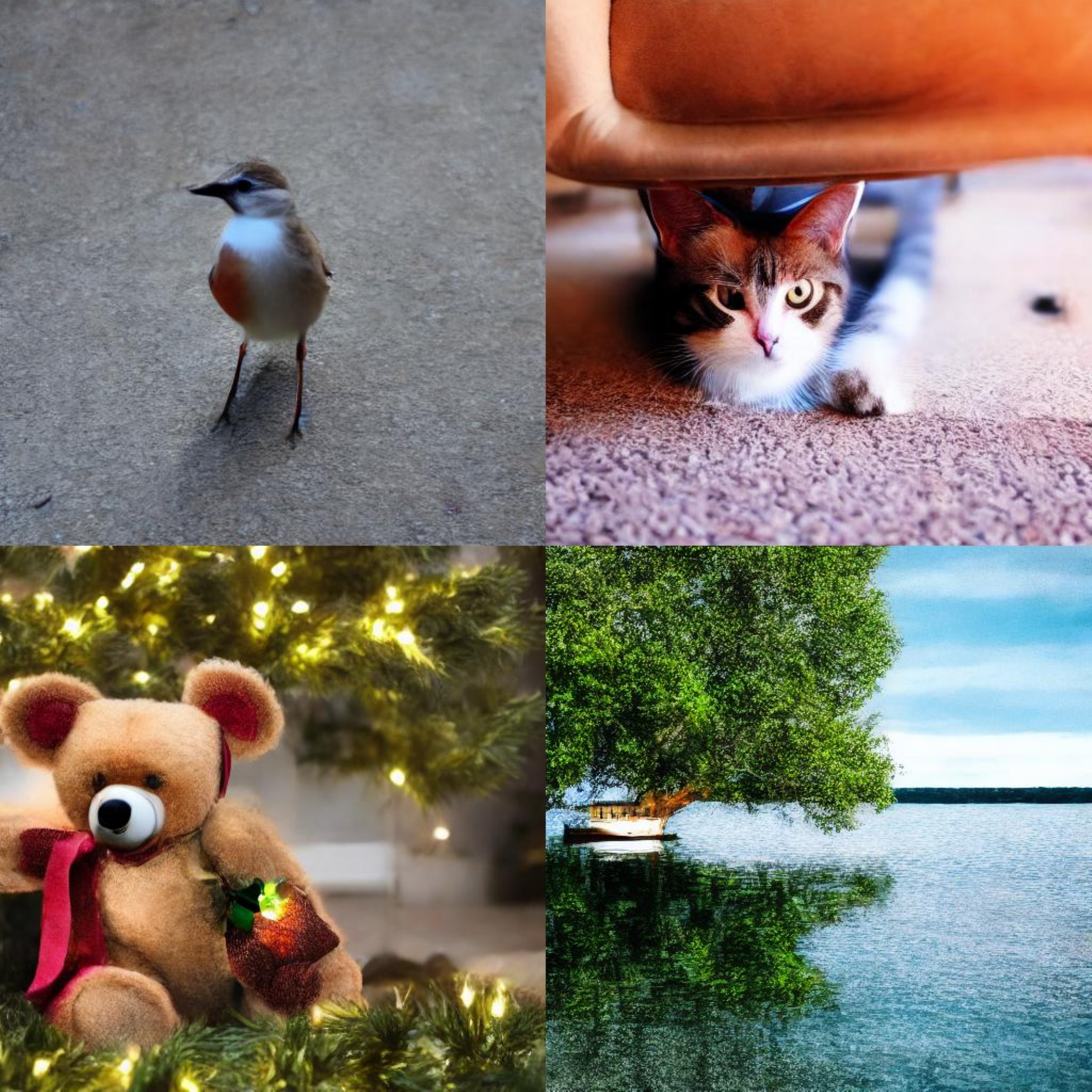}
    \label{fig:supp_mscoco_8nfe_dy}
    }
    \vspace{-1mm}
    \caption{Qualitative comparison on SD v1.5 \cite{sd1.5} (MS-COCO \cite{mscoco}) at 8 NFEs (4 Steps). Our method generates more detailed textures and coherent objects with reduced noise compared to iPNDM \cite{PNDM,iPNDM}. \textbf{Prompts}: (1) A small bird standing on top of a sidewalk. (2) a close up of a cat on the ground near a chair. (3) Teddy bear in front of a holly covered object. (4) An empty boat in the water near a tree.}
    \label{fig:supp_mscoco_8nfe}

\end{figure*}

\begin{figure*}[ht]
\centering

\subfloat[iPNDM]{
    \includegraphics[width=0.31\linewidth]{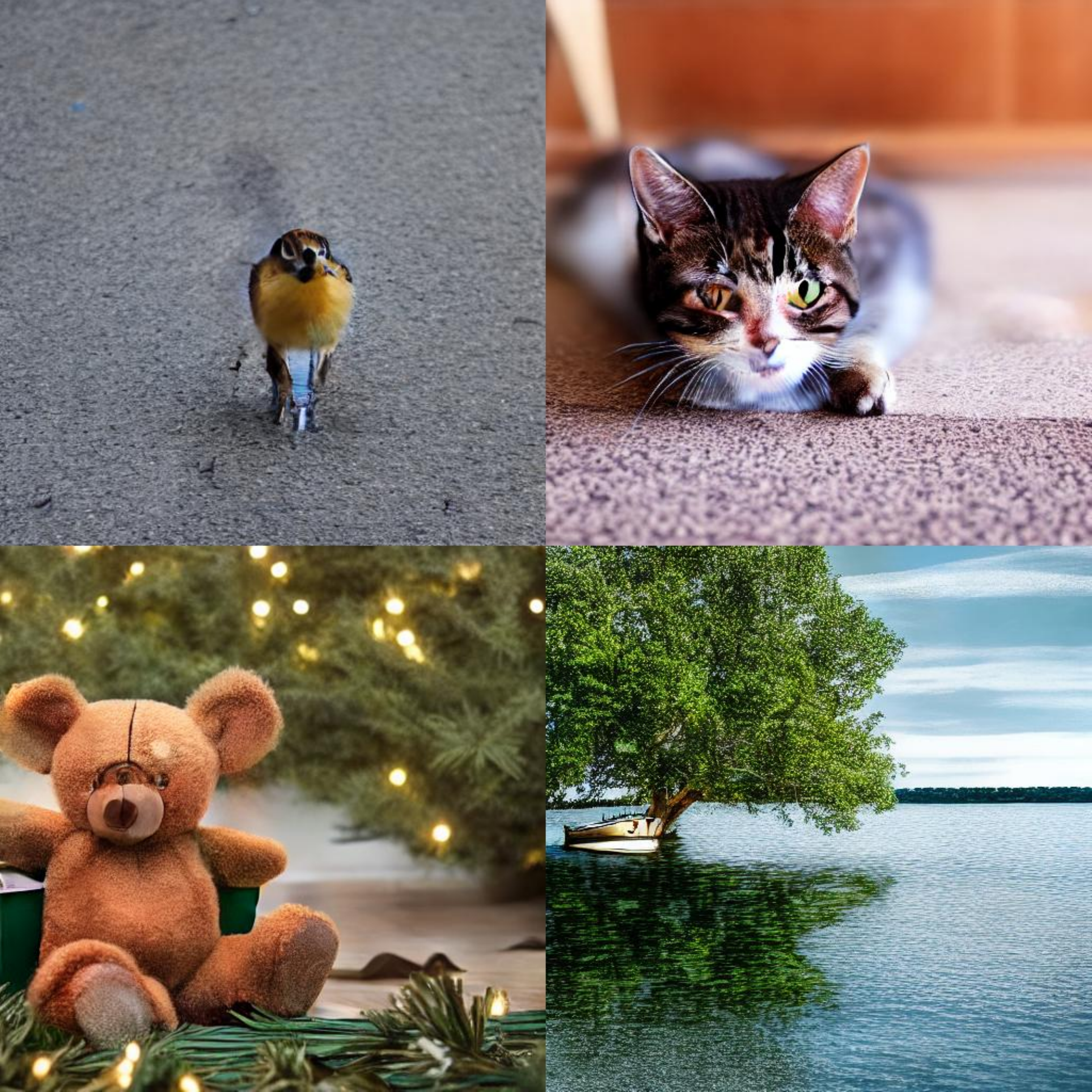}
    \label{fig:supp_mscoco_12nfe_ipndm}} 
\hspace{18mm}
\subfloat[\ours]{
    \includegraphics[width=0.31\linewidth]{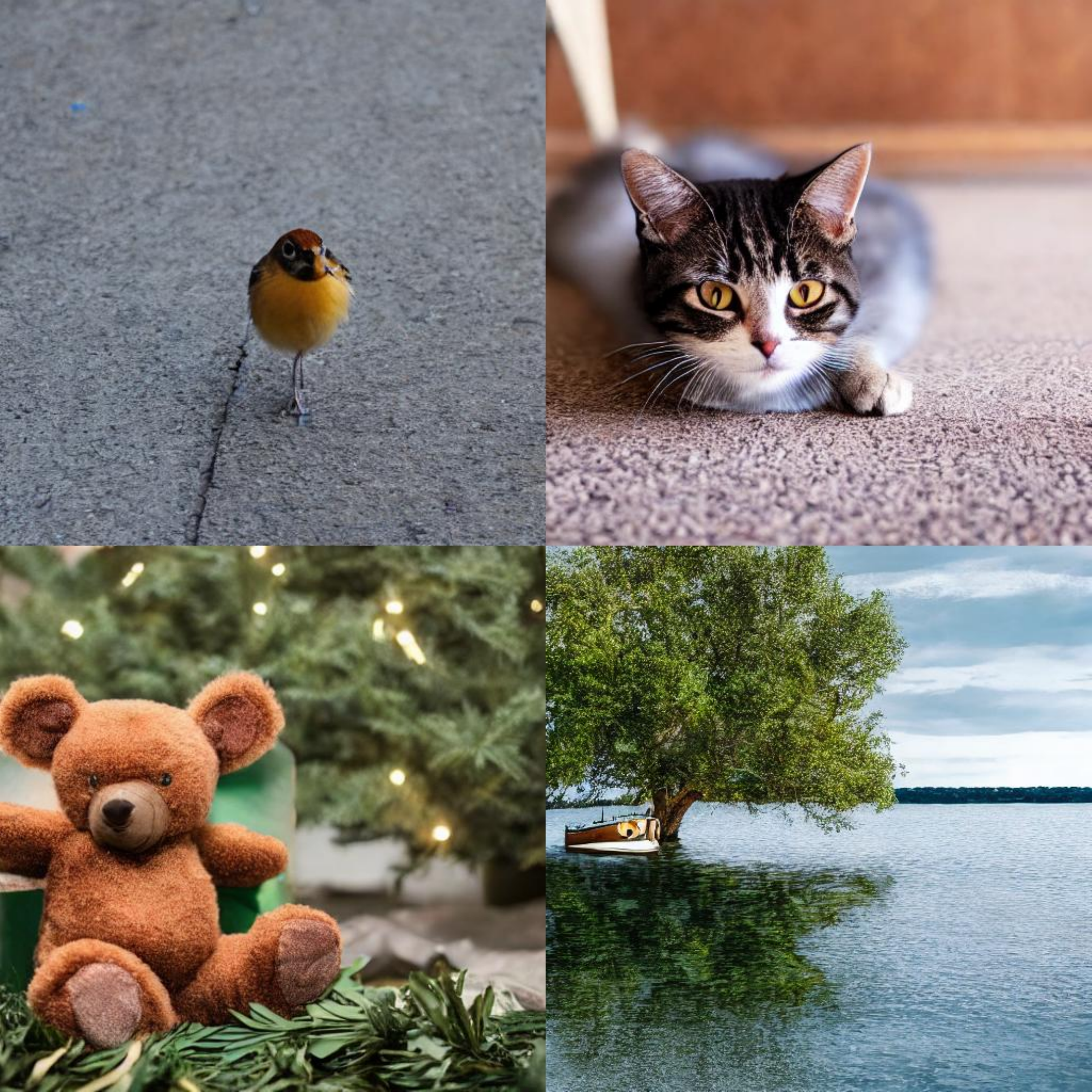}
    \label{fig:supp_mscoco_12nfe_dy}
    }
    \vspace{-1mm}
    \caption{Qualitative comparison on SD v1.5 \cite{sd1.5} (MS-COCO \cite{mscoco}) at 12 NFEs (6 Steps). Our method generates more detailed textures and coherent objects with reduced noise compared to iPNDM \cite{PNDM,iPNDM}. \textbf{Prompts}: (1) A small bird standing on top of a sidewalk. (2) a close up of a cat on the ground near a chair. (3) Teddy bear in front of a holly covered object. (4) An empty boat in the water near a tree.}
    \label{fig:supp_mscoco_12nfe}

\end{figure*}

\begin{figure*}[ht]
\centering

\subfloat[iPNDM]{
    \includegraphics[width=0.31\linewidth]{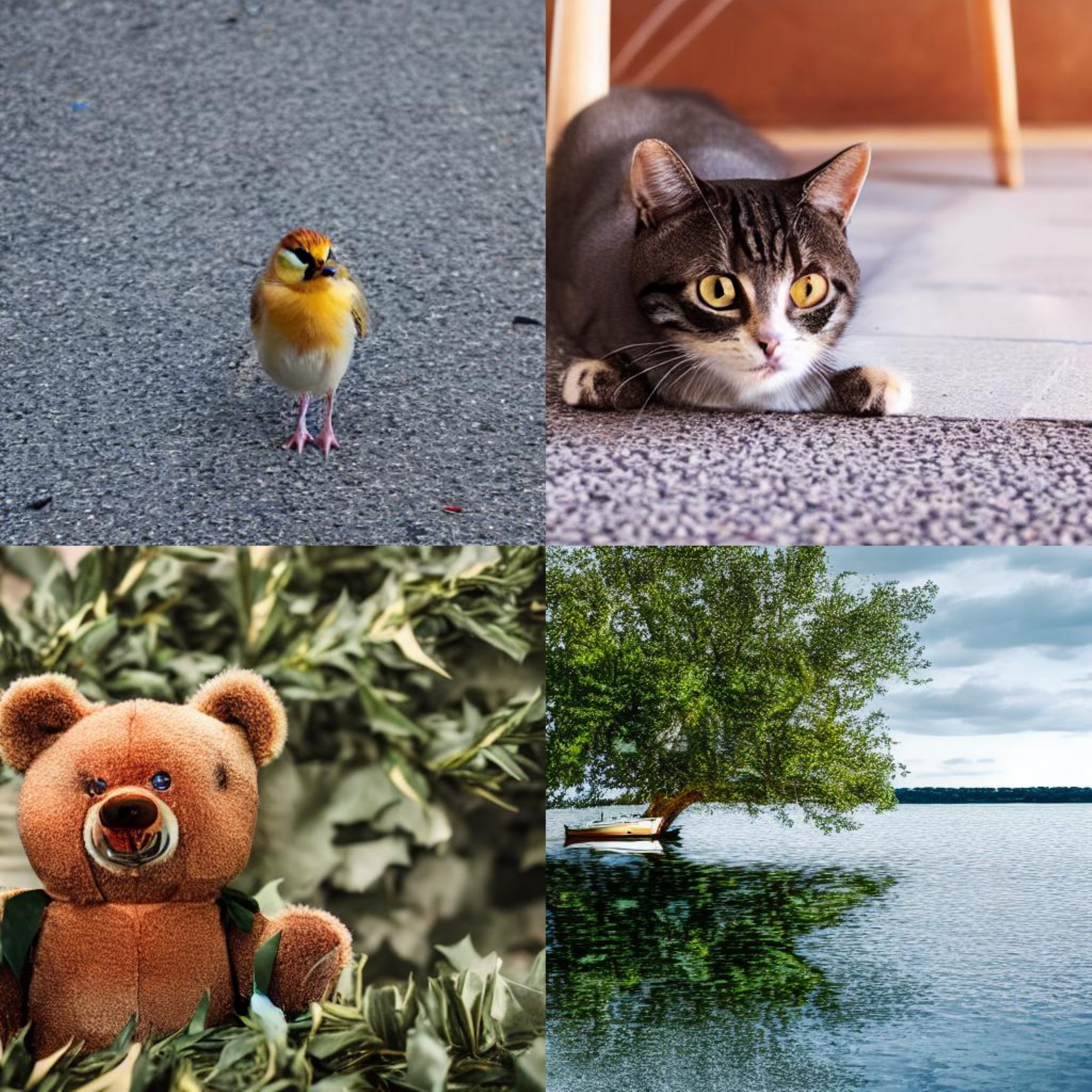}
    \label{fig:supp_mscoco_16nfe_ipndm}} 
\hspace{18mm}
\subfloat[\ours]{
    \includegraphics[width=0.31\linewidth]{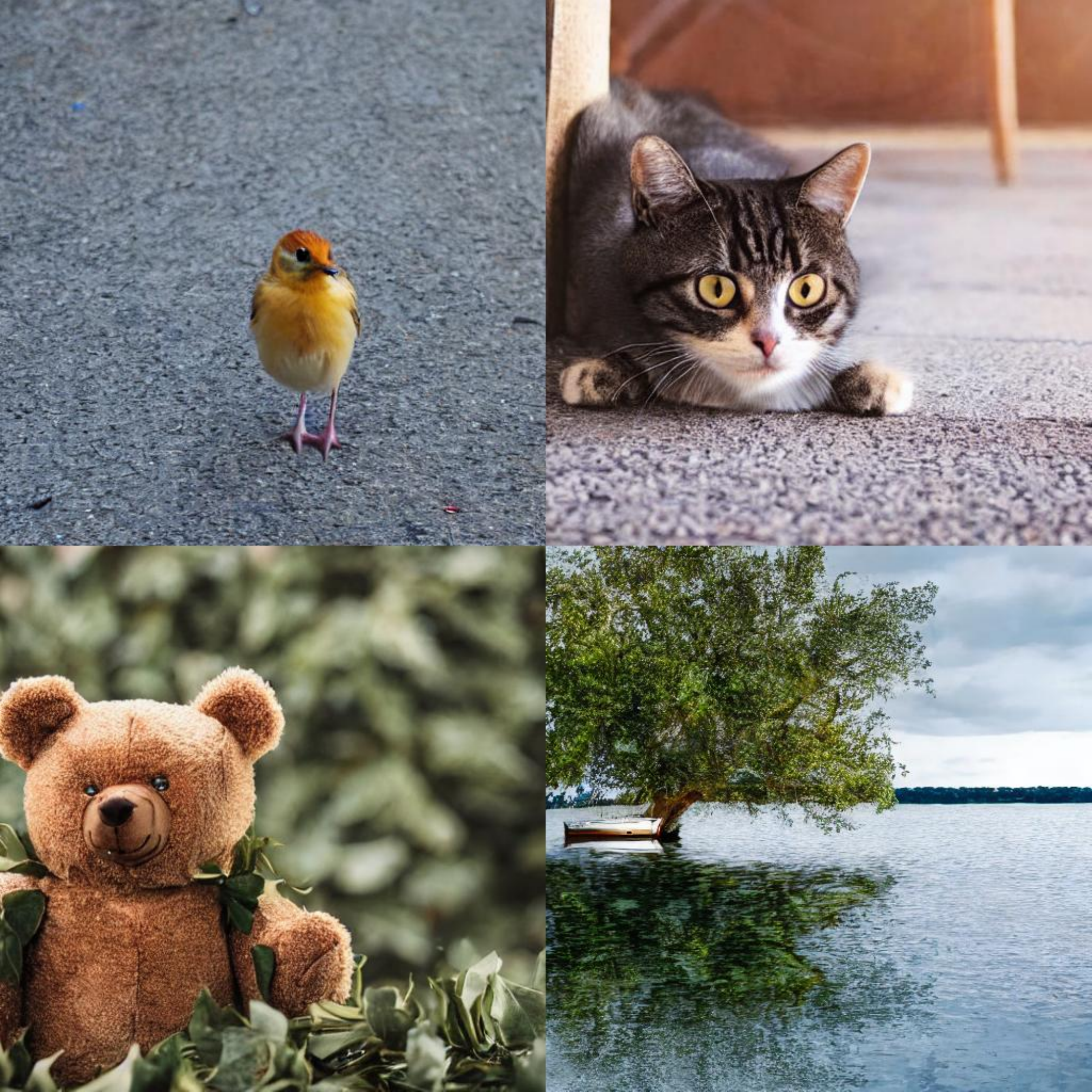}
    \label{fig:supp_mscoco_16nfe_dy}
    }
    \vspace{-1mm}
    \caption{Qualitative comparison on SD v1.5 \cite{sd1.5} (MS-COCO \cite{mscoco}) at 16 NFEs (8 Steps). Our method generates more detailed textures and coherent objects with reduced noise compared to iPNDM \cite{PNDM,iPNDM}. \textbf{Prompts}: (1) A small bird standing on top of a sidewalk. (2) a close up of a cat on the ground near a chair. (3) Teddy bear in front of a holly covered object. (4) An empty boat in the water near a tree.}
    \label{fig:supp_mscoco_16nfe}

\end{figure*}

\begin{figure*}
\centering
    \includegraphics[width=\linewidth]{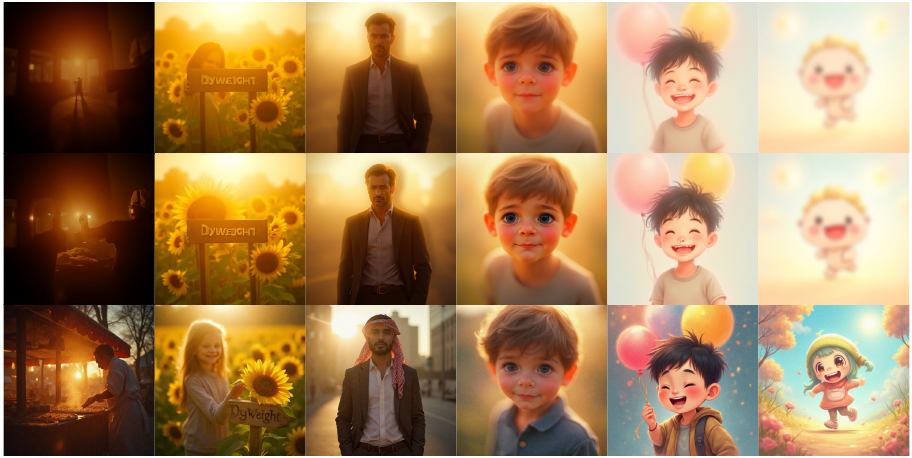}
    \caption{Qualitative comparison on FLUX.1-dev \cite{flux,flux_paper} at 5 NFEs. From top to bottom, the rows show results from DPM-Solver++(2M) \cite{dpmpp}, iPNDM(2M) \cite{PNDM,iPNDM}, and our \ours. Our method delivers superior visual fidelity, prompt alignment, and structural coherence.}
    \label{fig:supp_flux_5_1}
\end{figure*}

\begin{figure*}
\centering
    \includegraphics[width=\linewidth]{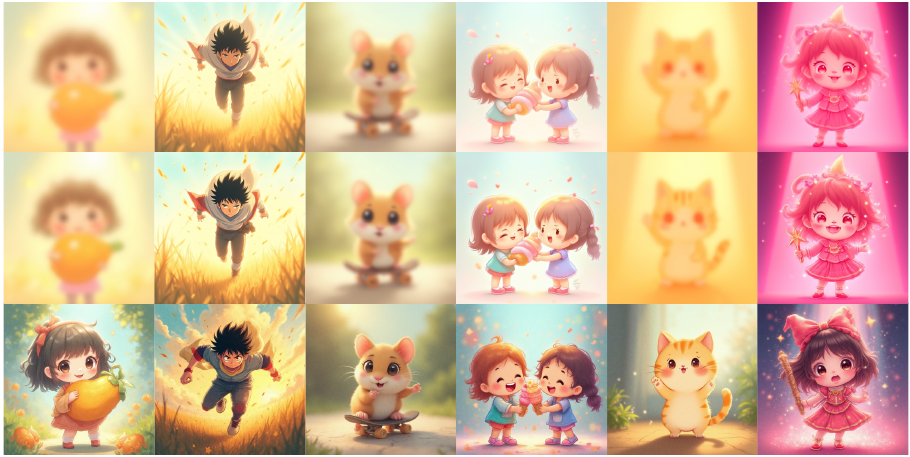}
    \caption{Qualitative comparison on FLUX.1-dev \cite{flux,flux_paper} at 5 NFEs. From top to bottom, the rows show results from DPM-Solver++(2M) \cite{dpmpp}, iPNDM(2M) \cite{PNDM,iPNDM}, and our \ours. Our method delivers superior visual fidelity, prompt alignment, and structural coherence.}
    \label{fig:supp_flux_5_2}
\end{figure*}

\begin{figure*}
\centering
    \includegraphics[width=\linewidth]{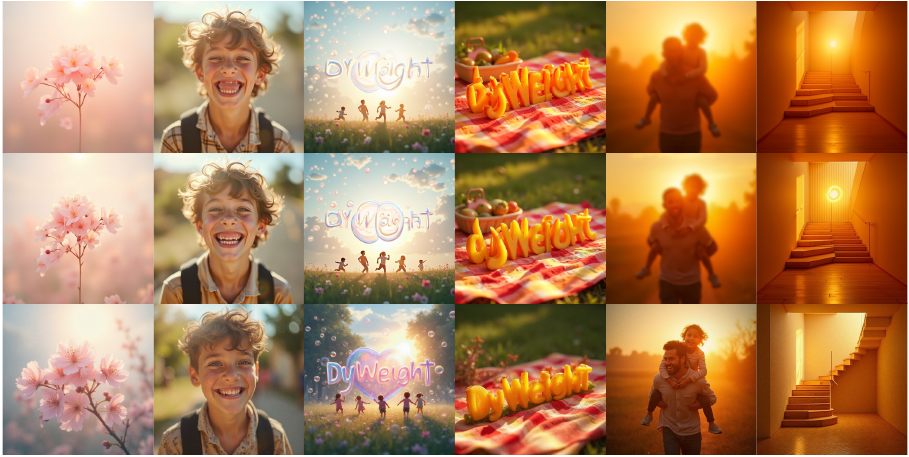}
    \caption{Qualitative comparison on FLUX.1-dev \cite{flux,flux_paper} at 7 NFEs. From top to bottom, the rows show results from DPM-Solver++(2M) \cite{dpmpp}, iPNDM(2M) \cite{PNDM,iPNDM}, and our \ours. Our method delivers superior visual fidelity, prompt alignment, and structural coherence.}
    \label{fig:supp_flux_7_1}
\end{figure*}

\begin{figure*}
\centering
    \includegraphics[width=\linewidth]{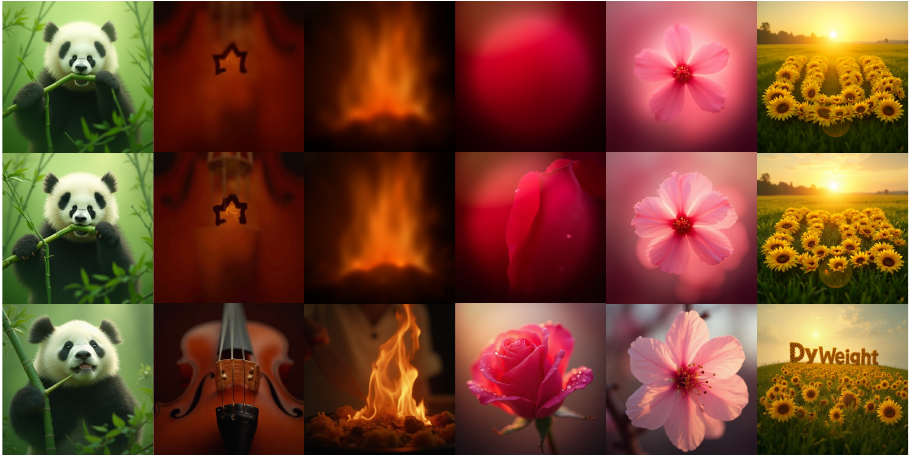}
    \caption{Qualitative comparison on FLUX.1-dev \cite{flux,flux_paper} at 7 NFEs. From top to bottom, the rows show results from DPM-Solver++(2M) \cite{dpmpp}, iPNDM(2M) \cite{PNDM,iPNDM}, and our \ours. Our method delivers superior visual fidelity, prompt alignment, and structural coherence.}
    \label{fig:supp_flux_7_2}
\end{figure*}

\begin{figure*}
\centering
    \includegraphics[width=0.85\linewidth]{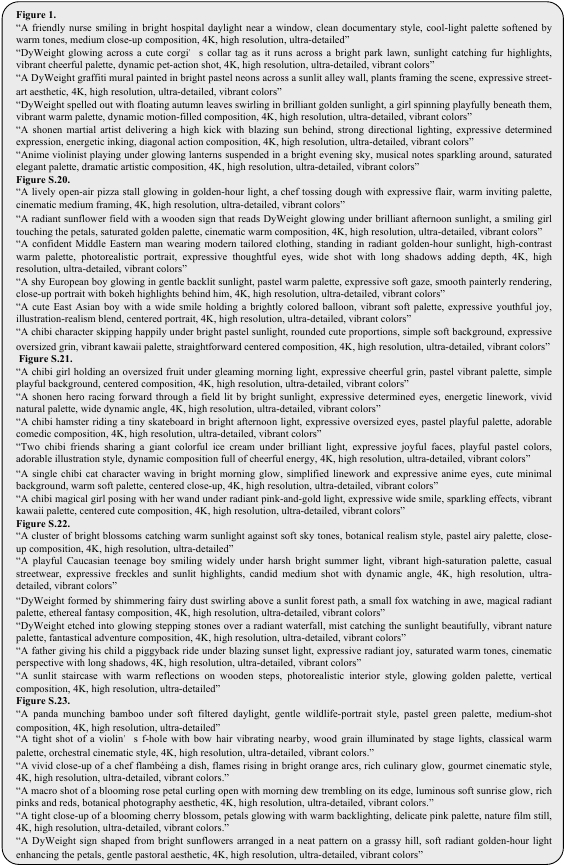}
    \vspace{-2mm}
    \caption{Comprehensive list of text prompts employed for the FLUX.1-dev \cite{flux,flux_paper} qualitative comparisons.}
    \label{fig:supp_flux_prompt}
\end{figure*}

In this section, we present extensive qualitative comparisons between \ours~and the baseline solver iPNDM~\cite{PNDM,iPNDM} across diverse benchmarks.
\cref{fig:supp_cifar_3nfe} -- \cref{fig:supp_afhq_5nfe} provide unconditional generation results on pixel-space datasets, including CIFAR-10~\cite{cifar}, FFHQ~\cite{ffhq}, and AFHQv2~\cite{afhqv2} at 3, 4, and 5 NFEs. \cref{fig:supp_imagenet_3nfe} -- \cref{fig:supp_imagenet_5nfe} demonstrate conditional generation results on ImageNet ($64 \times 64$) \cite{imagenet} in pixel space. \cref{fig:supp_lsun_3nfe} -- \cref{fig:supp_lsun_5nfe} display unconditional generation results on LSUN-Bedroom~\cite{lsun} in latent space. 
\cref{fig:supp_mscoco_8nfe} -- \cref{fig:supp_mscoco_16nfe} show text-to-image generation results using Stable Diffusion v1.5~\cite{sd1.5} on MS-COCO~\cite{mscoco} prompts at varying NFEs. 
\cref{fig:supp_flux_5_1} -- \cref{fig:supp_flux_7_2} present qualitative comparisons on the \textbf{state-of-the-art} text-to-image model FLUX.1-dev~\cite{flux,flux_paper} at 5 and 7 NFEs.
Finally, \cref{fig:supp_flux_prompt} lists the specific text prompts used for the FLUX.1-dev~\cite{flux,flux_paper} visualizations presented in \cref{fig:teaser} in the main paper and \cref{sec:supp_qualitative_results}.

\section{Limitation and future work}
Like other advanced learning-based solvers, DyWeight is fundamentally bounded by the local linearization assumption of ODE trajectories, making extreme one- or two-step generation (\eg, 1-2 NFEs) highly challenging without introducing perceptual artifacts. Furthermore, our current parameterization is coupled with predefined, static sampling budgets. Future work will explore integrating DyWeight with score-distillation paradigms to breach the 2-NFE barrier, alongside developing an NFE-agnostic learning mechanism to seamlessly generalize the learned weights across arbitrary, dynamic integration steps.

\clearpage

\begin{table*}[t]
\caption{Image generation FID$\downarrow$~\cite{fid} on CIFAR-10($32\times32$)~\cite{cifar}. The best results are in \textbf{bold}, the second best are \underline{underlined}.}
\label{tab:supp_cifar10}
\centering
\fontsize{12}{14}\selectfont
\setlength{\tabcolsep}{9pt}

\begin{tabular}{lccccccc}
\toprule
\multirow{2}{*}{Method} & \multicolumn{7}{c}{NFE} \\
\cmidrule{2-8}
 & 3 & 4 & 5 & 6 & 7 & 8 & 9 \\
\midrule
iPNDM \cite{iPNDM,PNDM}        & 24.55 & 13.92 &  7.77 &  5.07 & 4.04 & 3.22 & 2.83 \\
DPM-Solver++(3M) \cite{dpmpp}  & 55.76 & 22.41 &  9.94 &  5.97 & 4.29 & 3.37 & 2.99 \\
UniPC \cite{UniPC}             &109.60 & 45.20 & 23.98 & 11.14 & 5.83 & 3.99 & 3.21 \\
\midrule
AMED-Solver \cite{AMED}        & 18.49 & 17.18 &  7.59 &  7.04 & 4.36 & 5.56 & 3.67 \\ 
AMED-Plugin \cite{AMED}        & 10.81 &   -   &  6.61 &   -   & 3.65 &  -   & 2.63 \\ 
EPD-Solver \cite{EPD}          & \underline{10.40} &   -   &  4.33 &   -   & 2.82 &  -   & 2.49 \\
EPD-Plugin \cite{EPD}          & 10.54 &   -   &  4.47 &   -   & 3.27 &  -   & 2.42 \\
LD3 \cite{LD3}                 & 16.52 &  9.31 &  6.39 &  3.35 & 2.98 & 2.81 & 2.51 \\
DLMS \cite{DLMS}               &   -   &  \underline{4.52} &  \underline{3.23} &  2.81 & 2.53 & 2.43 & 2.37 \\ 
S4S-Alt \cite{s4s}             &   16.95   &  6.35 &  3.73 &  \underline{2.67} & \underline{2.52} & \underline{2.39} & \underline{2.31} \\ 
\rowcolor{lightCyan}
DyWeight (ours)                &  \textbf{8.16} &  \textbf{3.92} &  \textbf{3.02} &  \textbf{2.61} & \textbf{2.40} & \textbf{2.18} & \textbf{2.13} \\
\bottomrule
\end{tabular}
\end{table*}

\begin{table*}[t]
\caption{Image generation FID$\downarrow$~\cite{fid} on FFHQ($64\times64$)~\cite{ffhq}. The best results are in \textbf{bold}, the second best are \underline{underlined}.}
\label{tab:supp_ffhq}
\centering
\fontsize{12}{14}\selectfont
\setlength{\tabcolsep}{9pt}
\begin{tabular}{lccccccc}
\toprule
\multirow{2}{*}{Method} & \multicolumn{7}{c}{NFE} \\
\cmidrule{2-8}
 & 3 & 4 & 5 & 6 & 7 & 8 & 9 \\
\midrule
iPNDM \cite{iPNDM,PNDM}        & 27.72 & 20.07 & 13.80 &  8.61 & 7.16 & 5.43 & 4.98 \\
DPM-Solver++(3M) \cite{dpmpp}  & 66.07 & 30.06 & 13.47 &  8.25 & 6.20 & 5.20 & 4.77 \\
UniPC \cite{UniPC}             & 86.43 & 44.78 & 21.40 & 12.85 & 7.44 & 5.50 & 4.47 \\
\midrule
AMED-Solver \cite{AMED}        & 47.31 & 26.89 & 14.80 &  9.97 & 8,82 & 7.86 & 6.31 \\ 
AMED-Plugin \cite{AMED}        & 26.87 &   -   & 12.49 &   -   & 6.64 &   -  & 4.24 \\ 
EPD-Solver \cite{EPD}          & 21.74 &   -   &  7.84 &   -   & 4.81 &   -  & 3.82 \\
EPD-Plugin \cite{EPD}          & \underline{19.02} &   -   &  7.97 &   -   & 5.09 &   -  & 3.53 \\
LD3 \cite{LD3}                 & 23.86 & 17.96 & 10.36 &  5.97 & 4.38 & 3.50 & \underline{2.94} \\
DLMS \cite{DLMS}               &   -   &  \underline{9.63} &  6.85 &  5.82 & 5.16 & 4.81 & 4.23 \\ 
S4S-Alt \cite{s4s}               &   19.86   & 10.63 & \underline{6.25} &  \underline{4.62} & \underline{3.45} & \underline{3.15} & 3.00 \\ 
\rowcolor{lightCyan}
DyWeight (ours)                & \textbf{16.78} &  \textbf{9.17} &  \textbf{5.85} &  \textbf{3.93} & \textbf{3.39} & \textbf{2.89} & \textbf{2.77} \\
\bottomrule
\end{tabular}
\end{table*}

\begin{table*}[t]
\caption{Image generation FID$\downarrow$~\cite{fid} on AFHQv2($64\times64$)~\cite{afhqv2}. The best results are in \textbf{bold}, the second best are \underline{underlined}.}
\label{tab:supp_afhq}
\centering
\fontsize{12}{14}\selectfont
\setlength{\tabcolsep}{9pt}
\begin{tabular}{lccccccc}
\toprule
\multirow{2}{*}{Method} & \multicolumn{7}{c}{NFE} \\
\cmidrule{2-8}
 & 3 & 4 & 5 & 6 & 7 & 8 & 9 \\
\midrule
iPNDM \cite{iPNDM,PNDM}        & 15.53 &  8.73 &  5.58 & 3.81 & 3.19 & 2.72 & 2.48 \\
DPM-Solver++(3M) \cite{dpmpp}  & 35.05 & 21.04 & 10.63 & 6.24 & 4.47 & 3.86 & 3.44 \\
UniPC \cite{UniPC}             & 60.89 & 33.78 & 13.01 & 8.27 & 5.07 & 4.60 & 4.46 \\
\midrule
AMED-Solver \cite{AMED}        & 31.82 & 18.99 &  7.34 & 8.19 & 4.39 & 5.55 & 3.53 \\ 
LD3 \cite{LD3}                 & 17.94 &  9.96 &  6.09 & 3.63 & 2.97 & 2.63 & 2.27 \\ 
S4S-Alt \cite{s4s}             &   \underline{14.71}   & \underline{6.52} & \underline{3.89} &  \underline{2.70} & \underline{2.56} & \underline{2.29} & \underline{2.18} \\ 
\rowcolor{lightCyan}
DyWeight (ours)                &  \textbf{9.16} &  \textbf{5.38} &  \textbf{3.20} & \textbf{2.66} & \textbf{2.42} & \textbf{2.23} & \textbf{2.13} \\
\bottomrule
\end{tabular}
\end{table*}

\begin{table*}[t]
\caption{Image generation FID$\downarrow$~\cite{fid} on ImageNet($64\times64$)~\cite{imagenet}. The best results are in \textbf{bold}, the second best are \underline{underlined}.}
\label{tab:supp_imagenet}
\centering
\fontsize{12}{14}\selectfont
\setlength{\tabcolsep}{9pt}
\renewcommand{\arraystretch}{1.15} 
\begin{tabular}{lccccccc}
\toprule
\multirow{2}{*}{Method} & \multicolumn{7}{c}{NFE} \\
\cmidrule{2-8}
 & 3 & 4 & 5 & 6 & 7 & 8 & 9 \\
\midrule
iPNDM \cite{iPNDM,PNDM}        & 34.81 & 21.33 & 15.54 & 10.27 &  8.64 & 6.60 & 5.64 \\
DPM-Solver++(3M) \cite{dpmpp}  & 65.19 & 30.56 & 16.87 & 11.38 &  8.68 & 7.12 & 6.25 \\
UniPC \cite{UniPC}             & 91.38 & 55.63 & 24.36 & 14.30 &  9.57 & 7.52 & 6.34 \\
\midrule
AMED-Solver \cite{AMED}        & 38.10 & 32.69 & 10.74 & 10.63 &  6.66 & 7.71 & 5.44 \\ 
AMED-Plugin \cite{AMED}        & 28.06 &   -   & 13.83 &   -   &  7.81 &  -   & 5.60 \\ 
EPD-Solver \cite{EPD}          & \underline{18.28} &   -   &  \underline{6.35} &   -   &  5.26 &  -   & 4.27 \\
EPD-Plugin \cite{EPD}          & 19.89 &   -   &  8.17 &   -   &  \underline{4.81} &  -   & \underline{4.02} \\
LD3 \cite{LD3}                 & 27.82 & 17.03 & 11.55 &  7.53 &  5.63 & \underline{5.40} & 4.71 \\
DLMS \cite{DLMS}               &   -   & \underline{10.07} &  7.16 &  \underline{7.08} &  6.31 & 5.93 & 4.57 \\ 
\rowcolor{lightCyan}
DyWeight (ours)                & \textbf{17.37} & \textbf{9.62} &  \textbf{6.30} &  \textbf{6.15} &  \textbf{4.55} & \textbf{4.22} & \textbf{3.81} \\
\bottomrule
\end{tabular}
\end{table*}

\begin{table*}[t]
\caption{Image generation FID$\downarrow$~\cite{fid} on LSUN-Bedroom($256\times256$)~\cite{lsun}. The best results are in \textbf{bold}, the second best are \underline{underlined}.}
\label{tab:supp_lsun}
\centering
\fontsize{12}{14}\selectfont
\setlength{\tabcolsep}{9pt}
\renewcommand{\arraystretch}{1.15} 
\begin{tabular}{lccccccc}
\toprule
\multirow{2}{*}{Method} & \multicolumn{7}{c}{NFE} \\
\cmidrule{2-8}
 & 3 & 4 & 5 & 6 & 7 & 8 & 9 \\
\midrule
iPNDM \cite{iPNDM,PNDM}        & 43.31 & 26.14 & 18.46 & 12.81 & 11.03 &  9.49 & 7.89 \\
DPM-Solver++(3M) \cite{dpmpp}  &111.90 & 48.49 & 18.44 &  8.39 &  5.18 &  4.12 & 3.77 \\
UniPC \cite{UniPC}             &112.30 & 39.66 & 13.76 &  6.46 &  4.52 &  3.96 & 3.72 \\
\midrule
AMED-Solver \cite{AMED}        & 58.21 & 15.67 & 13.20 &  8.92 &  7.10 &  4.19 & 5.65 \\ 
AMED-Plugin \cite{AMED}        & 101.5 &   -   & 25.68 &  -   &  8.63 &  -   & 7.82 \\ 
EPD-Solver \cite{EPD}          & \underline{13.21} &   -   &  7.52 &  -   &  5.97 &  -   & 5.01 \\
EPD-Plugin \cite{EPD}          & 14.12 &   -   &  8.26 &  -   &  5.24 &  -   & 4.51 \\
LD3 \cite{LD3}                 & 14.62 &  8.48 &  5.93 &  4.52 &  4.16 &  4.22 & 3.98 \\
DLMS \cite{DLMS}               &   -   &  \underline{8.20} &  \underline{5.44} &  \underline{4.44} &  \underline{3.99} &  \underline{3.89} & \underline{3.70} \\ 
S4S-Alt \cite{s4s}             &   37.65 &  20.89 &  13.03 &  10.49 & 10.03 & 9.64 & - \\ 
\rowcolor{lightCyan}
DyWeight (ours)                & \textbf{9.82} &  \textbf{6.35} &  \textbf{4.97} &  \textbf{4.07} &  \textbf{3.87} &  \textbf{3.62} & \textbf{3.45} \\
\bottomrule
\end{tabular}
\end{table*}



\end{document}